\documentclass[letterpaper,11pt]{article}
\usepackage[margin=1in]{geometry}  

\usepackage{listings}
\usepackage{amsmath}
\usepackage{amsthm}
\usepackage{tikz}
\usepackage{caption,subcaption}
\usepackage{array}
\usepackage{mdwmath}
\usepackage{multirow}
\usepackage{mdwtab}
\usepackage{eqparbox}
\usepackage{multicol}
\usepackage{amsfonts}
\usepackage{tikz}
\usepackage{multirow,bigstrut,threeparttable}
\usepackage{amsthm}
\usepackage{array}
\usepackage{bbm}
\usepackage{epstopdf}
\usepackage{mdwmath}
\usepackage{mdwtab}
\usepackage{eqparbox}
\usepackage{tikz}
\usepackage{latexsym}
\usepackage[authoryear]{natbib}
\usepackage{amssymb}
\usepackage{bm}
\usepackage{amssymb}
\usepackage{graphicx}
\usepackage{mathrsfs}
\usepackage{epsfig}
\usepackage{psfrag}
\usepackage{setspace}
\usepackage{enumitem}
\usepackage[
            CJKbookmarks=true,
            bookmarksnumbered=true,
            bookmarksopen=true,
            hypertexnames=false,
            colorlinks=true,
            citecolor=red,
            linkcolor=blue,
            anchorcolor=red,
            urlcolor=blue
            ]{hyperref}
\usepackage[ruled]{algorithm2e}
\usepackage{algpseudocode}
\usepackage{stfloats}

\theoremstyle{plain}
\newtheorem{theorem}{Theorem}
\newtheorem{assumption}{Assumption}
\newtheorem{example}{Example}
\newtheorem{lemma}{Lemma}
\newtheorem{remark}{Remark}
\newtheorem{corollary}{Corollary}
\newtheorem{definition}{Definition}

\newtheorem{proposition}{Proposition}

\def \bP {\mathbb{P}}
\def \bE {\mathbb{E}}
\def \bR {\mathbb{R}}

\def \var {\mathsf{Var}}

\usepackage{xspace}

\newcommand{\stepa}[1]{\overset{\rm (a)}{#1}}
\newcommand{\stepb}[1]{\overset{\rm (b)}{#1}}
\newcommand{\stepc}[1]{\overset{\rm (c)}{#1}}
\newcommand{\stepd}[1]{\overset{\rm (d)}{#1}}
\newcommand{\stepe}[1]{\overset{\rm (e)}{#1}}
\newcommand{\stepf}[1]{\overset{\rm (f)}{#1}}

\newcommand{\ceil}[1]{{\left\lceil {#1} \right \rceil}}

\definecolor{myblue}{rgb}{.8, .8, 1}
\definecolor{mathblue}{rgb}{0.2472, 0.24, 0.6} 
\definecolor{mathred}{rgb}{0.6, 0.24, 0.442893}
\definecolor{mathyellow}{rgb}{0.6, 0.547014, 0.24}

\newcommand{\calN}{{\mathcal{N}}}

\newcommand{\calT}{{\mathcal{T}}}

\newcommand{\ZH}[1]{#1}
\newcommand{\Rtwo}[1]{#1}
\newcommand{\StochRev}[1]{#1}

\newcommand{\halmos}{\hfill\ensuremath{\square}}

\makeatletter
\renewenvironment{proof}[1]{%
  \par\addvspace{6pt}%
  \noindent{\itshape #1\enskip}%
  \ignorespaces
}{%
  \par\addvspace{0pt}%
}
\makeatother

\usepackage{cleveref}
\crefname{lemma}{Lemma}{Lemmas}
\Crefname{lemma}{Lemma}{Lemmas}
\crefname{thm}{Theorem}{Theorems}
\Crefname{thm}{Theorem}{Theorems}
\crefformat{equation}{(#2#1#3)}

\begin{document}

\title{Sequential Batch Learning in \\Finite-Action Linear Contextual Bandits}
\author{Yanjun Han, Zhengqing Zhou, Zihao Hu, Jose Blanchet, \\
Peter Glynn, Yinyu Ye, and Zhengyuan Zhou\thanks{Yanjun Han is with the Courant Institute of Mathematical Sciences and Center for Data Science, New York University. Zhengqing Zhou is with Susquehanna International Group. Zihao Hu is with the Department of Mathematics and IEDA, HKUST. Jose Blanchet, Peter Glynn, and Yinyu Ye are with the Department of Management Science and Engineering, Stanford University. Zhengyuan Zhou is with the Stern School of Business, New York University. Correspondence regarding this paper may be sent to \url{yanjunhan@nyu.edu} or \url{zz26@stern.nyu.edu}.}}

\maketitle

\begin{abstract}
We study the sequential batch learning problem in linear contextual bandits with finite action sets, where the decision maker is constrained to split incoming individuals into (at most) a fixed number of batches and can only observe outcomes for the individuals within a batch at the batch's end.
Compared with both standard online contextual-bandit learning and offline policy learning in contextual bandits, this sequential batch learning problem provides a finer-grained formulation of many personalized sequential decision making problems in practical applications, including medical treatment in clinical trials, product recommendation in e-commerce and adaptive experiment design in crowdsourcing.

We study two settings of the problem: one where the contexts are arbitrarily generated and the other where the context vectors are mutually independent across actions and time and follow a common Gaussian distribution.
In each setting, we establish a regret lower bound and provide an algorithm, whose regret upper bound nearly matches the lower bound.
As an important insight revealed therefrom, in the former setting, we show that the number of batches required to achieve the fully online performance is polynomial in the time horizon, while for the latter setting, a pure-exploitation algorithm with a judicious batch partition scheme achieves the fully online performance even when the number of batches is less than logarithmic in the time horizon. In the stochastic context setting, we additionally provide tight margin-based (i.e. instance-dependent) upper and lower regret bounds that delineate performance in terms of how difficult the problem instance is. Together, our results provide a near-complete characterization of sequential decision making in linear contextual bandits when batch constraints are present.
\end{abstract}


\tableofcontents

\section{Introduction}
\label{sec:intro}

With the rapid advances of digitization of the economy, massive amounts of user-specific data have become increasingly available. Among its varied implications, one that holds center-stage importance is the advent of the new era of data-driven personalized decision making: equipped with such user-specific data, decision makers across a wide variety of domains are now able to personalize the service decisions based on individuals' characteristics, thereby improving the outcomes. Fundamentally, such improved outcomes are achieved by intelligently exploring and exploiting the \Rtwo{heterogeneity} in a given population, which manifests itself as different individuals responding differently to the same treatments/recommendations/actions.
Such \Rtwo{heterogeneity} is ubiquitous across a variety of applications, including
medical treatment selection in clinical trials, product recommendation in marketing, 
order provisioning in inventory management,
hospital staffing in \StochRev{operating rooms, ad selection} in online advertising~\citep{bertsimas2007learning, LCLS2010, kim2011battle, he2012timing, chapelle2014, chen2015statistical, bastani2015online, schwartz2017customer, zhou2018evaluating, hespanhol2018family, ferreira2018online, ban2019big}.

Situated in this broader context and rising to materialize the value from personalization, contextual bandits have emerged to be the predominant mathematical framework that is at once rich and elegant. Its three modelling cores, contexts, actions, and rewards (representing individual features, recommendations and outcomes respectively), capture the salient aspects of personalized decision making and provide fertile ground for developing algorithms that contribute to making quality decisions at the fine-grained individual level.
As such, in the past decade, dedicated efforts have been devoted to this area, yielding a flourishing line of literature. Broadly speaking, the existing contextual bandits literature falls into two main categories: offline contextual bandits and online contextual bandits. 
In offline contextual bandits (\citet{langford2011doubly, zhang2012estimating,zhao2012estimating, zhao2014doubly, JMLR:v16:swaminathan15a, rakhlin2016bistro, athey2017efficient, kitagawa2018should, kallus2018confounding, zhou2018offline, deep-learning-logged-bandit-feedback}), the decision maker is given a full batch dataset that has been collected from historical observations.
The decision maker aims to learn from this dataset an effective policy (i.e. a function that maps contexts to decisions) that will be deployed in the future and that (hopefully) yields good performance.
As such, in offline contextual bandits,  the decision maker is not allowed to perform active learning: selecting a policy in this setting is one-shot and the decision maker is solely concerned with identifying the best policy using available data.

\ZH{On the other hand, in online contextual bandits, the decision-maker is an active participant in the learning process. As information arrives sequentially, the decision-maker adapts their strategy based on historical data, effectively choosing which feedback signals to reveal and incorporate into the model.} Typically, in such settings, a decision is made on the current individual based on all the past feedback, yielding an outcome that is immediately observed and incorporated to make the next decision. \ZH{A rich literature has studied this setting \citep{LCLS2010,rusmevichientong2010linearly,FCGS2010, rigollet2010nonparametric, chu2011contextual, goldenshluger2013linear, AG2013a, AG2013b, RV2014, russo2016information, JBNW2017,abeille2017linear,dimakopoulou2019balanced,li2017provably}. These works have developed various online contextual bandit algorithms, most notably those based on UCB and Thompson sampling. These methods are designed to effectively balance the trade-off between exploration and exploitation, which represents a key challenge in this domain.}
See further \citet{BCN2012,lattimore2020bandit,slivkins2019introduction} for three articulate surveys that more systematically describe the field.

However, in practice, it is often the case that neither setting provides a close approximation of the underlying reality. Specifically, that the decision maker can only perform one-shot policy learning--the key setting in offline contextual bandits--is simply too restrictive and pessimistic for almost all applications. On the other hand, assuming that the decision maker can constantly observe and incorporate feedback at a per-individual scale--the key setting \ZH{of} online contextual bandits--is also \ZH{an} over-simplification and too optimistic for many applications. In fact, reality often stands somewhere in between: decision makers across many applications are typically able to perform active learning and incorporate feedback from the past to adapt their decisions in the future; however, due to the physical and cost constraints, such adaptation is often limited to a fixed number of rounds of interaction.
For instance, in clinical trials~\citep{robbins1952some,thompson1933likelihood,kim2011battle}, each trial involves applying medical treatments to a group of patients, where the medical outcomes are observed and collected for the entire group at the end of the trial.
The data collected from previous trials are then analyzed to design the medical treatment selection schemes for the next trial.
Note that as the medical information from previous trials \StochRev{is} incorporated to inform
treatment selection in future trials, medical decision makers do \Rtwo{have the flexibility} in adaptive learning. However, such flexibility is limited since in practice, only a limited number of trials (e.g. $3$) can be conducted, far less than the number of patients in the trials, hence rendering the standard online learning models inapplicable here.
 Another example where adaptive learning is possible but with batch constraints is product recommendation in marketing~\citep{bertsimas2007learning, schwartz2017customer}.
 In this case, the marketer sends out product offers to a batch of customers at once when running a promotions campaign. Customers' feedback will then be batch collected at the end and analyzed to design the next round of promotions in the targeted population. Other examples include crowdsourcing~\citep{kittur2008crowdsourcing} and simulations~\citep{chick2009economic}, where both cases exhibit the characteristic that a single-run experiment consists of a batch of individuals.
 
 \subsection{Related Prior Work}
 \label{subsec:related_work}
 Motivated by these considerations, we study the problem of sequential batch learning in linear contextual bandits, where a decision maker can adaptively learn to make decisions subject to the batch constraints that, as described above, commonly occur in practice.  
 This \StochRev{batch-constrained bandit problem has} recently been studied in the simple multi-armed bandits (MAB) case. In particular, \cite{perchet2016batched} studied sequential batch learning in the 2-armed MAB problem,
 where they have given a successive elimination algorithm during each batch and established that 
 $O(\log \log T)$ batches are needed in order to achieve the same regret bound as in the standard online learning setting\ZH{, where $T$ denotes the time horizon}.  More recently, \cite{gao2019batched} generalized the results to the $K$-armed bandit setting and allowed the batch sizes to be chosen adaptively. \cite{Esfandiari_Karbasi_Mehrabian_Mirrokni_2021} has subsequently improved on the regret bound given in~\cite{gao2019batched} and additionally studied two extensions: 1) batched adversarial $K$-armed MABs and 2) batched (stochastic) infinite-armed MABs with a linear reward form, also known as linear bandits (which is different from linear contextual bandits).
 
 However, these initial efforts on MAB settings, despite providing interesting insights, cannot capture
 individuals' characteristics: in MABs, decisions can only be made at a population level (i.e. the decision maker aims to select an action that is the best for the entire population), rather than personalized at an individual level, which \StochRev{severely limits the practical applicability of MAB models}. In this paper, we fill in this gap by providing
 a comprehensive inquiry into sequential batch learning in linear contextual bandits with a finite number of actions, a \StochRev{canonical} setting where decisions are provisioned based on the individual features. 
 Our goal is then to delineate, in this more general setting, how the batch constraints impact the performance of adaptive decision making and characterize in depth the fundamental limits therein, thereby shedding light on how practical adaptive decision making can be most efficiently done in practice.
 
 \ZH{We demonstrate that batching efficiency diverges sharply across bandit paradigms. Stochastic settings (multi-armed, fixed-action linear, and contextual) permit highly efficient batching, requiring only $O(\log \log T)$ or $\Theta(\log \log (T/d^2))$ batches \Rtwo{(where $d$ denotes the context dimension)} to match fully online minimax regret. This efficiency arises because bounded context covariance limits initial regret to $\tilde{O}(1/\sqrt{d})$, enabling a large initial batch. Conversely, adversarial contextual bandits demand a heavy polynomial complexity of $\Theta(\sqrt{dT})$ batches, as adversaries can exploit delayed feedback with orthogonal contexts to force blind exploration and vastly more frequent policy updates.}

 
Our work is also related to but distinct from learning on bandits (either contextual or MABs) with delayed feedback~\citep{NAGS2010, DHKKLRZ2011, JGS2013, QK2015, GST2016, bis2019, zhou2019learning}: since the decision maker is not able to observe rewards in the interim of a batch (as they only come at the batch's end), feedback \StochRev{is} delayed from the decision maker's perspective.
 However, a key difference exists between learning in bandits and our sequential batch learning model: the former setting works with exogenous delays--drawn either from some stochastic process or from some arbitrary sequence--\Rtwo{that, unlike the delays in our setting, are neither influenced nor known by the decision maker}. The literature on learning in bandits with delayed feedback then develops adapted algorithms in the presence of delays and \StochRev{studies} how regret bounds scale as a function of the underlying delays. On the other hand, feedback delays in sequential batch learning are endogenous and arise as a consequence of the batch constraints; in particular, the decision maker in this setting \StochRev{has full discretion to choose} the batch sizes, which in turn \StochRev{determine} how the feedback for each individual is delayed. Consequently, the frameworks provided by the learning-in-bandits-with-delays literature--both the \StochRev{algorithms} and analyses--are not applicable here. Of course, it should also be pointed out that
 when viewed through the lens of learning with delayed feedback, the delays in our setting exhibit a particular structure: if a batch has size $B$, then the reward for the first item in the batch is delayed by \ZH{$B-1$} time units, the reward for the second item in the batch is delayed by \ZH{$B-2$} time units, and so on, and the reward for the last item in the batch has \StochRev{no delay}.
Consequently our results here also do not directly imply regret bounds in that literature.

\subsection{Our Contributions and Related Work}
\label{subsec:contributions}

\ZH{Throughout, we use standard asymptotic notation. We use $O(\cdot)$, $\Omega(\cdot)$ and $\Theta(\cdot)$ to denote asymptotic upper, lower and tight bounds, respectively. We use $\tilde{O}(\cdot)$, $\tilde{\Theta}(\cdot)$ to denote an asymptotic upper/tight bound that ignores polylogarithmic factors, and $\mathsf{polylog}(\cdot)$ to denote a polynomial in the logarithm of its argument.}
We study the sequential batch learning problem in linear contextual bandits when the number of actions is finite (and not too large, to be made precise later). Our main contributions are \StochRev{threefold}.
First, we consider the adversarial contexts setting, where the contexts can be generated arbitrarily or even adversarially. In this setting, we provide a UCB-style algorithm that is adapted to the sequential batch setting and achieves $\mathsf{polylog}(T)\cdot \ZH{\sqrt{\log K}}\left(\sqrt{dT} + \frac{dT}{M}\right)$ expected regret,
where \ZH{$K$ is the number of actions,} $d$ is the dimension of the context, $M$ is the number of allowed batches, and $T$ is the time horizon. This regret bound highlights that in the adversarial context setting, $\Theta(\sqrt{dT})$ batches--rather than $T$ batches--are sufficient to achieve the $\tilde{\Theta}(\sqrt{dT \log K})$ regret bound that is minimax optimal in the standard online learning setting. Further, we characterize the fundamental limits of learning in this setting by establishing a $c\cdot \left(\sqrt{dT} + \min \{\frac{T\sqrt{d}}{M}, \frac{T}{\sqrt{M}}\}\right)$ lower bound for the expected regret, where $c$ is some universal constant. 
This regret lower bound highlights that $\Omega(\sqrt{T})$ batches are needed for any algorithm to achieve the optimal performance for standard online learning (when no batch constraints are present). Consequently, our regret bound is tight up to at most a factor of $O(\sqrt{d \log K})$ (and other polylog factors). In the low-dimensional regime (i.e. $d$ considered a constant when compared to $T$), our regret bounds provide a complete characterization.

Second, we consider the stochastic contexts setting, where \StochRev{the context
vectors are mutually independent across actions and time and follow a common
Gaussian distribution}. In this case, we reveal an interesting phenomenon that stands in \StochRev{contrast} to the adversarial contexts setting: a simple pure-exploitation algorithm alone can achieve the $\tilde{\Theta}(\sqrt{dT \log K})$ regret bound (which, as mentioned above, is minimax optimal in standard online learning) using only $O(\log\log \frac{T}{d^2})$ \ZH{batches}, far less than what is required in the adversarial contexts setting. More specifically, we establish a $\mathsf{polylog}(T)\cdot \ZH{\sqrt{dT \log K}}\left(\frac{T}{d^2}\right)^{\frac{1}{2(2^M-1)}}$ upper bound
and a $c\cdot \ZH{\sqrt{dT \log K}}\left(\frac{T}{d^2}\right)^{\frac{1}{2(2^M-1)}}$ lower bound for the expected regret, respectively.  Consequently, up to polylog factors, our regret bound is exactly minimax optimal, indicating that
$\Theta(\log\log \frac{T}{d^2})$ batches are necessary and sufficient to achieve the optimal performance of the standard online learning setting. We focus on Gaussian contexts because they allow us to present the cleanest analysis that reveals the
nontrivially interesting nature of sequential batch learning, characterized by the above-mentioned regret bounds and their consequence. Such results can be extended to more general context distributions with appropriate covariate diversity assumptions (\cite{bastani2017mostly}), which we do not pursue here because the complex assumptions and the resulting notation/analysis would \StochRev{obscure} the  overall technical flow.

Third, under \StochRev{the stochastic-context model specified in
Section~\ref{subsec:adversarial_stochastic}}, we provide instance-dependent regret bounds and show that better bounds can be obtained for easy problem instances characterized by \StochRev{a large margin}. More specifically, we establish an upper bound of $\mathsf{polylog}(T)\cdot \ZH{\sqrt{\log K}} \cdot \frac{d^{3/2}}{\|\theta^\star\|_2} \left(\frac{T}{d^2}\right)^{\frac{1}{M}}$, where $\|\theta^\star\|_2$ is our characterization of the margin: the larger $\|\theta^\star\|_2$ is, the easier the problem becomes and the smaller the regret bound. Further, we show that this bound is essentially tight (up to polylogarithmic factors in $T$) by establishing that $\sup_{\theta^\star: \|\theta^\star\|_2 \le 1} \|\theta^\star\|_2 \cdot \mathbb{E}[R_T(\pi, \theta^\star)] \ge c \cdot \ZH{\sqrt{\log K}} \cdot d^{3/2} \left( \frac{T}{d^2} \right)^{\frac{1}{M}}$, for some constant $c$ independent of $T, M, d, K$, \ZH{where $\mathbb{E}[\cdot]$ denotes the expectation over the random environment (contexts, rewards, etc.) under the fixed parameter $\theta^\star$ and $R_T(\pi, \theta^\star)$ is the regret of the policy $\pi$ parameterized by $\theta^\star$}. Consequently, $\Theta(\log(T/d^2))$ batches are necessary and sufficient to achieve the optimal instance-dependent regret $\tilde{\Theta}(\ZH{\sqrt{\log K}} \cdot d^{3/2} / \|\theta^\star\|_2)$, \StochRev{where $\tilde{\Theta}$ hides $\mathsf{polylog}(T)$ factors.}

\ZH{In standard multi-armed bandit (MAB) settings, it is well known that adaptive batch sizes can significantly improve the regret's dependence on the number of arms $K$, for instance, by utilizing arm-elimination to dynamically shrink the active action set \Rtwo{\citep{Esfandiari_Karbasi_Mehrabian_Mirrokni_2021}}. \Rtwo{\StochRev{For the stochastic-context model specified in Section~\ref{subsec:adversarial_stochastic},} our matching upper and lower bounds show that the $\sqrt{\log K}$ dependence persists under sequential batching. For adversarial contexts, our UCB-style upper bound also contains a $\sqrt{\log K}$ factor, but the corresponding lower bound does not establish that this dependence is unavoidable. Whether a different adaptive batching method can improve the $K$-dependence in the adversarial-context setting remains open.}}


\ZH{Since our initial preprint~\cite{han2020sequential}, related works \StochRev{such as} \cite{ruan2021linear} \StochRev{have explored} a ``rare policy switch'' model allowing adaptive batch termination. While flexible, our framework requires upfront batch size determination, better reflecting practical planning constraints like marketing campaigns. Furthermore, while \cite{ruan2021linear} identifies the batch count needed to match online performance in stochastic settings (e.g., $M = \ceil{\log \log T}$), we uniquely characterize minimax optimal bounds for \StochRev{any given} batch constraint $M$ (e.g., $M \in \{3, 4\}$ for clinical trials). Finally, we extend our analysis to adversarial contexts and establish instance-dependent bounds \StochRev{(we also note that} \cite{zhang2021almost} proposes an empirical variant of \cite{ruan2021linear}\StochRev{)}.}

\section{Problem Formulation}
\label{sec:problem_formulation}
We introduce the problem of sequential batch learning on finite-action linear contextual bandits.

\subsection{Notation}
\label{subsec:notation}
For a positive integer $n$, let $[n]\triangleq\{1,\cdots,n\}$. For real numbers $a,b$, let $a\wedge b\triangleq \min\{a,b\}$. For a vector $v$, let $v^\top$ and $\|v\|_2$ be the transpose and $\ell_2$ norm of $v$, respectively. For square matrices $A,B$, let $\mathsf{Tr}(A)$ be the trace of $A$, and let $A\preceq B$ denote that the difference $B-A$ is symmetric and positive semi-definite. We adopt the standard asymptotic notations: for two non-negative sequences $\{a_n\}$ and $\{b_n\}$, let $a_n=O(b_n)$ iff $\limsup_{n\to\infty} a_n/b_n<\infty$, $a_n=\Omega(b_n)$ iff $b_n=O(a_n)$, and $a_n=\Theta(b_n)$ iff $a_n=O(b_n)$ and $b_n=O(a_n)$. We also write $\tilde{O}(\cdot), \tilde{\Omega}(\cdot)$ and $\tilde{\Theta}(\cdot)$ to denote the respective meanings within multiplicative logarithmic factors in $n$. For probability measures $P$ and $Q$, let $P\otimes Q$ be the product measure with marginals $P$ and $Q$. If measures $P$ and $Q$ are defined on the same probability space, we denote by $\mathsf{TV}(P,Q) = \frac{1}{2}\int |dP-dQ| $ and $ D_{\text{KL}}(P\|Q) = \int dP\log\frac{dP}{dQ}$ the total variation distance and Kullback--Leibler (KL) \StochRev{divergence} between $P$ and $Q$, respectively. \ZH{We write regret as $R_T(\textbf{Alg},\theta^\star)$ (Definition~\ref{def:regret}), and $\mathbb{E}[R_T(\textbf{Alg},\theta^\star)]$ for its expectation over the randomness of contexts and rewards for fixed $\theta^\star$.}

\subsection{Decision Procedure and Reward Structures}
\label{subsec:decision}
Let $T$ be the time horizon of the problem. At the beginning of each time $t\in [T]$, the decision maker observes a set of $K$ $d$-dimensional feature vectors (i.e. contexts) $\{x_{t,a} \mid a \in [K]\} \subseteq \bR^d$ corresponding to the $t$-th unit.
If the decision maker selects action $a \in [K]$, then a reward $r_{t,a} \in \bR$ corresponding to time $t$ is incurred (although not necessarily immediately observed).
We assume the mean reward is linear: that is, there exists an underlying (but unknown) parameter
$\theta^\star$ such that 
 $$r_{t,a} =  x_{t,a}^\top \theta^\star + \xi_t,$$
 where $\{\xi_t\}_{t=0}^{\infty}$ is a sequence of zero-mean independent sub-Gaussian random variables with a uniform upper bound on the sub-Gaussian constants. Without loss of generality and for notational simplicity,
 we assume each $\xi_t$ is $1$-sub-Gaussian:  $\mathbf{E}[e^{\lambda \xi_t}] \le e^{\frac{\lambda^2}{2}}, \forall t, \forall \lambda \in \bR$.
 Further, without loss of generality (via normalization), we assume $\|\theta^\star\|_2\le 1$.
 We denote by $a_t$ and $r_{t, a_t}$ the action chosen and the reward obtained at time $t$, respectively. 
 Note that both are random variables; in particular, $a_t$ \Rtwo{could be random} either because the action is randomly selected based on the contexts $\{x_{t,a} \mid a \in [K]\}$ or because the contexts $\{x_{t,a} \mid a \in [K]\}$
 are random, or both.

\ZH{As there are different (but equivalent) formulations of contextual bandits, we briefly discuss the meaning of the above abstract quantities and how they arise in practice through the following example.
	\begin{example}
	Consider the clinical-trial scenario for medical treatments mentioned in Section~\ref{sec:intro}. In this context, the index $t$ represents the $t$-th patient participating in the trial.
\begin{itemize}
    \item The raw characteristics $v_t$ encompass the individual-specific health traits of the $t$-th patient, such as age, blood pressure, genetic markers, and pre-existing medical conditions.
    \item The action $a_t \in [K]$ denotes the specific medical treatment or dosage assigned to the patient from a finite set of $K$ available options (e.g., different drugs or a placebo).
    \item The observed outcome $y_t(v_t, a_t)$ corresponds to the realized reward for patient $t$ after receiving treatment $a_t$; in the notation above, this is the quantity $r_{t,a_t}$ (e.g., a measurable health improvement such as reduction in a symptom's severity).
    \item For each $a \in [K]$, the feature map produces $\phi(v_t, a)$ and hence a context vector $x_{t,a} = \phi(v_t, a)$ before the action is chosen; after the learner selects $a_t$, the realized context is $x_{t,a_t}$. For example, $x_{t,a}$ might include main effects for the treatments as well as interaction terms between specific genetic markers in $v_t$ and treatment indicators.
\end{itemize}
The learning algorithm then uses these featurized contexts $\{x_{t,a}\}_{a \in [K]}$ and the linear contextual bandit framework to estimate $\theta^\star$, which maps features to the mean outcome $\mathbf{E}[y_t(v_t, a_t) \mid v_t, a_t] = x_{t,a_t}^\top \theta^\star$.
\end{example}}

\subsection{Sequential Batch Learning}
In the standard online learning setting, the decision maker immediately observes the reward $r_{t, a_t}$ after selecting action $a_t$ at time $t$. Consequently, in selecting $a_t$, the decision maker can base his decision on all the past contexts $\{x_{\tau,a} \mid a \in [K],  \tau\le t\}$ and all the past rewards $\{r_{\tau,a_\tau} \mid \tau\le t-1\}$. 

In contrast, we consider a \textit{sequential batch learning} setting, where
the decision maker is allowed to partition the $T$ units into at most $M$
batches, and the reward corresponding to each unit in a batch can only be
observed at the end of the batch. More specifically, given a maximum
\StochRev{number of batches} $M$, the decision maker needs to choose a sequential batch learning
algorithm \textbf{Alg} that has the following two components:
\begin{enumerate}
	\item A \emph{grid} $\calT=\{t_1,t_2,\cdots,t_M\}$, with $0 = t_0 < t_1<t_2<\cdots<t_M=T$.
		Intuitively, this grid partitions the $T$ units into $M$ batches: the $k$-th batch contains units $t_{k-1}+1$ to $t_k$. Note that without loss of generality, the decision maker will always choose a grid of $M$ batches (despite being allowed to choose \StochRev{fewer} than $M$ batches) since choosing \StochRev{fewer} than $M$ batches will only decrease the amount of information available and hence yields worse performance. More formally, for any sequential batch learning algorithm that uses a grid of size less than $M$, there exists another sequential batch learning algorithm that uses a grid of size $M$ whose performance is no worse. 
	
		\item A sequential batch policy $\pi = (\pi_1, \pi_2, \dots, \pi_T)$ such that each $\pi_t$ can only use reward information from all the prior batches (as well as all the contexts that can be observed up to $t$). More specifically, for any $t \in [T]$, define the batched history $\mathcal{H}^t = \{x_{\tau,a} \mid a \in [K]\}_{\tau=1}^t \cup \{r_{\tau, a_\tau}\}_{\tau=1}^{\StochRev{t_{j(t)-1}}}$, where \ZH{$j(t)\in\{1,2,\dots,M\}$ is the unique integer satisfying
	$t_{j(t)-1}<t\le t_{j(t)}$}. Intuitively, there are $j(t) - 1$ batches prior to unit $t$: only the rewards of those batches have been observed. With this definition, a sequential batch policy is any policy $\pi$ such that
    each $\pi_t$ is adapted to $\mathcal{H}^t$ for each $t \in [T]$.
\end{enumerate}

\begin{remark}\label{remark:grid}
$M = T$ yields the standard online learning setting, where the decision maker does not need to select a grid. Consequently, the sequential batch learning setting has a more complex decision space--one that entails selecting both the grid and the policy. 
\end{remark}

\subsection{Performance Metric: Regret}
To assess the performance of a given sequential batch learning algorithm
\textbf{Alg}, we compare its cumulative reward with that obtained by an
\textbf{optimal} policy that has prescient knowledge of the optimal action for
each unit (i.e., an oracle that knows $\theta^\star$). This is formalized by the
following definition of regret.

\begin{definition}\label{def:regret}
Let \textbf{Alg} $= (\calT, \pi)$ be a sequential batch learning algorithm.
The regret of \textbf{Alg} is:
\begin{align}\label{eq.regret}
R_T(\textbf{Alg},\theta^\star) \triangleq \sum_{t=1}^T \left( \max_{a\in [K]}x_{t,a}^\top \theta^\star  - x_{t,a_t}^\top \theta^\star \right),
\end{align}
where $a_1, a_2, \dots, a_T$ are actions generated by \textbf{Alg} in the online decision making process. 	
\end{definition}	

\begin{remark}
Although the form of regret defined here is the same as that in standard online learning, the goal here is much more ambitious, because batches induce delays in obtaining reward feedback, and hence the decision maker cannot immediately incorporate the feedback into his subsequent decision making process.
Nevertheless, we still make the performance comparison to the oracle that is used in the standard online learning setting.   	
Further, note that $R_T(\textbf{Alg},\theta^\star)$ is a random variable since, as mentioned earlier, $a_t$ \Rtwo{could be random}.
For simplicity, we will mostly focus on bounding the expected regret, where the expectation is taken with respect to all sources of randomness (to be made precise later in various settings). However, high-probability regret bounds can also be obtained in our setting, and we will discuss them in relevant places throughout the paper. 
\end{remark}	

\subsection{Adversarial Contexts \StochRev{versus} Stochastic Contexts}
\label{subsec:adversarial_stochastic}
The regret defined in Definition~\ref{def:regret} is also a function of all the contexts that arrive over a horizon of $T$.
In the bandits literature, depending on how these contexts are generated, there are two main categories.
The first category is adversarial contexts, where at each $t$, an adversary can choose the contexts $\{x_{t,a} \mid a \in [K]\}$ that will be revealed to the decision maker. The second category is stochastic contexts. \StochRev{In the stochastic model studied here, the collection $\{x_{t,a}:t\in[T],\,a\in[K]\}$ is mutually independent, and every context follows a Gaussian distribution with a common covariance matrix.}

In the standard online learning case (i.e., $M=T$), the optimal regret bounds under adversarial or stochastic contexts are both $\tilde{\Theta}(\sqrt{dT})$ (see~\cite{chu2011contextual} and Theorem \ref{thm.stochastic} below). However, it turns out that in the sequential batch setting, there is a sharp difference between the adversarial contexts case and the stochastic contexts case. 
In particular, it turns out that the power of the adversary in choosing the contexts has a profound impact on what regret bounds can be obtained. Consequently, we divide our presentation into these two distinct regimes: we study the adversarial contexts case in Section~\ref{sec:adversarial} and the stochastic contexts case in Section~\ref{sec:stochastic}.
In each of these two settings, we give upper and lower bounds for regret.
Finally, throughout the paper, we consider the standard low-dimensional contextual bandits regime where the action set is not too large, as made precise by the following assumption.
\begin{assumption}\label{aspn.TKd}
$K=O(\mathsf{poly}(d))$ and $T\ge d^2$.
\end{assumption}
That is, we are assuming that the number of actions cannot be too large (at most polynomial in the dimension of the context), for there is a phase transition on the regret when $K$ can be exponential in $d$ (see~\cite{abe2003reinforcement}). \Rtwo{Further, the time horizon must be sufficiently large to permit accurate estimation of $\theta^\star$. Even under isotropic Gaussian contexts with covariance $I_d/d$, the mean-squared $\ell_2$ estimation error scales as $d^2/T$. Hence, $T=\Omega(d^2)$ is necessary to estimate $\theta^\star$ with root-mean-squared $\ell_2$ error bounded by a constant.}
Equivalently, we are in the low-dimensional contextual bandits setting, where the number of covariates is small compared to the number of samples we receive over the entire horizon.

\subsection{Intuition of Technical Results}
\label{subsec:intuition}
Our main technical result demonstrates that $O(\log\log \frac{T}{d^2})$ batches are sufficient to achieve the optimal regret bound under \StochRev{the stochastic-context model specified in Section~\ref{subsec:adversarial_stochastic}}. This is significantly more efficient than the adversarial setting, which requires $\Theta(\sqrt{dT})$ batches to achieve the same fully online regret performance.
	To further appreciate this result, and to get some intuition into why such bounds are possible, consider the special case where $M = 3$ (not uncommon in a typical clinical trial) and \StochRev{the contexts are mutually independent across actions and time, with each context following} $\mathcal{N}(0, \frac{\mathbf{I}_d}{d})$ (dividing the identity matrix by $d$ simply ensures that the norm of each context is bounded by $1$ in mean square). \StochRev{This isotropic example is used only for intuition. Our upper bounds allow any common covariance matrix satisfying Assumption~\ref{assumption:cov}, while retaining independence across actions and time.}
In this case, our bound indicates that the optimal performance is $\tilde{O}(d^{\frac{5}{14}} T^{\frac{4}{7}})$, which is already quite close to $O(\sqrt{dT})$; further,
pure exploitation--selecting the best action given our current estimate--on each batch
is able to achieve this regret bound.
Why?
To get a rough sense, let's allocate the $T$ units into three batches in the following way: the first batch contains $O(d^{\frac{6}{7}} T^{\frac{4}{7}})$ units, the second batch contains $O(d^{\frac{2}{7}} T^{\frac{6}{7}})$ units and the third batch contains the rest.
Now, for the moment, let's assume that the contexts selected over time are \textit{\StochRev{i.i.d.}}: of course they are not \textit{\StochRev{i.i.d.}}, because how one context is chosen at time $t$ depends critically on how contexts were chosen in the previous times (otherwise, there is no learning that occurs); but for simplicity, let's assume they are. 
Then the regret incurred on the first batch--since we haven't observed anything and hence know nothing-- is $\tilde{O}(\frac{d^{\frac{6}{7}} T^{\frac{4}{7}}}{\sqrt{d}}) = \tilde{O}(d^{\frac{5}{14}} T^{\frac{4}{7}})$, since each unit incurs $\tilde{O}(\frac{1}{\sqrt{d}})$ regret (this is a consequence of the normalization as well as a separate but simple analysis of instantaneous regret).
	Now after observing the results from the first batch and \StochRev{running least-squares regression on those observations}, we have an estimate of the underlying \StochRev{parameter} that achieves a certain level of accuracy. 
How accurate is our estimate now? 
From standard theory on linear regression (which says that if each covariate sample is \StochRev{drawn i.i.d.} from a standard \StochRev{multivariate Gaussian} and there are $n$ samples, then with high probability $\|\hat{\theta} -\theta^\star\|_2 = O(\sqrt{\frac{d}{n}})$), we can deduce that after observing $O(d^{\frac{6}{7}} T^{\frac{4}{7}})$ samples in the first batch, we would be able to achieve the following estimation accuracy:
 $\|\hat{\theta} -\theta^\star\|_2 = O_p(\sqrt{d} \sqrt{\frac{d}{d^{\frac{6}{7}} T^{\frac{4}{7}}}}  ) = O(\frac{1}{d^{-\frac{4}{7}} T^{\frac{2}{7}}})$, where the extra $\sqrt{d}$ factor is again a result of our normalization on the covariance matrix.
	Now, \StochRev{a more accurate estimate of} $\theta^\star$ would yield smaller regret for each individual unit,
and if each individual's regret is proportional to $\|\hat{\theta} -\theta^\star\|_2$--as it turns out to be--then the total regret for the second batch is the number of units in that batch times each individual regret, yielding $\tilde{O}(\frac{1}{\sqrt{d}}) \times O(\frac{d^{\frac{2}{7}} T^{\frac{6}{7}}}{d^{-\frac{4}{7}} T^{\frac{2}{7}}}) = \tilde{O}(d^{\frac{5}{14}} T^{\frac{4}{7}})$, where the $\tilde{O}(\frac{1}{\sqrt{d}})$ factor is the proportionality constant that, as before, comes from the normalization factor on the covariance matrix.
Finally, after observing all the $O(d^{\frac{2}{7}} T^{\frac{6}{7}})$ units in the second batch, 
our estimation accuracy would further improve to 
\ZH{$\|\hat{\theta} -\theta^\star\|_2 = O_p(\sqrt{d} \sqrt{\frac{d}{d^{\frac{2}{7}} T^{\frac{6}{7}}}}  ) = O(\frac{1}{d^{-\frac{6}{7}} T^{\frac{3}{7}}})$.}
Consequently, since there are $O(T)$ units in the third batch, the total regret
in this batch would be--by applying the same line of reasoning as in the second batch--
 $\tilde{O}(\frac{1}{\sqrt{d}}) \times O(\frac{T}{d^{-\frac{6}{7}} T^{\frac{3}{7}}}) = \tilde{O}(d^{\frac{5}{14}} T^{\frac{4}{7}})$.
As such, aggregating over all three batches, we obtain $\tilde{O}(d^{\frac{5}{14}} T^{\frac{4}{7}})$ total regret.
	\StochRev{It remains to show that the selected contexts form a sufficiently well-conditioned matrix for the standard linear-regression estimation rate to apply, despite their non-i.i.d. selection. As shown below, this is indeed the case.}

	Additionally, we \StochRev{give} gap-dependent regret bounds--both upper and lower bounds--in the stochastic contexts setting (note that there is no notion of gap when the contexts can be arbitrary). These bounds are typically \StochRev{sharper than} their gap-independent counterparts (to which the above-mentioned regret bounds belong) when the gap is large; in particular, the dependence on $T$ would
be logarithmic rather than $\sqrt{T}$. Section~\ref{sec:gap} provides a more detailed discussion on this, including how the pure-exploitation algorithm should be (slightly) modified in this case to achieve the minimax optimal gap-dependent regret bound.
Finally, we mention that our analyses easily yield high-probability regret bounds as well for all settings, although for simplicity, we have chosen only to present bounds on expected regret.

\section{Learning with Adversarial Contexts}
\label{sec:adversarial}

In this section, we focus on the adversarial contexts case. Following \citet{auer2002using}, we assume \StochRev{that,} at time $t\in [T]$, the contexts $x_{t,1},\cdots,x_{t,K}$ are chosen by an oblivious adversary, with $\|x_{t,a}\|_2\le 1$ for any $a\in [K]$. 
We first state the main results in Section~\ref{subsec:adversarial_main}, which
characterize the upper and lower bounds of the regret.
We then give a UCB-based algorithm in the sequential batch setting in Section \ref{subsec:adversarial_UCB} and describe several important aspects of the algorithm, including a variant that is used for theoretical bound purposes. 
Next, in Section \ref{subsec.UCB_upperbound}, we summarize the regret upper-bound analysis for Theorem \ref{thm.adversarial} (with the full proof in Appendix~\ref{appendix.adversarial.upper}). 
Finally, we prove the regret lower bound in Section \ref{subsec.adversarial} and therefore establish that the previous upper bound is \StochRev{close to tight}. 

\subsection{Main Results}\label{subsec:adversarial_main}
\begin{theorem}\label{thm.adversarial}
	\begin{enumerate}
		\item Under Assumption~\ref{aspn.TKd}, there exists a sequential batch learning algorithm \textbf{Alg}= $(\calT, \pi)$, where $\calT$ is a uniform grid  defined by $t_m = \lfloor \frac{mT}{M}\rfloor$  and $\pi$ is explicitly defined in Section \ref{subsec:adversarial_UCB},
		such that:
		\begin{align*}
		\sup_{\theta^\star: \|\theta^\star\|_2\le 1} \mathbb{E}[R_T\left(\textbf{Alg}, \theta^\star\right)] \le O\left(\sqrt{\log(KT)}\log(T)\right)\cdot \left(\sqrt{dT} + \frac{dT}{M}\right).
		\end{align*}
		\item  Conversely, for $K=2$ and any sequential batch learning algorithm, we have:
		\begin{align*}
		\sup_{\theta^\star: \|\theta^\star\|_2\le 1} \mathbb{E}[R_T\left(\textbf{Alg}, \theta^\star\right)] \ge c\cdot \left(\sqrt{dT} + \left(\frac{T\sqrt{d}}{M}\wedge \frac{T}{\sqrt{M}}\right)\right),
		\end{align*}
		where $c>0$ is a universal constant independent of $(T,M,d)$. 
	\end{enumerate}
\end{theorem}


\ZH{A natural question arises regarding the choice of the uniform grid $\calT$ in Theorem \ref{thm.adversarial}. In the adversarial setting, a worst-case adversary can actively select contexts prior to a batch that are orthogonal to those within it. Because the learner cannot update its policy mid-batch, it is forced into blind exploration, causing the regret in any single batch $m$ to scale linearly with its length $B_m = t_m - t_{m-1}$. As our analysis reveals, the total regret bound scales as $\tilde{\mathcal{O}}(\sqrt{dT} + d B_{\max})$, where $B_{\max} = \max_m B_m$. To minimize this worst-case upper bound subject to the fixed horizon constraint $\sum_{m=1}^M B_m = T$, one must minimize $B_{\max}$. This minimax optimization uniquely yields the uniform grid $B_m = T/M$ for all $m \in [M]$, the optimality of which is rigorously confirmed by the arbitrary-grid lower bound in Statement 2.

Furthermore, while the proposed sequential batch learning algorithm (Algorithm \ref{algo.ucb}) is an intuitive adaptation of the standard online LinUCB algorithm, establishing its theoretical limits in the batched setting is highly non-trivial. First, standard online trace-determinant lemmas do not readily yield the optimal additive regret bound of $\tilde{\mathcal{O}}(\sqrt{dT} + dT/M)$ under delayed updates; bounding the cumulative variance over batches requires a refined algebraic analysis. Second, delayed batch feedback fundamentally breaks the standard conditional independence required to construct valid confidence ellipsoids. Overcoming this requires the careful construction of a phase-based master algorithm to decouple these complex dependencies, thereby proving that this straightforward adaptation is fundamentally minimax optimal.}

Our subsequent analysis easily gives high-probability regret upper bounds. However, for simplicity and to highlight more clearly the matching between the upper and lower bounds, 
we stick with presenting results on expected regret.
Theorem \ref{thm.adversarial} shows a polynomial dependence of the regret on the number of batches $M$ under adversarial contexts, and the following corollary is immediate. 
\begin{corollary}\label{cor.adversarial}
	Under adversarial contexts, $\Theta(\sqrt{dT})$ batches achieve the fully online regret $\tilde{\Theta}(\sqrt{dT})$. 
\end{corollary}

According to Corollary \ref{cor.adversarial}, $T$ batches are not necessary to achieve the fully online performance under adversarial contexts: $\Theta(\sqrt{Td})$ batches suffice. Since we are \text{not} in the high-dimensional regime (per Assumption~\ref{aspn.TKd}, $d \le \sqrt{T}$), the number of batches needed without any performance suffering is at most $O(T^{0.75})$, a sizable reduction from $O(T)$. Further, in the low-dimensional regime (i.e. when $d$ is a constant), only $O(\sqrt{T})$ batches are needed to achieve fully online performance.
Nevertheless, $O(\sqrt{dT})$ can still be a fairly large number. 
In particular, if only a constant number of batches are available, then the regret is \Rtwo{linear in $T$}. The lower bound shows that this linear-regret conclusion cannot be substantially improved under adversarial contexts.
This is because the power of the adversary under adversarial contexts is too strong when the learner only has a few batches: the adversary may simply pick any batch and choose all contexts \StochRev{preceding} this batch to be \StochRev{orthogonal to} the contexts within this batch, \StochRev{so that} the learner can learn nothing about the rewards in any given batch.

\subsection{A Sequential Batch UCB Algorithm}\label{subsec:adversarial_UCB}
The overall idea of the algorithm is that, at the end of every batch, the learner computes an estimate $\hat{\theta}$ of the unknown parameter $\theta^\star$ via ridge regression as well as a confidence set that contains $\theta^\star$ with high probability. Then, whenever the learner enters a new batch, at each time $t$ he simply picks the action with the largest upper confidence bound. Finally, we choose the uniform grid, i.e., $t_m = \lfloor \frac{mT}{M}\rfloor$ for each $m\in [M]$. The algorithm is formally illustrated in Algorithm \ref{algo.ucb}.

\begin{algorithm}[h!]
	\relscale{0.9}%
	\DontPrintSemicolon  
	\SetAlgoLined
	\BlankLine
	\caption{Sequential Batch UCB (SBUCB) \label{algo.ucb}}
	\textbf{Input:} time horizon $T$; context dimension $d$; number of batches $M$; tuning parameter $\gamma>0$.
	
	\textbf{Grid choice:} $\calT = \{t_1,\cdots,t_M\}$ with $t_m = \lfloor \frac{mT}{M}\rfloor$. 
	
	\textbf{Initialization:} $A_0 = I_d\in \bR^{d\times d}$, \Rtwo{$b_0={\bf 0}\in\bR^d$}, $\hat{\theta}_0={\bf 0}\in \bR^d$, $t_0 = 0$.
	
	\For{$m \gets 1$ \KwTo $M$}{
			\For{$t\gets t_{m-1}+1$ \KwTo $t_m$}{
				Choose $a_t = \arg\max_{a\in [K]} x_{t,a}^\top \hat{\theta}_{m-1} + \gamma\sqrt{x_{t,a}^\top A^{-1}_{m-1} x_{t,a}}$ (break ties arbitrarily). \\
			}
		Receive rewards in the $m$-th batch: $\{r_{t,a_t}\}_{t_{m-1}+1 \le t \le t_m}$. 
		
		$A_m =  A_{m-1} + \sum_{t=t_{m-1}+1}^{t_m}  x_{t,a_t}x_{t,a_t}^\top$. \\
		\Rtwo{$b_m=b_{m-1}+\sum_{t=t_{m-1}+1}^{t_m}r_{t,a_t}x_{t,a_t}$.} \\
		\Rtwo{$\hat{\theta}_m=A_m^{-1}b_m$.}
	}
\end{algorithm}

\begin{remark}
Note that when $M=T$ (i.e. the fully online setting), Algorithm~\ref{algo.ucb} degenerates to the standard LinUCB algorithm in~\cite{chu2011contextual}.
\end{remark}	

To analyze the sequential batch UCB algorithm, we need to first show that the constructed confidence bound is feasible. By applying \cite[Lemma 1]{chu2011contextual} to our setting, we immediately obtain the following concentration result that the estimated $\hat{\theta}_{m-1}$ is close to the true $\theta^\star$: 

\begin{lemma}\label{lemma.concentration}
	Fix any $\delta > 0$.
	For each $m\in [M]$, if for a fixed sequence of selected contexts $\{x_{t,a_t}\}_{t\in [t_m]}$ up to time $t_m$, the (random) rewards $\{r_{t,a_t}\}_{t\in [t_m]}$  are independent, then for each $t \in [t_{m-1}+1, t_m]$,
	with probability at least $1-\frac{\delta}{T}$, the following holds for all $a\in [K]$:
	\begin{align*}
	|x_{t,a}^\top (\hat{\theta}_{m-1} - \theta^\star)| \le \left(1+\sqrt{\frac{1}{2}\log\left(\frac{2KT}{\delta}\right)}\right)\sqrt{x_{t,a}^\top A_{m-1}^{-1}x_{t,a}}.
	\end{align*}
\end{lemma}

\begin{remark}
Lemma~\ref{lemma.concentration} rests on an important conditional independence assumption of the rewards $\{r_{t,a_t}\}_{t\in [t_m]}$. However, this assumption
does not hold in the vanilla version of the algorithm as given in Algorithm~\ref{algo.ucb}.
This is because a future selected action $a_t$ and hence the chosen context $x_{t,a_t}$ depends on  
the previous rewards. Consequently, by conditioning on $x_{t,a_t}$, previous rewards, say $r_{\tau_1}, r_{\tau_2}$ ($\tau_1, \tau_2 < t$), can become dependent. Note the somewhat subtle issue here on
the dependence of the rewards: when conditioning on $x_{t,a_t}$, the corresponding reward $r_t$ becomes independent of all the past rewards $\{r_\tau\}_{\tau < t}$. Despite this, when a future $x_{t^\prime, a_{t^\prime}}$ is revealed ($t^\prime > t$), these rewards (i.e. $r_t$ and all the rewards prior to $r_t$) become coupled again: what was known about $r_t$ now reveals information about the previous rewards $\{r_\tau\}_{\tau < t}$, because $r_t$ itself would not determine the selection of $x_{t^\prime, a_{t^\prime}}$:
all those rewards have influence over $x_{t^\prime, a_{t^\prime}}$. Consequently, a complicated dependence structure is thus created when conditioning on $\{x_{t,a_t}\}_{t\in [t_m]}$.

This lack of independence issue will be handled with a master algorithm variant of Algorithm~\ref{algo.ucb} presented in Section~\ref{subsec.UCB_upperbound}. 
Using the master algorithm to decouple dependencies is a standard technique in contextual bandits that was first developed in~\cite{auer2002using}. Subsequently, it has been used for the same purpose in~\cite{chu2011contextual, li2017provably}, among others. We end this subsection by pointing out that, strictly speaking, our regret upper bound is achieved only by this master algorithm, rather than Algorithm~\ref{algo.ucb}. However, we take the conventional view that the master algorithm is purely used as a theoretical construct (to resolve the dependence issue) rather than a practical algorithm that should actually be deployed in practice. In practice, Algorithm~\ref{algo.ucb} should be used instead. The complete proof of Theorem~\ref{thm.adversarial}, Statement~1, including the trace lemmas and the master--base analysis, is given in Appendix~\ref{appendix.adversarial.upper}.
\end{remark}

\subsection{Regret Analysis for Upper bound}\label{subsec.UCB_upperbound}

The complete proof of Theorem~\ref{thm.adversarial}, Statement~1, is given in Appendix~\ref{appendix.adversarial.upper}. We record two auxiliary lemmas there---Lemma~\ref{lemma.eigenvalue} (a trace inequality for nested PSD matrices) and Lemma~\ref{lemma.trace_sum} (bounding the cumulative batched variance term $\sum_m \sqrt{\mathsf{Tr}(A_{m-1}^{-1}X_m)}$)---which replace the standard fully-online trace--determinant tools when updates occur only at batch boundaries.

At a high level, the analysis proceeds in two steps. First, \emph{assuming} the reward sequence satisfies the conditional independence hypothesis of Lemma~\ref{lemma.concentration}, the UCB decomposition for the sequential batch update rule in Algorithm~\ref{algo.ucb} reduces regret to controlling $\sum_m \sqrt{\mathsf{Tr}(A_{m-1}^{-1}X_m)}$; Lemma~\ref{lemma.trace_sum} supplies the required bound, including a careful correction for using $A_{m-1}^{-1}$ rather than $A_m^{-1}$ in the trace. Second, vanilla SBUCB does not satisfy conditional independence; we therefore analyze the master--base pair SupSBUCB/BaseSBUCB below, which masks past samples by stages so that Lemma~\ref{lemma.concentration} applies. The additional $O(\log T)$ factor in the final regret bound is standard for this master construction~\cite{auer2002using,chu2011contextual}.

\textbf{Master--base algorithms.} SupSBUCB (Algorithm~\ref{algo:SupSBUCB}) orchestrates stage-wise elimination; BaseSBUCB (Algorithm~\ref{algo:BaseSBUCB}) performs ridge regression on the masked index set $\Psi_m^s$. Within each batch $m$, the sets $\Psi_{m}^s$ are carried forward across time steps as in the fully online construction; the only difference from the sequential setting is that reward information from batch $m$ is revealed only at the end of the batch, so the learner still evaluates UCB scores using only information available under the batching constraint.

\begin{algorithm}[h!]
	\relscale{0.9}%
	\DontPrintSemicolon  
	\SetAlgoLined
	\BlankLine
	\caption{SupSBUCB \label{algo:SupSBUCB}}
     \textbf{Inputs}: $T, M \in \mathbb{Z}_{++}$, Grid $\calT=\{t_1,t_2,\cdots,t_M\}$.
     
     $S\leftarrow \log(T), \Psi_1^s \leftarrow \emptyset$ for all $s\in[S]$
     
	\For{$m=1,2,\cdots,M$}{
		Initialize $\Psi_{m+1}^{s^{\prime}}\leftarrow \Psi_{m}^{s^{\prime}}$ for all $s^{\prime} \in [S]$.
		
	  \For{$t= t_{m-1}+1, \dots, t_m$}{
	    $s \leftarrow 1$ and $\hat{\mathcal{A}}_1 \leftarrow [K]$
		
		\textbf{Repeat:}
	
	    Use BaseSBUCB with $\Psi_{m}^s$ to compute $\theta_m^s$ and $A_m^s$
	    
	    For all $a \in \hat{\mathcal{A}}_s$, compute $w^s_{t,a} =\gamma \sqrt{x_{t,a}^T (A_{m}^s)^{-1}x_{t,a}}$, $\hat{r}_{t,a}^s = \langle \theta_m^s, x_{t,a} \rangle$ 
		
		\textbf{(a)} If $w^s_{t,a}\leq 1/\sqrt{T}$ for all $a\in \hat{\mathcal{A}}_s$,
		choose $a_t = \arg\max_{a\in \hat{\mathcal{A}}_s}\left( \hat{r}_{t,a}^s+w^s_{t,a}\right)$.

		\textbf{(b)} Else if $w^s_{t,a}\leq 2^{-s}$ for all $a \in \hat{\mathcal{A}}_s$,
		$\hat{\mathcal{A}}_{s+1}\leftarrow \{a\in \hat{\mathcal{A}}_s \,\,\vert\,\,\hat{r}_{t,a}^s+w^s_{t,a}\geq \max_{a^{\prime}\in \hat{\mathcal{A}}_s}(\hat{r}_{t,a^{\prime}}^s+w^s_{t,a^{\prime}})-2^{1-s}\}$, 
		
		\quad $s \leftarrow s+1$.
		
			\textbf{(c)} Else choose any $a_t \in \hat{\mathcal{A}}_{s}$ such that $w_{t,a_t}^s >2^{-s}$, \StochRev{update}
	    $\ZH{\Psi_{m+1}^{s} \leftarrow
		\Psi_{m+1}^{s} \cup \{t\}.}$

		\textbf{Until}{\quad an action $a_t$ is found.}
	}
}
\end{algorithm}

\begin{algorithm}[h!]
	\relscale{0.9}%
	\DontPrintSemicolon  
	\SetAlgoLined
	\BlankLine
	\caption{BaseSBUCB \label{algo:BaseSBUCB}}
	
		\textbf{Input}: $\Psi_m$.
	
		\ZH{$A_m = I_d+\sum_{\tau \in \Psi_m}x_{\tau,a_{\tau}}x^{\prime}_{\tau,a_{\tau}}$}
		
		$c_m = \sum_{\tau \in \Psi_m} r_{\tau,a_{\tau}}x_{\tau,a_{\tau}}$
	
		$\theta_m = A_m^{-1} c_m$
		
		\textbf{Return} $(\theta_m, A_m)$.
\end{algorithm}

More specifically, the master algorithm developed by~\cite{auer2002finite} has the following structure: each time step is divided into at most $\log T$ stages. At the beginning of each stage $s$, the learner computes the confidence interval using only the previous contexts designated as belonging to that stage and selects any action whose confidence interval has a large length (exceeding some threshold). If all actions \StochRev{have} a small confidence interval, then we end this stage, observe the rewards of the given contexts and move on to the next stage with a smaller threshold on the length of the confidence interval. In other words, conditional independence is obtained by successive manual masking and revealing of certain information. One can intuitively think of each stage $s$ as a color, and each time step $t$ is colored using one of the $\log T$ colors (if colored at all). When computing confidence intervals and performing regression, only previous contexts that have the same color are used, instead of all previous contexts. 

Adapting this algorithm to the sequential batch setting is not difficult: we merely keep track
of the sets $\Psi_{m}^s$ ($s \in [\log T]$) \StochRev{for each} batch $m$ (rather than \StochRev{for each} time step $t$ as in the fully online learning case). Note that we are still coloring each time step $t$. \StochRev{The difference lies} in the frequency at which we are running BaseSBUCB to compute the confidence bounds and rewards. 
\StochRev{Because of the close similarity,} we omit the details here and refer to \cite[Section 4.3]{auer2002using}. In particular, by establishing similar results to \cite[Lemma 15, Lemma 16]{auer2002using}, it is straightforward to show that the regret of the master algorithm SupSBUCB here is enlarged at most by a multiplicative factor of $O(\log T)$, which leads to the upper bound in Theorem \ref{thm.adversarial}. See Appendix~\ref{appendix.adversarial.upper} for the full derivation.

\ZH{\begin{remark}[Initialization with Empty Sets]
	If $\Psi_m^s$ is empty, the corresponding sums are zero. Hence,
	$c_m=\mathbf{0}$, $A_m^s=I_d$, and $\theta_m^s=\mathbf{0}$.
   \end{remark}}
\ZH{\begin{remark}[Dependence on $K$ under Adversarial Contexts]
	\Rtwo{Our upper bound includes an $O(\sqrt{\log K})$ factor arising from
	the confidence radius and the union bound over the $K$ available actions.
	This explains the dependence produced by our UCB-style analysis, but the
	corresponding adversarial-context lower bound does not contain a matching
	$\Omega(\sqrt{\log K})$ term. We therefore do not claim that this
	$K$-dependence is unavoidable. Whether it can be improved by a different
	adaptive batching method remains open.}
	\end{remark}}

\subsection{Regret Analysis for Lower bound}\label{subsec.adversarial}
In this section, we establish the regret lower bound and show that for any fixed grid $\calT=\{t_1,\cdots,t_M\}$ and any learner's policy on this grid, there exists an adversary who can make the learner's regret at least $\Omega(\sqrt{Td}+(T\sqrt{d}/M \wedge T/\sqrt{M}))$ even if $K=2$. Since the lower bound $\Omega(\sqrt{Td})$ has been proved in \cite{chu2011contextual} even in the fully online case, it remains to show the lower bound $\Omega(T\sqrt{d}/M \wedge T/\sqrt{M})$. Note that in the fully online case, the lower bound $\Omega(\sqrt{Td})$ given in \cite{chu2011contextual} is obtained under the same assumption $d^2 \le T$ as in Assumption~\ref{aspn.TKd}.

\begin{proof}{Proof of Theorem~\ref{thm.adversarial}, Statement~2.}
First we consider the case where $M\ge d/2$, and without loss of generality we may assume that $d'=d/2$ is an integer (if $d$ is odd, then we can take $d^\prime = \frac{d-1}{2}$ and modify the subsequent procedure only slightly). By an averaging argument, there must be $d'$ batches $\{i_1,i_2,\cdots,i_{d'}\}\subset [M]$ such that
\begin{align}\label{eq.large_batch}
\sum_{k=1}^{d'} \left(t_{i_k} - t_{i_k-1} \right) \ge \frac{d'T}{M}. 
\end{align}
\ZH{To establish the minimax lower bound over deterministic parameters, we employ the standard Bayesian reduction technique (see, e.g., \cite{tsybakov2009,lattimore2020bandit}) by assigning a prior over $\theta^\star$. Specifically, $\theta^\star$ is drawn randomly as follows:} Flip $d'$ independent fair coins to obtain $U_1,\cdots,U_{d'}\in \{1,2\}$, and set $\theta^\star = (\theta_1,\cdots,\theta_d)$ with
$\theta_{2k-1} = \frac{1}{\sqrt{d'}}\mathbbm{1}(U_k = 1), \theta_{2k} = \frac{1}{\sqrt{d'}}\mathbbm{1}(U_k = 2), \forall k\in [d']$.
(If $d$ is odd, then the last component $\theta_d$ is set to $0$.)

\ZH{Note that while the true parameter $\theta^\star$ remains fixed in any given instance, this randomized prior guarantees $\|\theta^\star\|_2=1$ almost surely. We can thus lower-bound the worst-case regret by the expected Bayesian regret under this prior.} Next the contexts are generated in the following manner: for $t\in (t_{m-1},t_{m}]$, if $m=i_k$ for some $k\in [d']$, set $x_{t,1}=e_{2k-1}, x_{t,2}=e_{2k}$, where $e_j$ is the $j$-th basis vector in $\bR^d$; otherwise, set $x_{t,1}=x_{t,2}={\bf 0}$. 

Now we analyze the regret of the learner under this environment. Clearly, for any $k\in [d']$, the learner has no information about whether $(\theta_{2k-1}, \theta_{2k}) = (1/\sqrt{d'},0)$ or $(0,1/\sqrt{d'})$ before entering the $i_k$-th batch, while an incorrect action incurs an \StochRev{instantaneous} regret $1/\sqrt{d'}$. Consequently, averaged over all possible coin flips $(U_1,\cdots,U_{d'})\in \{1,2\}^{d'}$, the expected regret is at least:
\begin{align*}
\frac{1}{2}\sum_{k=1}^{d'} \frac{\StochRev{t_{i_k} - t_{i_k-1}}}{\sqrt{d'}} \ge \frac{1}{2\sqrt{2}}\cdot \frac{T\sqrt{d}}{M}
\end{align*}
due to \eqref{eq.large_batch}, establishing the lower bound $\Omega\left(\frac{T\sqrt{d}}{M}\right)$ when $M\ge d/2$.

Next, in the case where $M < d/2$, choose $d^\prime = M$.
Here, since $d'=M$, we have \StochRev{$\sum_{m=1}^{M}(t_m-t_{m-1})=T$}. 
In this case, again flip $d'$ independent fair coins to obtain $U_1,\cdots,U_{d'}\in \{1,2\}$, and set $\theta^\star = (\theta_1,\cdots,\theta_d)$ with
$\theta_{2k-1} = \frac{1}{\sqrt{d'}}\mathbbm{1}(U_k = 1), \theta_{2k} = \frac{1}{\sqrt{d'}}\mathbbm{1}(U_k = 2), \forall k\in [d']$.
Set all remaining components of \StochRev{$\theta^\star$} to $0$.
The contexts are generated as follows: for $t\in (t_{m-1},t_{m}], 1\le m \le M$, set $x_{t,1}=e_{2m-1}, x_{t,2}=e_{2m}$.
In this case, we again average over all possible coin flips $(U_1,\cdots,U_{d'})\in \{1,2\}^{d'}$, and the expected regret is at least:
\begin{align*}
\frac{1}{2}\sum_{m=1}^{M} \frac{t_{m} - t_{m-1}}{\sqrt{d'}} = \frac{1}{2}\cdot \frac{T}{\sqrt{M}}
\end{align*}

Combining the above two cases yields a lower bound of  $ \Omega\left(\frac{T\sqrt{d}}{M}\wedge \frac{T}{\sqrt{M}}\right)$.\halmos
\end{proof}

\section{Learning with Stochastic Contexts}
\label{sec:stochastic}

{
In this section, we focus on the stochastic-context setting. The collection
$\{x_{t,a}:t\in[T],\,a\in[K]\}$ is mutually independent, and
\[
x_{t,a}\sim\mathcal{N}(0,\Sigma)
\]
\StochRev{for every $t\in[T]$ and $a\in[K]$, where the possibly unknown
covariance matrix $\Sigma$ is common to all actions and time steps.}
\StochRev{Although we begin with contexts that are independent across actions
and time and share a common covariance matrix, these restrictions can be
partially relaxed. In particular, Proposition~\ref{prop.heterogeneous_context}
allows correlations across actions and action-dependent covariance matrices,
while retaining independence across time.}
This is a simple setting that presents an interesting case for study:
at a population level, each of the $K$ actions is equally good. In
particular, if the decision maker is not allowed to personalize the action
based on the context and is therefore restricted to a single-action policy,
then all actions perform equally well. However, as we shall see, selecting
different actions based on the realized contexts allows the decision maker
to do much more. We next impose a condition on the covariance matrix.
}

 \begin{assumption}\label{assumption:cov}
 	The covariance matrix $\Sigma$ satisfies
 	$
 	\frac{\kappa}{d} \le \lambda_{\min}(\Sigma) \le \lambda_{\max}(\Sigma) \le \frac{1}{d}
 	$
 	for some numerical constant $\kappa>0$, where $\lambda_{\min}(\Sigma), \lambda_{\max}(\Sigma)$ denote the smallest and the largest eigenvalues of $\Sigma$, respectively. 
 \end{assumption}
 \ZH{
	\begin{remark}
		\Rtwo{The upper bound $\lambda_{\max}(\Sigma)\leq 1/d$ fixes the scale
		of the contexts and is without loss of generality up to rescaling. Given
		a covariance matrix $\Sigma_0$, the transformation
		$x_{t,a}\mapsto x_{t,a}/\sqrt{d\lambda_{\max}(\Sigma_0)}$, together
		with the inverse rescaling of $\theta^\star$, preserves the mean
		rewards and yields the stated upper bound. The substantive condition is
		the bounded condition number
		$\lambda_{\max}(\Sigma)/\lambda_{\min}(\Sigma)\leq 1/\kappa$.
		In practice, a related normalization can be obtained through principal
		component analysis (PCA) followed by whitening.}
	\end{remark}
 }
\Rtwo{The upper bound $\lambda_{\max}(\Sigma)\leq 1/d$ also gives
$\bE\|x_{t,a}\|_2^2=\operatorname{tr}(\Sigma)\leq 1$, matching the scale
constraint imposed on the adversarial contexts. The lower bound
$\lambda_{\min}(\Sigma)\geq\kappa/d$ ensures nondegenerate variation in every
direction. We treat $\kappa>0$ as a fixed constant and do not optimize its
dependence.}
 
The next theorem presents tight regret bounds for the stochastic contexts case.

\begin{theorem}\label{thm.stochastic}
	\Rtwo{Suppose $M=O(\log\log T)$.}
	\begin{enumerate}
	\item 
	{Under the stochastic-context model specified above and
	Assumptions~\ref{aspn.TKd} and~\ref{assumption:cov}, there exists a
	sequential batch learning algorithm $\textbf{Alg}=(\calT,\pi)$
	(explicitly defined in Section~\ref{subsec.pure-exp}) such that:}
	\begin{align*}
	\sup_{\theta^\star:\|\theta^\star\|_2\le 1}
	\mathbb{E}\!\left[
	R_T\!\left(\textbf{Alg},\theta^\star\right)
	\right]
	\le
	\widetilde{O}\!\left(
	\sqrt{\frac{dT\log K}{\kappa}}
	\left(\frac{T}{d^2}\right)^{\frac{1}{2(2^M-1)}}
	\right).
	\end{align*}
	\item 
	\ZH{{Conversely, even when contexts $x_{t,a} \sim \mathcal{N}(0, I_d/d)$ over all $a \in [K]$ and $t \in [T]$, assuming $T \ge d^2$, $d \ge 2$, and $K \ge 2$, for any sequential batch learning algorithm $\textbf{Alg}$ restricted to at most $M$ batches, we have:}
$$ \sup_{\theta^\star: \|\theta^\star\|_2 \le 1} \mathbb{E}[R_T(\textbf{Alg}, \theta^\star)] \ge c \cdot \sqrt{dT \log K} \left( \frac{T}{d^2} \right)^{\frac{1}{2(2^M - 1)}} $$
where $c > 0$ is a universal numerical constant independent of $(T, M, d, K)$.}
\end{enumerate}
\end{theorem}

{Theorem~\ref{thm.stochastic} completely characterizes the minimax regret for
sequential batch learning under the independent Gaussian context model
specified above and shows a doubly exponential dependence of the optimal
regret on the number of batches $M$. The following corollary is immediate.}
\begin{corollary}\label{cor.stochastic}
	{Under the independent Gaussian context model, it is necessary and
	sufficient to have $\Theta(\log\log(T/d^2))$ batches to achieve the
	fully online regret $\widetilde{\Theta}(\sqrt{dT})$.}
\end{corollary}

Compared with Corollary~\ref{cor.adversarial},
Corollary~\ref{cor.stochastic} shows that fully online performance is
attainable with far fewer batches. Theorem~\ref{thm.stochastic} also gives
tight regret rates up to logarithmic factors when fewer batches are
available.


\subsection{A Sequential Batch Pure-Exploitation Algorithm}\label{subsec.pure-exp}

In contrast to the adversarial contexts, under stochastic contexts the decision maker enjoys the advantage that he can choose to learn the unknown parameter $\theta^\star$ from any desired direction. In other words, the exploration of the learner is no longer subject to the adversary's restrictions, and strikingly, making decisions based on the best possible inference of $\theta^\star$ is already sufficient.

\begin{algorithm}[h!]
	\relscale{0.9}%
	\DontPrintSemicolon  
	\SetAlgoLined
	\BlankLine
	\caption{Sequential Batch Pure Exploitation (SBPE)	\label{algo.pure-exp}}
	\textbf{Input:} Time horizon $T$; context dimension $d$; number of batches $M$. \\
	\textbf{Set} $\beta = \Theta\left( \sqrt{T}\cdot \left(\frac{T}{d^2}\right)^{\frac{1}{2(2^M-1)}} \right)$\\
	\textbf{Grid choice}: $\calT = \{t_1,\cdots,t_M\}$, with $t_1 = \beta d, \quad t_m = \lfloor \beta\sqrt{t_{m-1}} \rfloor, \StochRev{m=2,3,\cdots,M.}$\\
	\textbf{Initialization:} $A = {\bf 0}\in \bR^{d\times d}$, \Rtwo{$b={\bf 0}\in\bR^d$}, $\hat{\theta}={\bf 0}\in \bR^d$\;
	\For{$m \gets 1$ \KwTo $M$}{
			\For{$t\gets t_{m-1}+1$ \KwTo $t_m$}{
				choose $a_t = \arg\max_{a\in [K]} x_{t,a}^\top \hat{\theta}$ {(break ties according to a fixed rule independent of the current contexts)}. \\
				receive reward $r_{t,a_t}$. 
			}
		$A\gets A + \sum_{t=t_{m-1}+1}^{t_m} x_{t,a_t}x_{t,a_t}^\top$. \\
		\Rtwo{$b\gets b+\sum_{t=t_{m-1}+1}^{t_m}r_{t,a_t}x_{t,a_t}$.} \\
		\Rtwo{$\hat{\theta}\gets A^{-1}b$.}
	}
\end{algorithm}

Algorithm~\ref{algo.pure-exp} selects actions greedily using the
least-squares estimate computed from previous batches and updates this
estimate at the end of each batch.

How do we select the grid $\calT$? Intuitively, in order to minimize overall regret, we must ensure that the regret incurred on each batch is not too large, because the overall regret is dominated by the batch that has the largest regret. Guided by this observation, we can see intuitively an optimal way of selecting the grid must ensure that each batch's regret is the same (at least orderwise in terms of the dependence of $T$ and $d$): for otherwise, there is a way of reducing the regret order in one batch and increasing the regret order in the other and the sum of the two will still have smaller regret order than before (which is dominated by the batch that has larger regret order). As we shall see later, the following grid choice satisfies this equal-regret-across-batches requirement:
\begin{align}\label{eq.minimax_grid}
t_1 = \beta d, \quad t_m = \lfloor \beta\sqrt{t_{m-1}} \rfloor, \qquad m=2,3,\cdots,M,
\end{align}
where the parameter $\beta = \Theta\left( \sqrt{T}\cdot \left(\frac{T}{d^2}\right)^{\frac{1}{2(2^M-1)}} \right)$ 
is chosen so that $t_M=T$. 

\ZH{The preceding grid is not merely a schedule for updating $\hat{\theta}$: within each batch the policy is purely greedy, so the covariates actually observed along the played actions are \emph{selected} using the current estimate and need not resemble i.i.d.\ draws from any fixed arm. This selection bias is precisely what makes greedy rules subtle in adaptive learning. The grid \eqref{eq.minimax_grid}, with $\beta$ chosen so that $t_M=T$, is calibrated so that, once such biased samples still accumulate sufficient directional information, the per-batch regret contributions are balanced in order across $m\in[M]$; otherwise the total regret would be dominated by a single batch. Thus the doubly exponential-type growth of the batch endpoints should be read jointly with the subsequent geometric argument establishing that greedy exploitation does not destroy estimation along hidden directions.}

\subsection{Regret Analysis for Upper bound}\label{subsec.stochastic_upperbound}
{The upper bound in Statement~1 of Theorem~\ref{thm.stochastic} uses two
distinct independence properties. First, mutual independence of the context
vectors across actions and time ensures that greedy selection does not
eliminate Gaussian variation orthogonal to the selection direction. This
property yields the lower bound on the minimum eigenvalue of the
selected-context design matrix in Lemma~\ref{lemma.equator}. Second, the
estimation argument in Lemma~\ref{lemma.difference} requires the reward noises
to be conditionally independent given the selected contexts. Because pure
exploitation can couple batches through past rewards, we obtain this second
form of independence through sample splitting. Specifically, the batched
pure-exploitation algorithm with sample splitting
(Algorithm~\ref{algo.sample_splitting}, BPE) uses disjoint subsamples
$T^{(m)}$ for estimation and satisfies Lemma~\ref{lemma.cond_indep}, at the
cost of an $O(M)=O(\log\log T)$ reduction in effective sample size. The
complete proof is given in Appendix~\ref{appendix.stochastic.upper}, and the
proofs of Lemmas~\ref{lemma.equator} and~\ref{lemma.difference} appear in
Section~\ref{sec:proof-main-lemmas}.}

\begin{algorithm}[h!]
	\relscale{0.9}%
	\DontPrintSemicolon  
	\SetAlgoLined
	\BlankLine
	\caption{Batched Pure Exploitation (BPE, with sample splitting)	\label{algo.sample_splitting}}
	\textbf{Input:} Time horizon $T$; context dimension $d$; number of batches $M$; grid $\calT = \{t_1,\cdots,t_M\}$\StochRev{, the same as} in Algorithm~\ref{algo.pure-exp}.\;
	\textbf{Initialization:} Partition each batch into $M$ intervals evenly, i.e.,
	{$(t_{m-1},t_m]=\bigcup_{j=1}^M T_m^{(j)}$ for every $m\in[M]$}. \;
	\For{$m \gets 1$ \KwTo $M$}{
		\If{$m=1$}{
			choose $a_t = 1$ and \Rtwo{receive} reward $r_{t,a_t}$ for any $t\in [1,t_1]$.
		}
		\Else{
			\For{$t\gets t_{m-1}+1$ \KwTo $t_m$}{
				choose $a_t = \arg\max_{a\in [K]} x_{t,a}^\top \hat{\theta}_{m-1}$ {(break ties according to a fixed rule independent of the current contexts)}. \\
				receive reward $r_{t,a_t}$. 
			}
		}
		$T^{(m)} \gets \cup_{m'=1}^m T_{m'}^{(m)}.$\\
		$A_m\gets \sum_{t\in T^{(m)}} x_{t,a_t}x_{t,a_t}^\top$. \\
		$\hat{\theta}_m \gets A_m^{-1}\sum_{t\in T^{(m)}} r_{t,a_t}x_{t,a_t}$.
	}
	\textbf{Output: resulting policy $\pi=(a_1,\cdots,a_T)$}.
\end{algorithm}
\ZH{
\subsection{Lower Bound}\label{subsec.stochastic_lower}
The core proof idea leverages a Bayesian framework by introducing an adversarial Gaussian prior over the true parameter to lower-bound the worst-case regret. The fundamental mechanism connects the algorithm's instantaneous regret to its statistical uncertainty, specifically the unestimated orthogonal variance of the parameter. Because the algorithm can only update its policy at discrete batch boundaries, its posterior precision matrix remains frozen throughout each batch, forcing it to incur unavoidable regret based on outdated information. By analyzing the cumulative regret across these frozen periods and mathematically optimizing the worst-case batch boundaries---either through a cascading telescoping sequence or the AM-GM inequality---the proofs definitively demonstrate that any batch allocation strategy must suffer a structurally bounded minimum regret. The proof details are given in Appendix~\ref{appendix.stochastic.lower}.

}
\ZH{
\subsection{Extensions and Generalizations}
\label{subsec:extension}
In this \Rtwo{section}, we provide several possible extensions \StochRev{of our problem setting} to show the versatility of the analysis framework. The upper bound in Theorem \ref{thm.stochastic} can be naturally generalized beyond the basic shared Gaussian setting. In the following, we extend our results to accommodate dependent and heterogeneous contexts, general sub-Gaussian contexts with anti-concentration, and action-dependent parameters (the disjoint setting). The proofs of these extensions are deferred to Section \ref{appendix.proof-prop}.


First, we relax the requirement that the contexts $x_{t,a}$ are independent across actions and share the same covariance matrix.

{
\begin{assumption}\label{assumption:heterogeneous_cov}
Let $x_t=(x_{t,1},\dots,x_{t,K})\in \bR^{dK}$ be the concatenated context vector. For every $t\in [T]$ we assume that $x_t\sim \calN(0,\Sigma)$, with $\Sigma\in \bR^{dK \times dK}$ and
\begin{align*}
\frac{\kappa}{d} \le \lambda_{\min}(\Sigma) \le \lambda_{\max}(\Sigma) \le \frac{1}{d}.
\end{align*}
Moreover, the contexts $\{x_t: t\in [T]\}$ are mutually independent over time. 
\end{assumption}

Compared with Assumption \ref{assumption:cov}, Assumption \ref{assumption:heterogeneous_cov} allows for heterogeneous context \StochRev{distributions} $x_{t,a}\sim \calN(0, \Sigma_a)$ with action-dependent covariance $\Sigma_a$, as well as correlations between $x_{t,a}$ across different actions $a$. The next result summarizes the regret guarantee achieved by \StochRev{the same} sequential batch learning algorithm, with an extra $\sqrt{\log K}$ factor due to the superconcentration of $\max_{i\le K} Z_i$ with i.i.d. $Z_i\sim \calN(0,1)$, i.e. $\var(\max_{i\le K} Z_i) \asymp \frac{1}{\log K}$. 
}

\begin{proposition}\label{prop.heterogeneous_context} 
\Rtwo{Suppose $M=\mathcal{O}(\log\log T)$.} {Under Assumptions \ref{aspn.TKd} and \ref{assumption:heterogeneous_cov}, the sequential batch learning algorithm \textbf{Alg} $=(\mathcal{T},\pi)$ achieves the following expected regret bound:}
$$ \sup_{\theta^\star: \|\theta^\star\|_2\le 1} \mathbb{E}[R_T\left(\textbf{Alg}, \theta^\star\right)] \le \mathsf{polylog}(T)\cdot\sqrt{\log K}\cdot \sqrt{\frac{dT\log K}{\kappa}}\left(\frac{T}{d^2}\right)^{\frac{1}{2(2^M-1)}}. $$ 
\end{proposition}

We now consider the disjoint setting, where the reward for each action $a\in[K]$ is governed by an unknown, action-specific parameter $\theta_a^\star\in\mathbb{R}^d$ such that $r_{t,a} = x_{t,a}^\top\theta_a^\star$. To handle this, the learner maintains $K$ separate sample covariance matrices and estimators. For each action $a\in[K]$ in the $m$-th batch, define:
$$ A_{m,a} = \sum_{t=1}^{t_m} \mathbbm{1}(a_t=a)x_{t,a}x_{t,a}^\top, \quad \text{and} \quad \hat{\theta}_{m,a} = A_{m,a}^{-1} \sum_{t=1}^{t_m} \mathbbm{1}(a_t=a)r_{t,a}x_{t,a}. $$

The greedy action selection rule at time $t$ naturally adapts to:
$$ a_t = \arg\max_{a\in[K]} x_{t,a}^\top \hat{\theta}_{m-1,a}. $$

		
		\begin{proposition}[Action-Dependent Parameters]\label{prop.action_dependent} 
			\Rtwo{Suppose $M=\mathcal{O}(\log\log T)$.}
			{Under Assumptions~\ref{aspn.TKd}
			and~\ref{assumption:heterogeneous_cov}, the modified sequential
			batch learning algorithm $\textbf{Alg}=(\mathcal{T},\pi)$ using
			the disjoint estimators achieves the following expected regret bound:}
			$$ \sup_{\{\theta^\star_a\}_{a=1}^K: \|\theta^\star_a\|_2\le 1} \mathbb{E}[R_T\left(\textbf{Alg}, \{\theta^\star_a\}_{a=1}^K\right)] \le \mathsf{polylog}(T)\cdot \sqrt{\frac{dT}{\kappa}}\left(\frac{T}{d^2}\right)^{\frac{1}{2(2^M-1)}},
			$$ 
            where the hidden constant depends on $K$.
			\end{proposition}

We note that the hidden constant in Proposition \ref{prop.action_dependent} depends exponentially on $K$, and such dependence is unavoidable for the greedy selection rule due to the action-specific parameter $\theta_a^\star\in \bR^d$. For instance, if $\hat{\theta}_{m-1,a}=0$ for some action \StochRev{$a\in [K]$}, then the greedy rule selects this action only if $x_{t,a'}^\top \hat{\theta}_{m-1,a'} \le 0$ for all $a'\neq a$, which occurs with an exponentially small probability $2^{1-K}$. 

We can also relax the Gaussian assumption entirely, provided the context distributions satisfy standard sub-Gaussian and anti-concentration properties.

\begin{assumption}[General Distributional Conditions]\label{assumption:general_distribution}
	{The collection $\{x_{t,a}:t\in[T],\,a\in[K]\}$ is mutually
	independent. Each context $x_{t,a}$ has a symmetric distribution
	satisfying the following two conditions:}
	\begin{enumerate} 
		\item \textbf{Sub-Gaussianity and Well-Conditioned Covariance:} The contexts are $\sigma_x^2$-sub-Gaussian random vectors, {where $\sigma_x^2\le C_x/d$ for some numerical constant $C_x>0$ independent of $(K,T,d)$,} with bounded norms ($\|x_{t,a}\|_2 \le 1$ almost surely) and a covariance matrix $\Sigma$ satisfying $\frac{\kappa}{d} \le \lambda_{\min}(\Sigma) \le \lambda_{\max}(\Sigma) \le \frac{1}{d}$.
		\item \textbf{Conditional Diversity (Orthogonal Decoupling):} There exists a universal constant $c_1 > 0$ such that for any fixed unit vector $v \in \mathbb{S}^{d-1}$, if we orthogonally decompose the context as $x_{t,a} = (x_{t,a}^\top v) v + w_{t,a}$ where $w_{t,a} \perp v$, then conditioned on the parallel projection $x_{t,a}^\top v$, the orthogonal component $w_{t,a}$ has zero conditional mean and its conditional covariance is lower bounded:
		$$ \mathbb{E}\left[w_{t,a} \mid x_{t,a}^\top v\right] = \mathbf{0}, \quad \text{and} \quad \mathbb{E}\left[ w_{t,a} w_{t,a}^\top \;\middle|\; x_{t,a}^\top v \right] \succeq \frac{c_1}{d} (I_d - v v^\top) \quad \text{almost surely.} $$
	\end{enumerate} 
	\end{assumption}
		{Assumption~\ref{assumption:general_distribution} encompasses mutually
		independent context vectors with suitable bounded, rotationally
		symmetric marginal distributions, as well as suitable bounded marginal
		distributions with independent symmetric coordinates that satisfy the
		stated conditional diversity condition. Independence across context
		vectors and conditional diversity within each vector jointly control
		the selection bias introduced by pure exploitation.}
	
	\begin{proposition}[General Distributional Bounds]\label{prop.general_distribution} 
	\Rtwo{Suppose $M=\mathcal{O}(\log\log T)$.} Under Assumption \ref{assumption:general_distribution}, the sequential batch learning algorithm \textbf{Alg} $=(\mathcal{T},\pi)$ achieves the following expected regret bound:
	$$ \sup_{\theta^\star: \|\theta^\star\|_2\le 1} \mathbb{E}[R_T\left(\textbf{Alg}, \theta^\star\right)] \le \mathcal{O}(\sqrt{\log K}) \cdot \mathsf{polylog}(T) \cdot \sqrt{\frac{dT}{\kappa}}\left(\frac{T}{d^2}\right)^{\frac{1}{2(2^M-1)}}. $$ 
	The expected regret exhibits an optimal $\mathcal{O}(\sqrt{\log K})$ dependence on the number of arms.
	\end{proposition}
	In follow-up work, \citet{ren2024dynamic} focus on the high-dimensional sparse regime and allow general context distributions. They also allow arbitrary cross-arm dependence among $\{x_{t,a}\}_{a=1}^{K}$ within each time step.
}

\section{Problem-Dependent Regret Bounds}\label{sec:gap}
While the preceding sections established problem-independent bounds, stochastic contexts with well-separated arms (e.g., $\|\theta^\star\|_2 > 0$) admit sharper problem-dependent guarantees that improve upon the $\Theta(\sqrt{dT})$ worst-case rate. The following theorem characterizes this for $K$ arms (introducing a $\sqrt{\log K}$ factor); its proof is deferred to Appendix~\ref{appendix.problem-dependent}.
\ZH{
\begin{theorem}\label{thm.problem-dependent}
	\Rtwo{Suppose $M=O(\log T)$.} Assume without loss of generality $\|\theta^\star\|_2 > 0$.
	\begin{enumerate}
		\item 
		{Under Assumption~\ref{aspn.TKd}, suppose that the context
		vectors are mutually independent across actions and time, with
		$x_{t,a}\sim\mathcal{N}(0,\Sigma)$, where $\Sigma$ satisfies
		Assumption~\ref{assumption:cov}. Then there exists a sequential batch
		learning algorithm $\textbf{Alg}=(\calT,\pi)$ (explicitly defined
		below) that achieves the following regret:}
		\begin{align*}
	   \mathbb{E}[R_T\left(\textbf{Alg}, \theta^\star\right)] \le \mathsf{polylog}(T)\cdot\sqrt{\log K}\cdot \frac{d^{3/2}}{\sqrt{\kappa}\|\theta^\star\|_2} \left(\frac{T}{d^2}\right)^{\frac{1}{M}}.
		\end{align*}
		\item {Conversely, suppose the contexts $x_{t,a} \sim \mathcal{N}(0, I_d/d)$ are mutually independent over all $a \in [K]$ and $t \in [T]$. Let $d \ge 2$, $K \ge 2$, and the horizon $T$ satisfy $T/d^2 \ge 2^M$. For any sequential batch learning algorithm $\textbf{Alg}$ restricted to at most $M$ batches, we have:}
\[
\sup_{\theta^\star: \|\theta^\star\|_2 \le 1} \|\theta^\star\|_2 \cdot \mathbb{E}[R_T(\textbf{Alg}, \theta^\star)] \ge c \cdot \sqrt{\log K} \cdot d^{3/2} \left( \frac{T}{d^2} \right)^{\frac{1}{M}},
\]
where $c > 0$ is a universal numerical constant independent of $(T, M, d, K)$.
	\end{enumerate}
\end{theorem}
}
\begin{corollary}
In this setting, it is necessary and sufficient to have $\Theta(\log(T/d^2))$ batches to achieve the optimal problem-dependent regret $\tilde{\Theta}(d^{3/2} / \|\theta^\star\|_2)$. 
Here we are not aiming to get the tightest dependence on $\log T$ (note that $\tilde{\Theta}(\cdot)$ hides polylog factors). 
\end{corollary}
\ZH{

This improved bound is achieved by Algorithm~\ref{algo.sample_splitting} (batched pure-exploitation with sample splitting) using a geometric grid $\mathcal{T}' = \{t_1', \dots, t_M'\}$, defined by $t_1' = bd^2$ and $t_m' = \lfloor b t_{m-1}' \rfloor$ for $m \ge 2$, where $b = \Theta((T/d^2)^{1/M})$ ensures $t_M' = T$.

This architecture ensures two critical properties:
\begin{enumerate}
    \item \textbf{Sequential independence:} Sample splitting trains $\hat{\theta}_{m-1}$ exclusively on independent subsets, which prevents design matrix skew and enables sharp unregularized OLS estimates.
    \item \textbf{Fractional control:} The geometric grid maintains a uniform fractional batch length $\Delta t_m / t_{m-1} = O(b)$. Since expected instantaneous regret in batch $m$ scales as $O(1/t_{m-1})$, each batch contributes $O(b)$ regret. Summing these $M$ bounds, alongside the initial $t_1'$ exploration phase, directly yields the $\mathcal{O}((T/d^2)^{1/M})$ overall scaling.
\end{enumerate}
}


\ZH{
\section{Experimental Results}
\label{sec:experimental_results}

\subsection{Experimental Setup: Adversarial Contexts}

To evaluate the empirical performance of our proposed batched contextual bandit algorithms, we conduct numerical experiments on two real-world classification datasets from the UCI Machine Learning Repository, transforming them into contextual bandit tasks. We compare our proposed batch-constrained algorithms SBUCB and SupSBUCB against a fully sequential baseline LinUCB \citep{chu2011contextual}.

We formulate the multi-class classification problem as a contextual bandit problem where the learner only observes the binary reward $r_t \in \{0, 1\}$ for the chosen action (class prediction) $a_t$ at time $t$. To map the classification data into the linear contextual bandit framework, let $v_t \in \bR^p$ denote the processed feature vector for the $t$-th sample. For each action $a \in [K]$, we construct the $d$-dimensional context-action feature vector $x_{t,a} \in \bR^d$ (where $d = p \times K$) by mapping $v_t$ into the $a$-th block of a zero vector:
$$x_{t,a} = (\mathbf{0}^\top, \dots, \mathbf{0}^\top, \underbrace{v_t^\top}_{a\text{-th block}}, \mathbf{0}^\top, \dots, \mathbf{0}^\top)^\top.$$
This disjoint feature design allows the model parameter $\theta^\star \in \bR^d$ to learn class-specific weights. The chosen action's feature is thus denoted as $x_{t,a_t}$. To align with the adversarial context setting, we do not permute the data entries.

The Mushroom dataset \citep{mushroom_73} contains $T_{total} = 8,124$ samples and $K = 2$ classes (edible vs. poisonous). The $22$ original categorical features are transformed using one-hot encoding, resulting in a raw feature dimension $p=117$, and an effective dimension $d = p \times 2=234$. The Covertype dataset contains $T_{total} = 581,012$ samples and $K = 7$ classes (forest cover types). The dataset contains $p = 54$ raw features, yielding an effective dimension $d = 54 \times 7 = 378$.

\begin{figure}[t]
    \centering
    \begin{subfigure}[t]{0.47\textwidth}
        \centering
        \includegraphics[width=\textwidth]{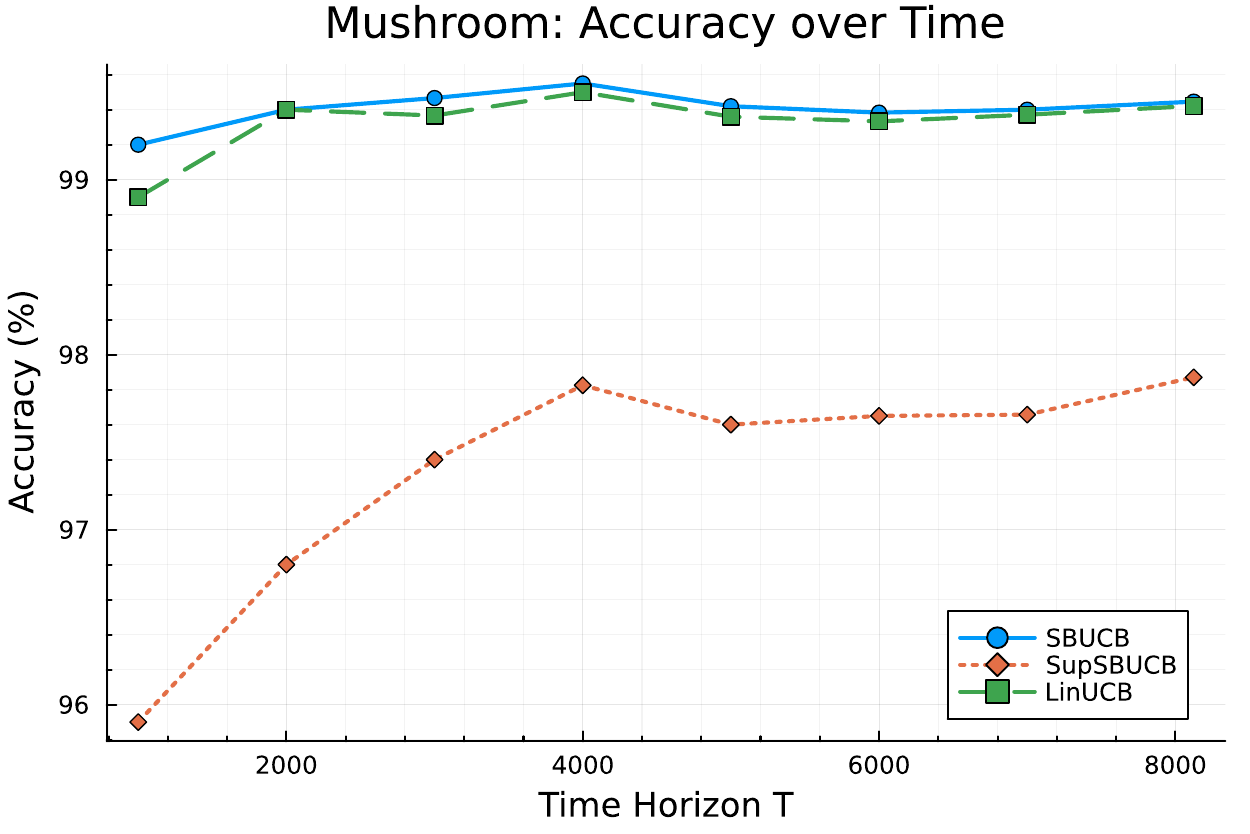}
        \caption{Mushroom}
        \label{fig:mushroom_results}
    \end{subfigure}\hfill
    \begin{subfigure}[t]{0.47\textwidth}
        \centering
        \includegraphics[width=\textwidth]{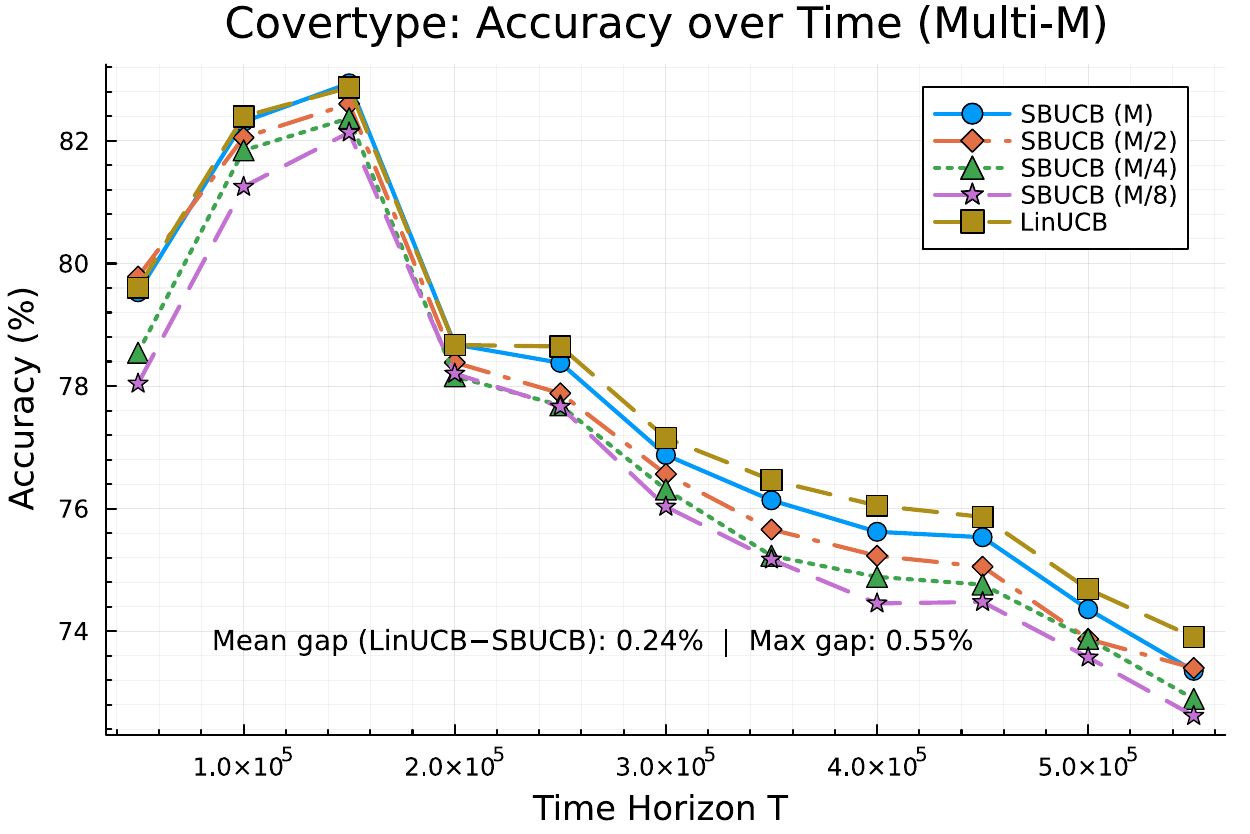}
        \caption{Covertype}
        \label{fig:covertype_results}
    \end{subfigure}

    \caption{Average cumulative reward (accuracy) over the time horizon $T$.}
    \label{fig:datasets_results}
\end{figure}

For the batched algorithms (SBUCB and SupSBUCB), the number of batches $M$ is dynamically determined based on the time horizon $T$ and the effective dimension $d$. Specifically, we set $M = \max\left(1, \left\lfloor \sqrt{\frac{d \log(1 + T/d) T}{\log(2)}} \right\rfloor\right)$, with the grid boundaries defined as $t_m = \lfloor m T / M \rfloor$. The exploration tuning parameter is set to $\gamma = 6.0$ across all algorithms. \Rtwo{At horizon $T$, we measure classification accuracy as
\[
\operatorname{Accuracy}(T)=\frac{1}{T}\sum_{t=1}^T
\mathbf{1}\{a_t=y_t\}=\frac{1}{T}\sum_{t=1}^T r_t,
\]
where $y_t$ is the true class label. We report this quantity averaged over
$50$ independent trials. Because the oracle predicts the correct class and
receives reward one in every round, $R_T/T=1-\operatorname{Accuracy}(T)$ in
this construction. Accuracy is therefore an interpretable classification
metric equivalent to normalized regret. We use it for the real-data
classification experiments in this subsection, whereas Section~6.2 reports
cumulative regret to examine its theoretical scaling with $T$.} Results are
evaluated at various horizons up to $T = T_{total}$. We only include SupSBUCB in the comparison on the Mushroom dataset, since SupSBUCB is \StochRev{15--20 times} slower than SBUCB, and the Covertype dataset is too large for SupSBUCB to run in a reasonable time.

As shown in Figures \ref{fig:mushroom_results} and \ref{fig:covertype_results}, SBUCB achieves performance nearly identical to the LinUCB baseline, while SupSBUCB exhibits a noticeable drop in accuracy. Although SupSBUCB resolves conditional independence issues to secure minimax optimal bounds, the conservative data-splitting mechanics required for such theoretical rigor incur a clear empirical cost. Conversely, SBUCB serves as a practical and efficient alternative that maintains parity with fully sequential performance.

Notably, Figure \ref{fig:covertype_results} reveals a distinct trend where the classification accuracy initially increases but subsequently declines as the time horizon $T$ expands. We attribute this non-monotonic behavior directly to our experimental design choice of not permuting the data entries. Because the original Covertype dataset exhibits inherent structural grouping, streaming the \StochRev{unshuffled} data introduces non-stationarity, effectively simulating an adversarial context setting. Consequently, the algorithms must continuously adapt to shifting feature and class distributions, which naturally drags down the cumulative average accuracy over time compared to a randomized, i.i.d. setting.

To further investigate the sensitivity of our approach to batch frequency, we evaluated SBUCB on the Covertype dataset using reduced batch quantities of $M/2$, $M/4$, and $M/8$. While a lower batch count leads to a mild degradation in average reward, the classification accuracy remains remarkably competitive even at $M/8$. This highlights the algorithm's robustness to coarse updating schedules; for instance, while LinUCB requires $T = 581,012$ sequential updates, SBUCB achieves comparable results with only $M = 4,821$ batches—a 120-fold reduction in update frequency. These findings confirm that making decisions based on batched historical inferences is sufficient for effective learning in practical environments.

\subsection{Experimental Setting: Stochastic Contexts}
We evaluate SBPE and BPE against two exploration-free baselines across four
stochastic environments: Lasso Batch Greedy Learning (LBGL)
\citep{ren2024dynamic} and Ridge Batch Greedy Learning (RBGL)
\citep{bastani2017mostly}. LBGL uses Lasso for high-dimensional sparse
features, whereas RBGL uses ridge regression.

{To illustrate our stochastic-context results and assess robustness
beyond the exact Gaussian model,} we set the context dimension $d = 10$ and the number of actions $K = 10$. The time horizon $T$ is evaluated from $50,000$ to $100,000$ in increments of $5,000$. The results for each horizon are averaged over 50 independent trials.
\begin{figure}
    \centering
    \includegraphics[width=0.90\textwidth]{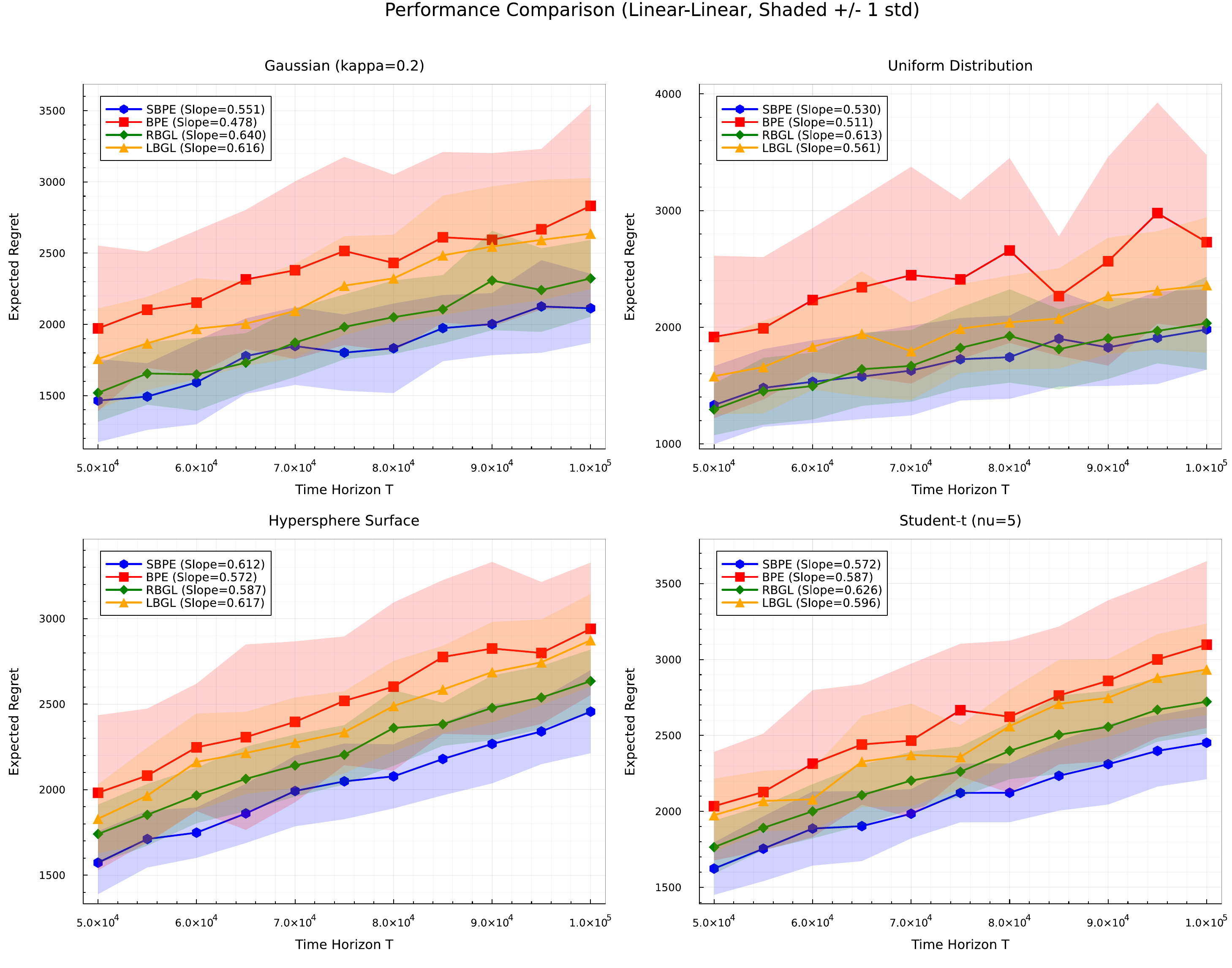}
    \caption{Plot of cumulative expected regret vs. time horizon. The annotated slopes represent the asymptotic scaling calculated from a log-log regression.}
    \label{fig:stochastic_regret}
\end{figure}
{In these synthetic experiments, contexts are generated independently
across actions and time.} The contexts $x_{t,a}$ are drawn from four distributions to stress robustness to different feature geometries: \emph{Gaussian} ($\kappa=0.2$), with a multivariate normal whose covariance has eigenvalues uniformly spaced in $[\kappa/d, 1/d]$ and a random orthogonal rotation, so $\lambda_{\min}=\kappa/d$; \emph{uniform} on $[-1,1)^d$, scaled by $1/\sqrt{d}$ for comparable expected $\ell_2$ norm; \emph{hypersphere} draws from a standard Gaussian followed by projection onto the unit sphere; and \StochRev{\emph{Student's $t$}} ($\nu=5$) heavy-tailed multivariate $t$ noise.
For the \StochRev{stochastic-context setting}, the number of batches required by the four algorithms we consider has a doubly-logarithmic dependence on the time horizon. Specifically, the total number of batches for all four algorithms is dynamically set to $M = \max(2, \lceil 0.2 \log_2(\log_2(T/d^2)) \rceil)$. Given $M$, the batch grid boundaries $t_1, t_2, \dots, t_M$ are constructed recursively as follows:
\begin{itemize}
    \item \textbf{SBPE and BPE:} Both algorithms employ a recursive grid where the first batch ends at $t_1 = \lfloor \rho d \rfloor$, and subsequent batches end at $t_m = \lfloor \rho \sqrt{t_{m-1}} \rfloor$ for $1 < m < M$. The multiplier is defined as $\rho = a_{\text{scale}} \sqrt{T} (T/d^2)^{\frac{1}{2(2^M-1)}}$.     We set the scaling constant to $a_{\text{scale}} = 0.4$ for SBPE and BPE.

    \item \textbf{RBGL and LBGL:} The baseline algorithms utilize a similar recursive structure $t_m = \lfloor b_{\text{val}} \sqrt{t_{m-1}} \rfloor$, initialized at $t_1 = \lfloor b_{\text{val}} \sqrt{d} \rfloor$. The scaling multiplier is defined as $b_{\text{val}} = \sqrt{T} (T/d)^{\frac{1}{2(2^M-1)}}$. For LBGL, the theoretical sparsity parameter $s_0$ is conservatively set to $d$ (as the optimal parameter is dense in this setting), making its batch schedule mathematically identical to that of RBGL.
\end{itemize}

To ensure numerical stability and valid batch intervals in practice, all schedules strictly enforce $t_m \ge t_{m-1} + 1$, and the final batch boundary is strictly anchored to the horizon, $t_M = T$. The primary evaluation metric is the expected cumulative regret, with shaded regions representing $\pm 1$ standard deviation to assess algorithm stability.



From Figure \ref{fig:stochastic_regret}, we can see that SBPE demonstrates strong empirical performance in all four environments. While the regret of BPE has a similar slope to SBPE in all four environments, the regret of BPE is larger than that of SBPE. We note that BPE is a theoretically optimal algorithm, while SBPE is a practical algorithm that is easy to implement and performs well in practice. This is similar to the case of adversarial contexts, where theoretically optimal SupSBUCB exhibits inferior performance compared to SBUCB.
}

\ZH{
	\subsection{Experimental Setting: Problem-Dependent Regret}

We conduct experiments on the Avazu Click-Through Rate (CTR) prediction dataset, which aims to predict whether a customer will click on an ad. The Avazu dataset contains approximately $40.4$ million entries, and there are $22$ raw features for each entry. The categorical strings are transformed into numerical vectors using the hashing trick, mapped to a dimension of $d=40$, and normalized by $1/\sqrt{d}$. We treat this massive preprocessed dataset as a global pool of available ad features.

To simulate the ad recommendation environment, at each time step $t \in [T]$, the environment samples $K=2$ distinct ad entries from the Avazu feature pool uniformly at random. The $d$-dimensional feature vector of the $a$-th sampled ad serves directly as the context $x_{t,a} \in \mathbb{R}^d$ for arm $a$. Unlike the classification reduction setting where features are zero-padded into disjoint blocks, this recommendation setting utilizes a shared linear model. That is, all arms share a single, global underlying parameter $\theta^\star \in \mathbb{R}^d$ that represents the general user preference.

Because the raw Avazu dataset and the binary ad click outcomes might not strictly follow the linear model assumption, we adopt a semi-synthetic reward setting. The optimal $\theta^\star$ is computed by running a linear regression on the dataset, and the synthetic reward for choosing arm $a$ is generated as $r_{t,a} = x_{t,a}^\top\theta^\star + \epsilon_{t,a}$, where $\epsilon_{t,a}$ is zero-mean Gaussian noise with variance $0.01$. We evaluate our BPE algorithm and the two baselines, LBGL and RBGL, on this dataset across various time horizons up to $T = 40,428,967$. We report the average performance over $50$ independent trials, visualizing the expected regret versus the time horizon on a log-log scale. In our experiments, the total number of batches for all algorithms is dynamically set to $M = \max(2, \lceil \log(T/d^2) \rceil)$. We implement and compare the following batch scheduling strategies:

\begin{itemize}
    \item \textbf{BPE (Geometric Batch Schedule):} BPE employs a geometric batch schedule. Specifically, the end time of the $m$-th batch is defined as $t'_m = \lfloor d^2 \cdot b^m \rfloor$ for $1 \le m < M$, where the growth factor is calculated as $b = (T/d^2)^{1/M}$.     The final batch naturally concludes at exactly $t'_M = T$.

    \item \textbf{RBGL and LBGL (Original vs. Unified Batch Schedules):} We refer to the batch schedules in \citet{ren2024dynamic} as the \textit{original batch schedules}; they induce a recursive, non-uniform grid. The first batch ends at $t'_1 = \lfloor b_{\text{val}} \sqrt{d} \rfloor$, and subsequent batches are recursively set to $t'_m = \lfloor b_{\text{val}} \sqrt{t'_{m-1}} \rfloor$ for $1 < m < M$, where the scaling constant is $b_{\text{val}} = \sqrt{T (T/d)^{1/(2^M-1)}}$. Under this schedule, the length of the first batch is $\Omega(\sqrt{dT})$, during which RBGL and LBGL suffer substantial regret. To ensure a fair comparison, we also evaluate RBGL and LBGL using the geometric grid schedule from BPE, which we refer to as the \textit{unified batch schedule}.
\end{itemize}

\begin{figure}[htbp]
    \centering
    \includegraphics[width=0.75\textwidth]{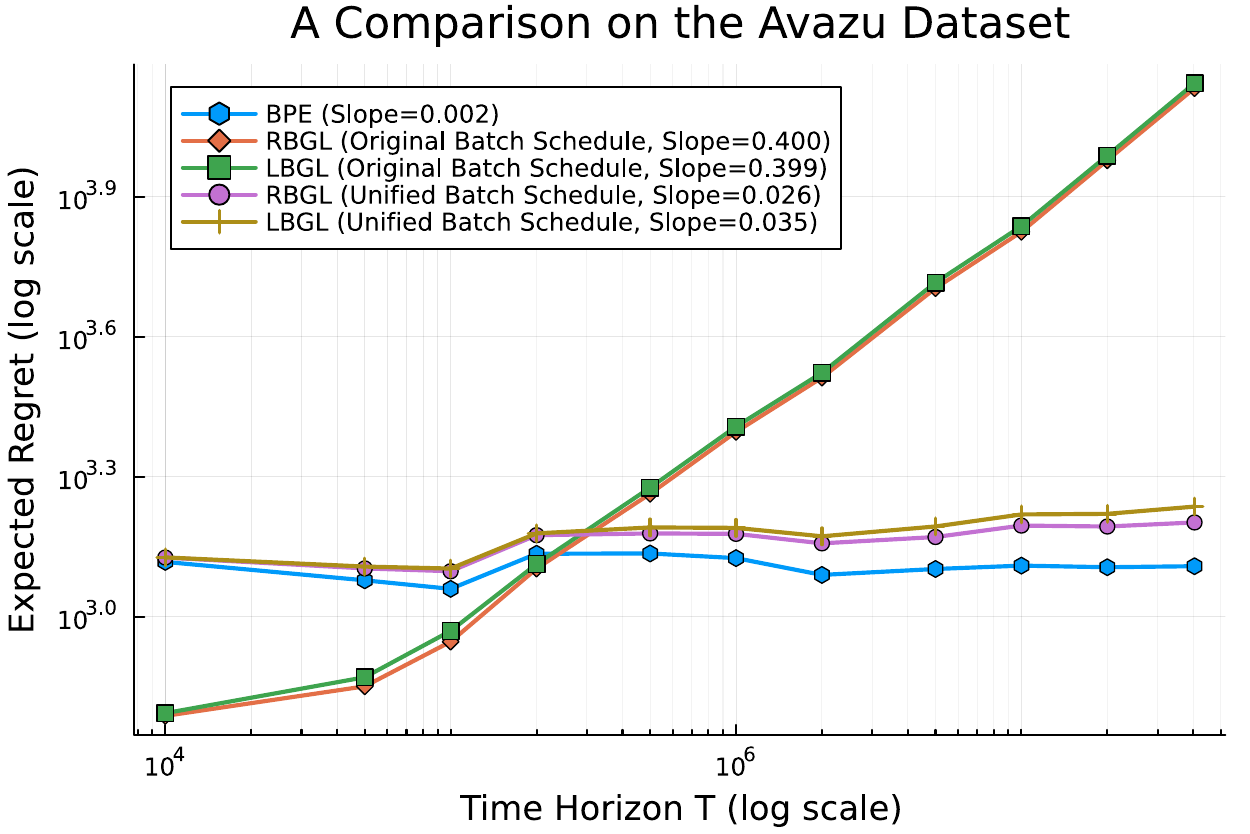}
    \caption{Log-log plot of Expected Regret vs. Time Horizon $T$ for the Avazu dataset.}
    \label{fig:problem_dependent_regret}
\end{figure}

\StochRev{Figure \ref{fig:problem_dependent_regret} shows that BPE has a nearly
flat regret slope of $0.002$. Under their original schedules, RBGL and LBGL
have slopes of $0.400$ and $0.399$, respectively, approaching the worst-case
$\mathcal{O}(\sqrt{T})$ rate. The unified schedule reduces these slopes to
$0.026$ and $0.035$, but both baselines remain inferior to BPE in overall
regret and asymptotic growth.}}
\section{Conclusion}

\StochRev{We study sequential batch learning in linear contextual bandits and
characterize its algorithms and theoretical performance. Our central finding
is that the nature of the contexts---adversarial or stochastic---fundamentally
affects both the minimax regret and the algorithms that attain it.}

\section*{Data and Code Accessibility}
Code and instructions for reproducing the numerical results are provided as
supplemental material. The public datasets are available from the UCI Machine
Learning Repository and Kaggle, with access instructions in the README file.

\appendix
\section{Definitions and Auxiliary Results}\label{appendix.auxiliary}

\begin{definition}
Let $(\mathcal{X}, \mathcal{F})$ be a measurable space and $P$, $Q$
be two probability measures on $(\mathcal{X}, \mathcal{F})$. 
\begin{enumerate}
	\item The total-variation distance between $P$ and $Q$ is defined as:
	$$ \mathsf{TV}(P,Q) = \sup_{A \in \mathcal{A}} |P(A) - Q(A)|.$$
	\item The KL-divergence between $P$ and $Q$ is:
	\begin{equation*}
	D_{\text{\rm KL}}(P\|Q) = \begin{cases}
	\int \log \frac{dP}{dQ} dP \text{\quad if $P << Q$} \\
	+\infty \text{\quad otherwise}
	\end{cases}
	\end{equation*}
\end{enumerate}

\end{definition}
\begin{lemma}\cite[Lemma 2.6]{tsybakov2009}\label{lemma.TV_KL}
	Let $P$ and $Q$ be any two probability measures on the same measurable space. Then
	\begin{align*}
	1- \mathsf{TV}(P,Q) \ge \frac{1}{2}\exp\left(-D_{\text{\rm KL}}(P\|Q)\right). 
	\end{align*}
\end{lemma}

\begin{lemma}\cite[Theorem 6.1]{wainwright2019high}
	\label{lemma.wishart}
Let $x_1,x_2,\cdots,x_n\sim \calN(0,I_d)$ be i.i.d. random vectors. Then for any $\delta>0$, 
\begin{align*}
\bP\left(\sigma_{\max}\left(\frac{1}{n}\sum_{i=1}^n x_ix_i^\top\right) \ge 1+\sqrt{\frac{d}{n}}+\delta \right) \le \exp\left(-\frac{n\delta^2}{2}\right),
\end{align*}
where $\sigma_{\max}(A)$ denotes the largest singular value of $A$. 
\end{lemma}

\section{Proof of Theorem~\ref{thm.adversarial}, Statement~1}
\label{appendix.adversarial.upper}

We begin with a fundamental result from linear algebra derived from the von Neumann trace inequality.

\begin{lemma}\label{lemma.eigenvalue}
Suppose that $A$ and $B$ are two PSD matrices with $A\preceq B$. Let $\lambda_1\ge \lambda_2\ge \cdots\ge \lambda_n$ be the eigenvalues of $A$, and $\mu_1\ge \mu_2\ge\cdots\ge \mu_n$ be the eigenvalues of $B$. Then
\begin{align*}
	\mathsf{Tr}(B^{-1}(B-A)) \le \sum_{i=1}^n \frac{\mu_i-\lambda_i}{\mu_i}. 
\end{align*}
\end{lemma}
\begin{proof}{Proof of Lemma~\ref{lemma.eigenvalue}.}
It is equivalent to prove that $\mathsf{Tr}(B^{-1}A)\ge \sum_{i=1}^n \lambda_i/\mu_i$. Applying the eigen-decomposition to the matrix $B$, it suffices to consider the case where $B = \text{diag}(\mu_1,\cdots,\mu_n)$. In this scenario, 
\begin{align*}
	\mathsf{Tr}(B^{-1}A) = \sum_{i=1}^n \frac{A_{ii}}{\mu_i} \ge \sum_{i=1}^n \frac{a_i}{\mu_i}, 
\end{align*}
where $a_1\ge a_2\ge \cdots\ge a_n$ is a decreasing rearrangement of $(A_{ii})_{i\in [n]}$, and the last inequality is due to the celebrated rearrangement inequality. Denote this quantity by the function $g(A_{11},A_{22},\cdots,A_{nn})$; then clearly $g$ is invariant with the permutation of inputs, and is also concave as it is the minimum of several affine functions. Schur's famous result claims that $(\lambda_1,\cdots,\lambda_n)$ majorizes $(A_{11},\cdots,A_{nn})$ (see, e.g. \cite[Chapter 9, Theorem B.1]{marshall2011inequalities}), thus by the majorization inequality, we have
\begin{align*}
	\sum_{i=1}^n \frac{a_i}{\mu_i} = g(A_{11},A_{22},\cdots,A_{nn}) \ge g(\lambda_1,\lambda_2,\cdots,\lambda_n) = \sum_{i=1}^n \frac{\lambda_i}{\mu_i},
\end{align*}
as claimed.\halmos
\end{proof}

\begin{lemma}\label{lemma.trace_sum}
Define $X_m = \sum_{t=t_{m-1}+1}^{t_m} x_{t,a_t}x_{t,a_t}^\top$. We have:
\begin{align*}
\sum_{m=1}^M \sqrt{ \mathsf{Tr}(A_{m-1}^{-1} X_m)} \le \log(T+1)\cdot \left(\sqrt{Md} + d\sqrt{\frac{T}{M}} \right). 
\end{align*}
\end{lemma} 
\begin{proof}{Proof of Lemma~\ref{lemma.trace_sum}.}
We start by noting that with the above notation, we have $A_m=A_{m-1}+X_m$ for any $m\in [M]$ with $A_0=I_d$. Let $\lambda_{m,1}\ge \lambda_{m,2}\ge \cdots\ge \lambda_{m,d}$ be all eigenvalues of $A_m$, then Lemma \ref{lemma.eigenvalue} shows that
\begin{align*}
	\mathsf{Tr}(A_{m}^{-1} X_m) = \mathsf{Tr}(A_{m}^{-1} (A_m - A_{m-1}) ) \le \sum_{j=1}^d \frac{\lambda_{m,j} - \lambda_{m-1,j}}{\lambda_{m,j}}, 
\end{align*}
and consequently
\begin{align}\label{eq:base_term}
\sum_{m=1}^M \sqrt{ \mathsf{Tr}(A_{m}^{-1} X_m) } &\le \sum_{m=1}^M \sqrt{\sum_{j=1}^d \frac{\lambda_{m,j} - \lambda_{m-1,j}}{\lambda_{m,j}}} \nonumber \\
&\le \sqrt{M}\cdot \sqrt{\sum_{j=1}^d \sum_{m=1}^M \frac{\lambda_{m,j} - \lambda_{m-1,j}}{\lambda_{m,j}}} \nonumber \\
&\le \sqrt{M}\cdot \sqrt{\sum_{j=1}^d \int_{\lambda_{0,j}}^{\lambda_{M,j}} \frac{dx}{x}} \le \sqrt{Md\log(T+1)}, 
\end{align}
where the last inequality follows from $\lambda_{M,j}\le \lambda_{0,j} + \sum_{t=1}^T \|x_{t,a_t}\|_2^2\le 1+T$. Using similar arguments, we could also have
\begin{align}\label{eq:base_square_term}
	\sum_{m=1}^M \mathsf{Tr}(A_{m}^{-1} X_m) \le d\log(T+1). 
\end{align}

Note that in the above inequality \eqref{eq:base_term}, the inversion is for $A_m$ instead of $A_{m-1}$. So in the sequel, we also need to upper bound the difference in doing so. We have the following chain of inequalities: 
\begin{equation}
\begin{aligned}\label{eq:difference_term}
&\sum_{m=1}^M \sqrt{\mathsf{Tr}(A_{m-1}^{-1}X_m)} - \sum_{m=1}^M \sqrt{\mathsf{Tr}(A_{m}^{-1}X_m)} \nonumber \stepa{\le} \sum_{m=1}^M \frac{\mathsf{Tr}(A_{m-1}^{-1}X_m) - \mathsf{Tr}(A_{m}^{-1}X_m) }{\sqrt{\mathsf{Tr}(A_{m-1}^{-1}X_m)}} \nonumber \\
&\stepb{=} \sum_{m=1}^M \frac{\mathsf{Tr}(A_m^{-1}X_mA_{m-1}^{-1}X_m) }{\sqrt{\mathsf{Tr}(A_{m-1}^{-1}X_m)}} \nonumber 
\stepc{\le} \sum_{m=1}^M \sqrt{\mathsf{Tr}(X_m^{1/2}A_m^{-1}X_mA_{m-1}^{-1}X_mA_m^{-1}X_m^{1/2} ) } \nonumber\\
&\stepd{\le} \sum_{m=1}^M \sqrt{\frac{T}{M}}\cdot \sqrt{\mathsf{Tr}((A_m^{-1}X_m)^2)} \nonumber 
\stepe{\le} \sum_{m=1}^M \sqrt{\frac{T}{M}}\cdot \mathsf{Tr}(A_m^{-1}X_m) 
\stepf{\le} \sqrt{\frac{T}{M}}\cdot d\log(T+1), 
\end{aligned}
\end{equation}
where 
\begin{itemize}
	\item (a) is due to the inequality $\sqrt{a}-\sqrt{b}\le (a-b)/\sqrt{a}$ for $a\ge b>0$;
	\item (b) follows from $A_m = A_{m-1}+X_m$ and simple matrix algebra; 
	\item (c) follows from the matrix \StochRev{Cauchy--Schwarz} inequality
	\begin{align*}
		\mathsf{Tr}(Y^\top Z) \le \sqrt{\mathsf{Tr}(YC^{-1}Y^\top)\cdot \mathsf{Tr}(ZCZ^\top) }, \qquad C\succ 0,  
	\end{align*}
	with $Y = X_m^{1/2}, Z = X_m^{1/2}A_m^{-1}X_mA_{m-1}^{-1}$, and $C = A_{m-1}$; 
	\item (d) uses $X_m=\sum_{t=t_{m-1}+1}^{t_m} x_{t,a_t}x_{t,a_t}^\top \preceq (t_m-t_{m-1})I_d = (T/M) I_d$ and $A_{m-1}\succeq A_0 = I_d$; 
	\item (e) uses $\sqrt{\mathsf{Tr}(X^2)}\le \mathsf{Tr}(X)$ for any PSD matrix $X$; 
	\item (f) makes use of the inequality \eqref{eq:base_square_term}. 
\end{itemize} 

Finally, the claimed lemma follows from \eqref{eq:base_term} and \eqref{eq:difference_term}. 
\halmos
\end{proof}

\begin{proof}{Proof of Theorem~\ref{thm.adversarial}, Statement~1.}
\begin{enumerate}
\item \textbf{Regret bound under conditional independence assumption.}

For a given $\delta > 0$, set the hyper-parameter $\gamma$ in Algorithm~\ref{algo.ucb} to be $1+\sqrt{\frac{1}{2}\log(\frac{2KT}{\delta})}$ for the entire proof.
Under the conditional independence assumption in Lemma~\ref{lemma.concentration},
by a simple union bound over all $t \in [T]$, we have with probability at least $1 - \delta$, the following event holds:
\begin{align*}
\forall m \in [M], \forall t \in [t_{m-1}+1, t_m], \forall a\in [K], \quad |x_{t,a}^\top (\hat{\theta}_{m-1} - \theta^\star)| \le \gamma\sqrt{x_{t,a}^\top A_{m-1}^{-1}x_{t,a}}.
\end{align*}
On this high probability event (with probability $1 - \delta$), we can bound the regret as follows:
\begin{align}
R_T(\textbf{Alg},\theta^\star) &= \sum_{t=1}^T \left( \max_{a\in [K]}x_{t,a}^\top \theta^\star  - x_{t,a_t}^\top \theta^\star \right)
= \sum_{m=1}^M \sum_{t=t_{m-1}+1}^{t_m} \left( \max_{a\in [K]}x_{t,a}^\top \theta^\star  - x_{t,a_t}^\top \theta^\star \right) \\
& \le \sum_{m=1}^M \sum_{t=t_{m-1}+1}^{t_m} \left( \max_{a\in [K]} \Big(x_{t,a}^\top \hat{\theta}_{m-1} +  \gamma\sqrt{x_{t,a}^\top A_{m-1}^{-1}x_{t,a}}\Big) - x_{t,a_t}^\top \theta^\star \right) \\
& = \sum_{m=1}^M \sum_{t=t_{m-1}+1}^{t_m} \left( x_{t,a_t}^\top \hat{\theta}_{m-1} +  \gamma\sqrt{x_{t,a_t}^\top A_{m-1}^{-1}x_{t,a_t}} - x_{t,a_t}^\top \theta^\star \right) \\
& = \sum_{m=1}^M \sum_{t=t_{m-1}+1}^{t_m} \left( x_{t,a_t}^\top (\hat{\theta}_{m-1} - \theta^\star) +  \gamma\sqrt{x_{t,a_t}^\top A_{m-1}^{-1}x_{t,a_t}}  \right)\\
& \le  
\sum_{m=1}^M \sum_{t=t_{m-1}+1}^{t_m} 2\gamma\sqrt{x_{t,a_t}^\top A_{m-1}^{-1}x_{t,a_t}} = 2\gamma\cdot \sum_{m=1}^M \sum_{t=t_{m-1}+1}^{t_m} 1\cdot \sqrt{x_{t,a_t}^\top A_{m-1}^{-1} x_{t,a_t}} \nonumber \\ \label{eq.regret_ucb}
&\le 2\gamma\sqrt{\frac{T}{M}}\cdot \sum_{m=1}^M \sqrt{\sum_{t=t_{m-1} +1}^{t_m} x_{t,a_t}^\top A_{m-1}^{-1} x_{t,a_t}} = 
 2\gamma\sqrt{\frac{T}{M}}\cdot \sum_{m=1}^M \sqrt{ \mathsf{Tr}(A_{m-1}^{-1} X_m)}, 
\end{align}
where the inequality in~\eqref{eq.regret_ucb} follows from \StochRev{Cauchy--Schwarz} and the choice of a uniform grid (without loss of generality we assume that $T/M$ is an integer).

Next, setting $\delta = \frac{1}{T}$ (and hence resulting in $\gamma = 1+\sqrt{\frac{1}{2}\log\left(2KT^2\right)}$) and  applying Lemma~\ref{lemma.trace_sum} to the upper bound in~\eqref{eq.regret_ucb}, we immediately obtain that again on this high-probability event:
\begin{align}
&R_T(\textbf{Alg},\theta^\star) \le  2\left(\sqrt{\frac{1}{2}\log\left(2KT^2\right)}+1\right)\log(T+1)\sqrt{\frac{T}{M}}\left(\sqrt{Md} + d\sqrt{\frac{T}{M}} \right)\\ 
&\label{eq.x}= \mathsf{polylog}(T)\cdot (\sqrt{dT} + \frac{dT}{M}).
\end{align}
Consequently, taking the expectation of $R_T(\textbf{Alg},\theta^\star)$ yields the same bound as in Equation~\eqref{eq.x}, since with probability at most $\frac{1}{T}$, the total regret over the entire horizon is at most $T$ (each time accumulates at most a regret of $1$ by the normalization assumption).
Since the regret bound is independent of $\theta^\star$, it 
immediately follows that 
$\sup_{\theta^\star: \|\theta^\star\|_2\le 1} \mathbb{E}[R_T(\textbf{Alg},\theta^\star)]
\le \mathsf{polylog}(T)\cdot (\sqrt{dT} + \frac{dT}{M}).$ 

\item \textbf{Building a Master algorithm that satisfies conditional independence}

To complete the proof, we need to validate the conditional independence assumption in Lemma~\ref{lemma.concentration}. Since the length of the confidence intervals does not depend on the random rewards, this task can be done by using a master algorithm SupSBUCB (Algorithm~\ref{algo:SupSBUCB}), which runs in $O(\log T)$ stages at each time step $t$ similar to \cite{auer2002using}, which is subsequently adopted in the linear contextual bandits setting~\cite{chu2011contextual} and then in the generalized linear contextual bandits setting~\cite{li2017provably} for the same purpose of meeting the conditional independence assumption. 
Note that SupSBUCB is responsible for selecting the actions $a_t$ and it does so by calling BaseSBUCB (Algorithm~\ref{algo:BaseSBUCB}), which merely performs regression.
This master-base algorithm pair has by now become a standard trick to get around the conditional dependency in the vanilla UCB algorithm for a variety of contextual bandits problems (by sacrificing at most $O(\log T)$ regret). The pseudocode is given in Section~\ref{subsec.UCB_upperbound}.

More specifically, the master algorithm developed by~\cite{auer2002finite} has the following structure: each time step is divided into at most $\log T$ stages. At the beginning of each stage $s$, the learner computes the confidence interval using only the previous contexts designated as belonging to that stage and selects any action whose confidence interval has a large length (exceeding some threshold). If all actions \StochRev{have} a small confidence interval, then we end this stage, observe the rewards of the given contexts and move on to the next stage with a smaller threshold on the length of the confidence interval. In other words, conditional independence is obtained by successive manual masking and revealing of certain information. One can intuitively think of each stage $s$ as a color, and each time step $t$ is colored using one of the $\log T$ colors (if colored at all). When computing confidence intervals and performing regression, only previous contexts that have the same color are used, instead of all previous contexts. 

Adapting this algorithm to the sequential batch setting is not difficult: we merely keep track
of the sets $\Psi_{m}^s$ ($s \in [\log T]$) \StochRev{for each} batch $m$ (rather than \StochRev{for each} time step $t$ as in the fully online learning case). Note that we are still coloring each time step $t$. \StochRev{The difference lies} in the frequency at which we are running BaseSBUCB to compute the confidence bounds and rewards. 
\StochRev{Because of the close similarity,} we omit the details here and refer to \cite[Section 4.3]{auer2002using}. In particular, by establishing similar results to \cite[Lemma 15, Lemma 16]{auer2002using}, it is straightforward to show that the regret of the master algorithm SupSBUCB here is enlarged at most by a multiplicative factor of $O(\log T)$, which leads to the upper bound in Theorem \ref{thm.adversarial}. 
\end{enumerate}
\halmos
\end{proof}

\section{Proof of Theorem~\ref{thm.stochastic}}
\label{appendix.stochastic.upper}

\subsection{Regret Analysis for the Upper Bound}
We now turn to establishing the upper bound in Theorem \ref{thm.stochastic}. 
We again execute a two-step program. First, we prove that Algorithm \ref{algo.pure-exp} with the grid $\calT=\{t_1,\cdots,t_M\}$ in \eqref{eq.minimax_grid} attains the regret upper bound in Theorem \ref{thm.stochastic}, assuming the conditional independence assumption (cf. Lemma \ref{lemma.difference}) holds. Second, similar to the master algorithm in the previous section, we then modify Algorithm \ref{algo.pure-exp} slightly to validate this condition. One thing to note here is that, unlike in the adversarial contexts case, here the modification is much simpler, as we shall see later.

We start by establishing that the least squares estimator $\hat{\theta}$ is close to the true parameter $\theta^\star$ at the beginning of every batch with high probability. By the theory of least squares, this would be obvious if the chosen contexts $x_{t,a_t}$ were i.i.d. Gaussian. However, since the action $a_t$ depends on all contexts $(x_{t,a})_{a\in [K]}$ available at time $t$, the probability distribution of $x_{t,a_t}$ may be far from isotropic. Consequently, a priori, there might be one or more directions in the context space that were never chosen, hence yielding inaccurate estimation of $\theta^\star$ along that (or those) direction(s). However, as we shall see next, this is not a concern: we establish that the matrix formed by the selected contexts \StochRev{is reasonably} well-conditioned, despite being selected in a greedy fashion.

\begin{lemma}\label{lemma.equator}
{
	Under the stochastic-context model specified in
	Section~\ref{sec:stochastic}, consider any batch $m\in[M]$. Suppose
	that $\hat{\theta}_{m-1}$ is measurable with respect to
	$\mathcal{H}_{t_{m-1}}$, is held fixed throughout batch $m$, and
	\[
	a_t
	=
	\arg\max_{a\in[K]}x_{t,a}^{\top}\hat{\theta}_{m-1},
	\qquad
	t_{m-1}<t\le t_m,
	\]
	where ties are resolved according to a fixed rule independent of the
	current contexts. For the grid in~\eqref{eq.minimax_grid}, with
	$M=O(\log\log T)$, conditional on $\mathcal{H}_{t_{m-1}}$, with
	probability at least $1-O(T^{-4})$,
	\[
	\lambda_{\min}\!\left(
	\sum_{t=t_{m-1}+1}^{t_m}
	x_{t,a_t}x_{t,a_t}^{\top}
	\right)
	\ge
	c\,\frac{\kappa(t_m-t_{m-1})}{d},
	\]
	where $c>0$ is a numerical constant independent of
	$(K,T,d,m,\kappa)$. Consequently, the same bound holds
	unconditionally with probability at least $1-O(T^{-4})$.
}
\end{lemma}

The proof of Lemma~\ref{lemma.equator} is given in Section~\ref{sec:proof-main-lemmas} below. Based on Lemma \ref{lemma.equator}, we are ready to show that the least squares estimator $\hat{\theta}$ is close to the true parameter $\theta^\star$ with high probability. For $m\in [M]$, let $\hat{\theta}_m$ be the estimate at the end of \StochRev{the} $m$-th batch, and $A_m = \sum_{t=1}^{t_m} x_{t,a_t}x_{t,a_t}^\top$ be the regression matrix. 
\begin{lemma}\label{lemma.difference}
	For each $m\in [M]$, if the rewards $\{r_{t,a_t}\}_{t\in [t_m]}$ up to time $t_m$ are mutually independent given the selected contexts $\{x_{t,a_t}\}_{t\in [t_m]}$, then with probability at least $1-O(T^{-3})$,
	\begin{align*}
	\|\hat{\theta}_m - \theta^\star\|_2 \le Cd\cdot \sqrt{\frac{\log T}{\kappa t_m}}
	\end{align*}
	for a numerical constant $C>0$ independent of $(K,T,d,m,\kappa)$. 
\end{lemma}

Lemma~\ref{lemma.difference} shows that given the conditional independence assumption, the estimator $\hat{\theta}$ given by pure exploitation essentially achieves the rate-optimal estimation of $\theta^\star$ even if one purely explores. This now positions us well to prove the upper bound of Theorem \ref{thm.stochastic}. Of course, bear in mind that when using Algorithm~\ref{algo.pure-exp}, the conditional independence assumption does not hold, for the choice of future contexts depends on the rewards in the previous batches. Therefore, we will use sample splitting to build another master algorithm to gain independence at the cost of the sample size reduction by a multiplicative factor of $M$ (recall that $M = O(\log\log T)$). The following proof implements these two steps; note that in this setting, the master algorithm is entirely different from and much simpler than the one given in the adversarial case.

\begin{proof}{Proof of Theorem~\ref{thm.stochastic}, Statement~1.}
	\noindent

\begin{enumerate}
\item \textbf{Regret bound under conditional independence assumption.}
	 Consider \StochRev{the} $m$-th batch with any $m\ge 2$, and any time point $t$ inside this batch. By the definition of $a_t$, we have $x_{t,a_t}^\top \hat{\theta}_{m-1}\ge x_{t,a}^\top \hat{\theta}_{m-1}$ for any $a\in [K]$. Consequently, 
\begin{align*}
\max_{a\in [K]} (x_{t,a} - x_{t,a_t})^\top \theta^\star &\le \max_{a\in [K]} (x_{t,a} - x_{t,a_t})^\top (\theta^\star - \hat{\theta}_{m-1}) \\
&\le \max_{a,a'\in [K]} (x_{t,a} - x_{t,a'})^\top (\theta^\star - \hat{\theta}_{m-1}) \\
&\le 2\max_{a\in [K]} |x_{t,a}^\top (\theta^\star - \hat{\theta}_{m-1})|. 
\end{align*}
For fixed $a\in [K]$, marginally we have $x_{t,a}\sim \calN(0,\Sigma)$ independent of $\hat{\theta}_{m-1}$. Therefore, conditioning on the previous contexts and rewards, we have $x_{t,a}^\top (\theta^\star - \hat{\theta}_{m-1})\sim \calN(0,\sigma^2)$ with
$$
\sigma^2 = (\theta^\star - \hat{\theta}_{m-1})^\top \Sigma (\theta^\star - \hat{\theta}_{m-1}) \le \frac{\|\theta^\star - \hat{\theta}_{m-1}\|_2^2}{d}
$$
by Assumption \ref{assumption:cov}. By a union bound over $a\in [K]$, with probability at least $1-O(T^{-3})$ over the randomness in the current batch we have
\begin{align*}
\max_{a\in [K]} (x_{t,a} - x_{t,a_t})^\top \theta^\star \le  2\max_{a\in [K]} |x_{t,a}^\top (\theta^\star - \hat{\theta}_{m-1})| = O\left(\|\theta^\star - \hat{\theta}_{m-1} \|_2 \cdot \sqrt{\frac{\log(KT)}{d}}\right). 
\end{align*}
Applying Lemma \ref{lemma.difference} and another union bound, there exists some numerical constant $C'>0$ such that with probability at least $1-O(T^{-3})$, the \StochRev{instantaneous} regret at time $t$ is at most
\begin{align*}
\max_{a\in [K]} (x_{t,a} - x_{t,a_t})^\top \theta^\star \le C'\sqrt{\log(KT)\log T}\cdot \sqrt{\frac{d}{\kappa t_{m-1}}}. 
\end{align*}
Now taking the union bound over $t\in [T]$, the total regret incurred after the first batch is at most
\begin{align}\label{eq.later_batch}
\sum_{m=2}^M C'\sqrt{\log(KT)\log T}\cdot t_m\sqrt{\frac{d}{\kappa t_{m-1}}} \le C'\sqrt{\frac{\log(KT)\log T}{\kappa}}M\cdot \beta\sqrt{d}
\end{align}
with probability at least $1-O(T^{-2})$, where the inequality is due to the choice of the grid in \eqref{eq.minimax_grid}. 

As for the first batch, the \StochRev{instantaneous} regret at any time point $t$ is at most the maximum of $K$ Gaussian random variables $\calN(0,(\theta^\star)^\top \Sigma \theta^\star)$. Since $\|\theta^\star\|_2\le 1$ and $\lambda_{\max}(\Sigma)\le 1/d$, we conclude that the \StochRev{instantaneous} regret is at most $C''\sqrt{\log(KT)/d}$ for some constant $C''>0$ with probability at least $1-O(T^{-3})$. Now by a union bound over $t\in [t_1]$, with probability at least $1-O(T^{-2})$ the total regret in the first batch is at most
\begin{align}\label{eq.first_batch}
C''\sqrt{\log(KT)/d}\cdot t_1 = C''\sqrt{\log(KT)}\cdot \beta\sqrt{d}. 
\end{align}

Combining \eqref{eq.later_batch} and \eqref{eq.first_batch} with the
definition of $\beta$ in Algorithm~\ref{algo.pure-exp}, we obtain
\[
O\!\left(
M\sqrt{\log(KT)\log T}
\sqrt{\frac{dT}{\kappa}}
\left(\frac{T}{d^2}\right)^{\frac{1}{2(2^M-1)}}
\right)
\]
under the conditional independence condition. Since
$M=O(\log\log T)$ and $K=O(\mathsf{poly}(d))$, this is
\[
\widetilde{O}\!\left(
\sqrt{\frac{dT\log K}{\kappa}}
\left(\frac{T}{d^2}\right)^{\frac{1}{2(2^M-1)}}
\right).
\]

\item \textbf{Building a Master algorithm that satisfies conditional independence}

We start by proposing a sample splitting based master algorithm (see Algorithm~\ref{algo.sample_splitting}) that ensures that when restricting to the subset of observations used for constructing $\hat{\theta}$, the rewards are conditionally independent given the contexts. 
The key modification in Algorithm \ref{algo.sample_splitting} lies in the computation of the estimator $\hat{\theta}_{m}$ after the first $m$ batches. Specifically, instead of using all past contexts and rewards before $t_m$, we only use the past observations inside the time frame $T^{(m)}\subsetneq [t_m]$ to construct the estimator. The key property of the time frames is the disjointness, i.e., $T^{(1)},\cdots,T^{(M)}$ are pairwise disjoint. Then the following lemma shows that the conditional independence condition holds within each time frame $T^{(m)}$.

\begin{lemma}\label{lemma.cond_indep}
		For each $m\in [M]$, the rewards $\{r_{t,a_t}\}_{t\in T^{(m)}}$ are \StochRev{conditionally independent given} the selected contexts $\{x_{t,a_t}\}_{t\in T^{(m)}}$. 
\end{lemma}
\begin{proof}{Proof of Lemma~\ref{lemma.cond_indep}.}
	For $t\in T^{(m)}$, the action $a_t$ only depends on the contexts $\{x_{t,a}\}_{a\in [K]}$ at time $t$ and the past estimators $\hat{\theta}_1, \cdots, \hat{\theta}_{m-1}$. However, for any $m'\in [m-1]$, the estimator $\hat{\theta}_{m'}$ only depends on the contexts $x_{\tau,a_\tau}$ and rewards $r_{\tau,a_\tau}$ with $\tau\in T^{(m')}$. Repeating the same arguments for the action $a_\tau$ with $\tau\in T^{(m')}$, we conclude that $a_t$ only depends on the contexts $\{x_{\tau,a}\}_{a\in [K],\tau\in \cup_{m'\le m-1} T^{(m')}\cup \{t\}}$ and rewards $\{r_{\tau,a_\tau} \}_{\tau\in \cup_{m'\le m-1} T^{(m')}}$. Consequently, by the disjointness of $T^{(m)}$ and $\cup_{m'\le m-1} T^{(m')}$, the desired conditional independence holds.\halmos
\end{proof}

By Lemma \ref{lemma.cond_indep}, the conditional independence condition of Lemma \ref{lemma.difference} holds for Algorithm \ref{algo.sample_splitting}. Sample splitting reduces the effective sample size by at most a factor of $M=O(\log\log T)$ and therefore introduces only an additional polylogarithmic factor. This factor is absorbed by $\widetilde{O}(\cdot)$, which proves the upper bound stated in Theorem~\ref{thm.stochastic}.
\end{enumerate}
\halmos
\end{proof}
\ZH{
\subsection{Regret Analysis for the Lower Bound}
\label{appendix.stochastic.lower}

Before presenting the proofs, we formally establish the filtration and several local notations used throughout this section. Recall from Section~\ref{subsec:decision} that the algorithm operates over a batch grid $\mathcal{T} = \{t_1, \dots, t_M\}$. For any batch $m \in [M]$, let $\mathcal{H}_{t_{m-1}}$ denote the $\sigma$-algebra generated by all historical information fully observed by the end of batch $m-1$:
\[
\mathcal{H}_{t_{m-1}} \triangleq \sigma\left( \{x_{\tau,a} \mid a \in [K]\}_{\tau=1}^{t_{m-1}} \cup \{a_\tau, r_{\tau, a_\tau}\}_{\tau=1}^{t_{m-1}} \right).
\]
During batch $m$ (i.e., for steps $t \in (t_{m-1}, t_m]$), the policy $\pi_t$ is adapted to $\mathcal{H}_{t_{m-1}}$ and the newly arrived set of contexts, which we denote as $X_t \triangleq \{x_{t,a}\}_{a=1}^K$. Furthermore, while $r_{t,a}$ denotes the realized reward in our general formulation, we use $r_t$ in the subsequent analysis to denote the expected conditional \emph{instantaneous regret} at time $t$:
\[
r_t \triangleq \max_{a \in [K]} x_{t,a}^\top \theta^\star - x_{t,a_t}^\top \theta^\star.
\]

The argument proceeds in three steps. First, Lemma~\ref{lemma.geometric_regret_identity} in Appendix~\ref{appendix.stochastic.lower} lower bounds the conditional expected instantaneous regret by a geometric quantity involving the component of $\theta^\star$ orthogonal to the direction induced by a policy based only on information available before the current batch. Second, Lemma~\ref{lemma.orthogonal_trace} lower bounds the orthogonal variance under a Gaussian prior using the posterior precision matrix, which remains frozen within each batch. Third, the adversary chooses a family of priors $\rho_j$ and truncation events $E_j$ targeting different batch indices $j$, combines the resulting per-step bounds into a regret lower bound of order $\sqrt{d}\,t_j/\sqrt{t_{j-1}}$ for each $j$ (see \eqref{eq:lower_bound_tj} in the appendix), and optimizes over $j$ via a weighted geometric-mean cascade along the grid $\mathcal{T}$.

\begin{lemma}[Universal Geometric Regret Identity]\label{lemma.geometric_regret_identity}
    For any step $t \in (t_{m-1}, t_m]$, let $u_t$ be an $\mathcal{H}_{t_{m-1}}$-measurable unit vector. For independent contexts $x_{t,a} \sim \mathcal{N}(0, I_d/d)$, the expected conditional instantaneous regret satisfies:
    \[
    \mathbb{E}_{X_t}[r_t \mid \theta^\star] \ge \frac{C_K}{2\sqrt{d}\|\theta^\star\|_2} \|(I - u_t u_t^\top)\theta^\star\|_2^2
    \]
    where $g_1,\ldots,g_K$ are independent standard Gaussian random variables and $C_K=\mathbb{E}[\max_{a\in[K]}g_a]=\Theta(\sqrt{\log K})$.
\end{lemma}

\begin{proof}{Proof of Lemma~\ref{lemma.geometric_regret_identity}.}
	    At time $t$, new contexts $X_t = \{x_{t,a}\}_{a=1}^K$ arrive independently of the history $\mathcal{H}_{t_{m-1}}$. Let $v_t = \mathbb{E}_{X_t}[x_{t, a_t} \mid \mathcal{H}_{t_{m-1}}]$. Because the contexts are isotropic Gaussians, for \StochRev{any policy}, the norm of the conditionally expected selected context is bounded by the optimal greedy context: 
    \[ 
    \|v_t\|_2 = \sup_{\|w\|_2 \le 1} \mathbb{E}_{X_t}[x_{t, a_t}^\top w \mid \mathcal{H}_{t_{m-1}}] \le \sup_{\|w\|_2 \le 1} \mathbb{E}_{X_t}\left[\max_{a \in [K]} x_{t,a}^\top w\right] = \frac{C_K}{\sqrt{d}}, 
    \]
    where $g_1,\ldots,g_K$ are independent standard Gaussian random variables and the classical estimate for Gaussian maxima gives $C_K=\mathbb{E}[\max_{a\in[K]}g_a]=\Theta(\sqrt{\log K})$.
    
    The expected conditional instantaneous regret obeys:
    \[
    \mathbb{E}_{X_t}[r_t \mid \theta^\star] = \mathbb{E}_{X_t}\left[\max_{a} x_{t,a}^\top \theta^\star\right] - \mathbb{E}_{X_t}[x_{t,a_t}^\top \theta^\star] = \frac{C_K}{\sqrt{d}}\|\theta^\star\|_2 - v_t^\top \theta^\star.
    \]
    
    Let $u_t = v_t / \|v_t\|_2$ if $v_t \ne 0$ (otherwise, an arbitrary unit vector). We utilize the elementary algebraic inequality $a - cb \ge \frac{1}{2a}(a^2 - b^2)$ for $a \ge 0$, $b \in [-a, a]$, and $c \in [0,1]$. The preceding bound $\|v_t\|_2\le C_K/\sqrt{d}$ implies that $c=\|v_t\|_2/(C_K/\sqrt{d})\in[0,1]$. Substituting $a = \|\theta^\star\|_2$, $b = u_t^\top \theta^\star$, and this value of $c$, we map the regret to the orthogonal variance:
    \begin{align*}
    \mathbb{E}_{X_t}[r_t \mid \theta^\star] &= \frac{C_K}{\sqrt{d}} \left( \|\theta^\star\|_2 - c (u_t^\top \theta^\star) \right) \\
    &\ge \frac{C_K}{2\sqrt{d}\|\theta^\star\|_2} \left( \|\theta^\star\|_2^2 - (u_t^\top \theta^\star)^2 \right) \\
    &= \frac{C_K}{2\sqrt{d}\|\theta^\star\|_2} \|(I - u_t u_t^\top)\theta^\star\|_2^2.
    \end{align*}
\halmos
\end{proof}

\begin{lemma}[Orthogonal Posterior Variance Bound]\label{lemma.orthogonal_trace}
Assume that the adversary draws
$\theta^\star\sim\mathcal{N}(0,\nu^2I_d/d)$. Let
$\Sigma_{m-1}^{-1}
=(d/\nu^2)I_d
+\sum_{\tau=1}^{t_{m-1}}x_{\tau,a_\tau}x_{\tau,a_\tau}^\top$
be the posterior precision matrix conditional on
$\mathcal{H}_{t_{m-1}}$. For any
$\mathcal{H}_{t_{m-1}}$-measurable unit vector $u_t$, the posterior
uncertainty orthogonal to $u_t$ satisfies
\[
\mathbb{E}\!\left[
\left\|(I-u_tu_t^\top)\theta^\star\right\|_2^2
\right]
\ge
\mathbb{E}\!\left[
\operatorname{Tr}\!\left((I-u_tu_t^\top)\Sigma_{m-1}\right)
\right]
\ge
\frac{(d-1)^2}
{\mathbb{E}[\operatorname{Tr}(\Sigma_{m-1}^{-1})]}.
\]
Furthermore,
$\mathbb{E}[\operatorname{Tr}(\Sigma_{m-1}^{-1})]
\le d^2/\nu^2+\gamma_Kt_{m-1}$,
where $\gamma_K=O(1)$.
\end{lemma}

\begin{proof}{Proof of Lemma~\ref{lemma.orthogonal_trace}.}
    Conditional on $\mathcal{H}_{t_{m-1}}$, the posterior distribution of $\theta^\star$ is exactly Gaussian: $\theta^\star\mid\mathcal{H}_{t_{m-1}}\sim\mathcal{N}(\bar{\theta}_{m-1},\Sigma_{m-1})$. Decomposing the conditional second moment into its posterior-mean and posterior-variance components gives
    \begin{align*}
        \mathbb{E}[\|(I - u_t u_t^\top)\theta^\star\|_2^2 \mid \mathcal{H}_{t_{m-1}}] &= \|(I - u_t u_t^\top)\bar{\theta}_{m-1}\|_2^2 + \text{Tr}((I - u_t u_t^\top)\Sigma_{m-1}) \\
        &\ge \text{Tr}((I - u_t u_t^\top)\Sigma_{m-1}).
    \end{align*}
    
    Because $u_t$ is a unit vector,
    \[
    u_t^\top\Sigma_{m-1}u_t
    \le \lambda_{\max}(\Sigma_{m-1}).
    \]
    Therefore,
    \[
    \operatorname{Tr}\!\left((I-u_tu_t^\top)\Sigma_{m-1}\right)
    =
    \operatorname{Tr}(\Sigma_{m-1})
    -u_t^\top\Sigma_{m-1}u_t
    \]
    is bounded below by the sum of the $d-1$ smallest eigenvalues of
    $\Sigma_{m-1}$.
    
    Let $\lambda_1 \ge \dots \ge \lambda_d > 0$ denote the eigenvalues of the precision matrix $\Sigma_{m-1}^{-1}$. The $d-1$ smallest eigenvalues of $\Sigma_{m-1}$ are precisely $\lambda_1^{-1}, \dots, \lambda_{d-1}^{-1}$. By the Arithmetic-Harmonic Mean (AM-HM) eigenvalue inequality across the remaining dimensions, we have:
    \[
    \sum_{i=1}^{d-1} \lambda_i^{-1} \ge \frac{(d-1)^2}{\sum_{i=1}^{d-1} \lambda_i} \ge \frac{(d-1)^2}{\text{Tr}(\Sigma_{m-1}^{-1})}.
    \]
    Thus, $\text{Tr}((I - u_t u_t^\top)\Sigma_{m-1}) \ge \frac{(d-1)^2}{\text{Tr}(\Sigma_{m-1}^{-1})}$.
    
    Taking the unconditional expectation of both sides (and applying Jensen's inequality for the convex function $f(x)=1/x$ as necessary for the final transition), we establish the first part of the lemma:
    \[
    \mathbb{E}[\|(I - u_t u_t^\top)\theta^\star\|_2^2] \ge \mathbb{E}[\text{Tr}((I - u_t u_t^\top)\Sigma_{m-1})] \ge \mathbb{E} \left[ \frac{(d-1)^2}{\text{Tr}(\Sigma_{m-1}^{-1})} \right] \ge \frac{(d-1)^2}{\mathbb{E}[\text{Tr}(\Sigma_{m-1}^{-1})]}.
    \]
    
    Because the observation model is linear Gaussian, the exact posterior precision matrix is $\Sigma_{m-1}^{-1} = \frac{d}{\nu^2} I_d + \sum_{\tau=1}^{t_{m-1}} x_{\tau, a_\tau} x_{\tau, a_\tau}^\top$. Taking the unconditional expectation of its trace gives:
    \[
    \mathbb{E}[\text{Tr}(\Sigma_{m-1}^{-1})] = \frac{d^2}{\nu^2} + \sum_{\tau=1}^{t_{m-1}} \mathbb{E}[\|x_{\tau, a_\tau}\|_2^2].
    \]
    Let
    \[
    \gamma_K
    =
    \mathbb{E}\!\left[
    \max_{a\in[K]}\|x_{\tau,a}\|_2^2
    \right].
    \]
    Since $d\|x_{\tau,a}\|_2^2\sim\chi_d^2$, standard chi-square
    concentration combined with a union bound over the $K$ actions gives
    \[
    \gamma_K
    \le
    1+C\left(
    \sqrt{\frac{\log K}{d}}
    +
    \frac{\log K}{d}
    \right)
    \]
    for a universal constant $C>0$. By Assumption~\ref{aspn.TKd},
    $K=O(\mathsf{poly}(d))$, and hence $\log K=O(\log d)$. Consequently,
    \[
    \gamma_K
    \le
    1+O\left(
    \sqrt{\frac{\log d}{d}}
    +
    \frac{\log d}{d}
    \right)
    =
    1+o(1)
    =
    O(1).
    \]
    Therefore, we conclude:
    \[
    \mathbb{E}[\text{Tr}(\Sigma_{m-1}^{-1})] \le \frac{d^2}{\nu^2} + \gamma_K t_{m-1}.
    \]
\halmos
\end{proof}

\begin{proof}{Proof of Theorem~\ref{thm.stochastic}, Statement~2.}
    Let $s = \min\{m \in [M] : t_m \ge d^2\}$. The adversary constructs $M-s+1$ distinct testing priors targeted at different batch intervals. For each targeted batch $j\in\{s+1,\dots,M\}$, the adversary selects the prior $\rho_j=\mathcal{N}(0,\nu_j^2I_d/d)$ with $\nu_j=d/(2\sqrt{\gamma_Kt_{j-1}})$. Since $t_{j-1}\ge t_s\ge d^2$ and $\gamma_K\ge1$, it follows that $\nu_j\le1/2$.

    Let $E_j$ be the event $\{ \|\theta^\star\|_2 \le 2\nu_j \}$, ensuring $\|\theta^\star\|_2 \le 1$. Because event $E_j$ mandates $1/\|\theta^\star\|_2 \ge 1/(2\nu_j)$, Lemma~\ref{lemma.geometric_regret_identity} guarantees the expected conditional instantaneous regret satisfies:
    \[
    \mathbb{E}[r_t \mid \theta^\star] \mathbbm{1}_{E_j} \ge \frac{C_K}{4\nu_j\sqrt{d}} \|(I - u_t u_t^\top)\theta^\star\|_2^2 \mathbbm{1}_{E_j}.
    \]
    Using $\mathbbm{1}_{E_j}+\mathbbm{1}_{E_j^c}=1$ and linearity of expectation, we have
    \[
    \mathbb{E}_{\theta^\star\sim\rho_j}
    [X\mathbbm{1}_{E_j}]
    =
    \mathbb{E}_{\theta^\star\sim\rho_j}[X]
    -
    \mathbb{E}_{\theta^\star\sim\rho_j}
    [X\mathbbm{1}_{E_j^c}].
    \]
    To bound the tail term, let
    $Y=d\|\theta^\star\|_2^2/\nu_j^2\sim\chi_d^2$. Then
    \[
    \mathbb{E}_{\theta^\star\sim\rho_j}\!\left[
    \|\theta^\star\|_2^2
    \mathbbm{1}_{\{\|\theta^\star\|_2>2\nu_j\}}
    \right]
    =
    \frac{\nu_j^2}{d}
    \mathbb{E}\!\left[
    Y\mathbbm{1}_{\{Y>4d\}}
    \right]
    \le 0.092\nu_j^2,
    \]
    where the last inequality follows from the chi-square tail calculation
    $\sup_{d\ge2}d^{-1}\mathbb{E}[Y\mathbbm{1}_{\{Y>4d\}}]
    =5e^{-4}<0.092$. Applying Lemma~\ref{lemma.orthogonal_trace} to bound the posterior uncertainty orthogonal to $u_t$ under the prior, together with this tail bound, we obtain the Bayesian expected instantaneous regret:
    \[
    \mathbb{E}_{\theta^\star \sim \rho_j} \left[ \mathbb{E}[r_t \mid \theta^\star] \mathbbm{1}_{E_j} \right] \ge \frac{C_K}{4\nu_j\sqrt{d}} \left( \frac{(d-1)^2}{d^2/\nu_j^2 + \gamma_K t_{m-1}} - 0.092 \nu_j^2 \right).
    \]

    Consider the total regret accumulated up to $t_j$. For any step $t \le t_j$, its associated batch $m$ satisfies $t_{m-1} \le t_{j-1}$. Therefore, substituting $\nu_j$ into the denominator above forces $d^2/\nu_j^2 + \gamma_K t_{m-1} \le 5\gamma_K t_{j-1}$.
    
    For $d \ge 2$, the trace term algebraically guarantees: $\frac{(d-1)^2}{5\gamma_K t_{j-1}} = \frac{4(d-1)^2}{5d^2} \nu_j^2 \ge 0.2 \nu_j^2$. Subtracting the tail error leaves a strictly positive geometric margin of $(0.2 - 0.092)\nu_j^2 = 0.108\nu_j^2$. Consequently, the Bayesian expected instantaneous regret for every single step $1 \le t \le t_j$ is uniformly bounded by:
    \[
    \mathbb{E}_{\theta^\star \sim \rho_j} \left[ \mathbb{E}[r_t \mid \theta^\star] \mathbbm{1}_{E_j} \right] \ge \frac{C_K}{4\nu_j\sqrt{d}} (0.108 \nu_j^2) = 0.027 \frac{C_K}{\sqrt{d}} \nu_j.
    \]
    Summing this constant lower bound uniformly from step 1 up to $t_j$ forces:
    \begin{equation}
    \mathbb{E}_{\theta^\star \sim \rho_j} \left[ \mathbb{E} [R_T\left(\textbf{Alg}, \theta^\star\right)] \mathbbm{1}_{E_j} \right] \ge \sum_{t=1}^{t_j} \mathbb{E}_{\theta^\star \sim \rho_j} \left[ \mathbb{E}[r_t \mid \theta^\star] \mathbbm{1}_{E_j} \right] \ge t_j \left( 0.027 \frac{C_K}{\sqrt{d}} \frac{d}{2\sqrt{\gamma_K t_{j-1}}} \right) = c_0 \frac{C_K}{\sqrt{\gamma_K}} \sqrt{d} \frac{t_j}{\sqrt{t_{j-1}}}. \label{eq:lower_bound_tj}
    \end{equation}

    By standard minimax theory, the supremum of the expected regret over the unit ball is lower-bounded by the Bayesian expected regret over any prior distribution supported on (or bounded within) the unit ball. Because the event $E_j$ ensures $\|\theta^\star\|_2 \le 1$, we have:
    \[
    \sup_{\|\theta^\star\|_2 \le 1} \mathbb{E}[R_T(\textbf{Alg}, \theta^\star)] \ge \mathbb{E}_{\theta^\star \sim \rho_j} \left[ \mathbb{E} [R_T\left(\textbf{Alg}, \theta^\star\right)] \mathbbm{1}_{E_j} \right].
    \]
    By mirroring the logic in \eqref{eq:lower_bound_tj} for the initialization limit $j=s$ (setting $\nu_s = 1/2$), we obtain a bound of $c_0 \frac{C_K}{\sqrt{\gamma_K}} \frac{t_s}{\sqrt{d}}$. Since the global minimax regret bounds the maximum over any targeted adversarial choice $j$, we acquire the sequence maximum:
    \[
    \sup_{\|\theta^\star\|_2 \le 1} \mathbb{E}[R_T\left(\textbf{Alg}, \theta^\star\right)] \ge c_0 \frac{C_K}{\sqrt{\gamma_K}} \max \left\{ \frac{t_s}{\sqrt{d}}, \sqrt{d} \frac{t_{s+1}}{\sqrt{t_s}}, \sqrt{d} \frac{t_{s+2}}{\sqrt{t_{s+1}}}, \dots, \sqrt{d} \frac{T}{\sqrt{t_{M-1}}} \right\}.
    \]
    
    Let $\alpha$ denote the maximum inside braces in the preceding display. We now bound $\alpha$ using a weighted geometric-mean argument along the batch grid. Raising each inequality to an exponentially increasing weight yields
    \begin{equation*}
    \begin{aligned}
    \alpha &\ge t_s d^{-1/2}, \\
    \alpha^{2^k} &\ge d^{2^{k-1}} t_{s+k}^{2^k} t_{s+k-1}^{-2^{k-1}} \qquad (k=1,\ldots,M-s-1), \\
    \alpha^{2^{M-s}} &\ge d^{2^{M-s-1}} T^{2^{M-s}} t_{M-1}^{-2^{M-s-1}}.
    \end{aligned}
    \end{equation*}
    Multiplying the entire cascade perfectly collapses the internal boundary values $t_k$, as the exponent of each internal $t_k$ algebraically resolves to $0$. Summing the cascading exponents of $d$ universally yields:
    \[
    \alpha^{2^{M-s+1}-1} \ge d^{2^{M-s}-1.5} T^{2^{M-s}} \implies \alpha \ge \sqrt{dT} \left( \frac{T}{d^2} \right)^{\frac{1}{2(2^{M-s+1}-1)}}.
    \]
    Noting $s \ge 1 \implies 2^{M-s+1}-1 \le 2^M-1$, re-substituting $\alpha$ with $C_K = \Omega(\sqrt{\log K})$ and $\gamma_K \le \mathcal{O}(1)$ fully captures the scale parameters, resulting in the final Minimax limit across any grid partition for general $K$:
    \[
    \sup_{\|\theta^\star\|_2 \le 1} \mathbb{E}[R_T\left(\textbf{Alg}, \theta^\star\right)] \ge c \cdot \sqrt{dT \log K} \left( \frac{T}{d^2} \right)^{\frac{1}{2(2^M-1)}}.
    \]
\halmos
\end{proof}
}
\section{Proof of Main Lemmas}\label{sec:proof-main-lemmas}
\subsection{Proof of Lemma~\ref{lemma.equator}}
{
Fix a batch $m\in[M]$, let $n_m=t_m-t_{m-1}$
and condition throughout on the pre-batch history
$\mathcal{H}_{t_{m-1}}$. Under this conditioning,
$\hat{\theta}_{m-1}$ is fixed, while the context vectors in the current
batch remain mutually independent.

Let $y_{t,a}=\Sigma^{-1/2}x_{t,a}$.
Then the collection
$
\{y_{t,a}:t_{m-1}<t\le t_m,\ a\in[K]\}
$
is mutually independent, and every $y_{t,a}$ follows
$\mathcal{N}(0,I_d)$. Define
\[
B
=
\frac{1}{n_m}
\sum_{t=t_{m-1}+1}^{t_m}
y_{t,a_t}y_{t,a_t}^{\top}.
\]

We first characterize the distribution of the selected context. Suppose
that $\hat{\theta}_{m-1}\neq 0$, and let
\[
v
=
\frac{\Sigma^{1/2}\hat{\theta}_{m-1}}
{\|\Sigma^{1/2}\hat{\theta}_{m-1}\|_2}.
\]
For each action $a\in[K]$, decompose
\[
y_{t,a}
=
S_{t,a}v+W_{t,a},
\qquad
S_{t,a}=v^\top y_{t,a},
\qquad
W_{t,a}=(I_d-vv^\top)y_{t,a}.
\]
Because the whitened contexts are mutually independent standard Gaussian
vectors, the variables $\{S_{t,a}\}_{a\in[K]}$ are independent standard
Gaussians, the vectors $\{W_{t,a}\}_{a\in[K]}$ are independent with
\[
W_{t,a}\sim\mathcal{N}(0,I_d-vv^\top),
\]
and the score vector $(S_{t,1},\ldots,S_{t,K})$ is independent of the
collection $(W_{t,1},\ldots,W_{t,K})$. Since
\[
a_t=\arg\max_{a\in[K]}S_{t,a},
\]
is determined entirely by the scores, selecting $a_t$ does not alter the
distribution of the corresponding orthogonal component. Consequently,
$W_{t,a_t}\sim\mathcal{N}(0,I_d-vv^\top)$ and is independent of
$S_{t,a_t}=\max_{a\in[K]}S_{t,a}$.
Conditional on the pre-batch history, the selected pairs are also
independent across $t$ because the context collections are independent
across time.

Apply a common orthogonal rotation that maps $v$ to $e_d$. In the
rotated coordinate system,
\[
y_{t,a_t}
\sim
\mathcal{N}(0,I_{d-1})\otimes\nu_K,
\]
where $\nu_K$ is the distribution of
\[
M_K=\max_{a\in[K]}Z_a
\]
for independent standard Gaussian variables $Z_1,\ldots,Z_K$.

If $\hat{\theta}_{m-1}=0$, the fixed context-independent tie-breaking
rule selects a fixed action. In this case,
$y_{t,a_t}\sim\mathcal{N}(0,I_d)$ independently across $t$. To treat
the two cases uniformly, write the selected whitened context in the
rotated coordinate system as
\[
y_{t,a_t}=(G_t,Q_t),
\qquad
G_t\sim\mathcal{N}(0,I_{d-1}),
\]
where $G_t$ is independent of $Q_t$ and
\[
Q_t=
\begin{cases}
M_{K,t}:=\max_{a\in[K]}Z_{t,a},
&\hat{\theta}_{m-1}\neq0,\\
Z_t,
&\hat{\theta}_{m-1}=0,
\end{cases}
\]
with $Z_{t,1},\ldots,Z_{t,K},Z_t$ independent standard Gaussian
variables. In either case,
\[
\mathbb{P}(Q_t\ge1)\ge\mathbb{P}(Z_t\ge1)>0,
\]
and the absolute tail of $Q_t$ is bounded by a union bound over at most
$K$ standard Gaussian variables.

We next establish an anti-concentration bound in an arbitrary fixed
direction. Let
$y=(G,Q)$ have the preceding product distribution, where
$G\sim\mathcal{N}(0,I_{d-1})$ is independent of $Q$, and fix a unit
vector
\[
u=(u_\perp,u_d)
\in
\mathbb{R}^{d-1}\times\mathbb{R}.
\]
We show that there exist numerical constants $c_1,c_2>0$, independent
of $(d,K)$, such that
\begin{equation}\label{eq.large_prob_fixed_u}
\mathbb{P}(|u^\top y|\ge c_1)\ge c_2.
\end{equation}

Suppose first that $|u_d|<1/2$. Conditional on $Q$, the variable
$u^\top y$ is Gaussian with mean $u_dQ$ and variance
\[
\|u_\perp\|_2^2=1-u_d^2\ge\frac34.
\]
For a Gaussian random variable, the probability of leaving a symmetric
interval is minimized when its mean is zero. Therefore,
\[
\mathbb{P}(|u^\top y|\ge c_1)
\ge
\mathbb{P}\!\left(
\left|\mathcal{N}(0,3/4)\right|\ge c_1
\right),
\]
which is bounded below by a positive numerical constant.

Now suppose that $|u_d|\ge1/2$. On the event $\{Q\ge1\}$,
\[
|u_dQ|\ge\frac12.
\]
Conditional on $Q$, the variable $u_\perp^\top G$ is a centered
Gaussian independent of $Q$. Consequently,
\[
\mathbb{P}\!\left(
|u_\perp^\top G+u_dQ|\ge\frac14
\,\middle|\,
Q
\right)
\ge\frac12
\]
on $\{Q\ge1\}$. Moreover,
\[
\mathbb{P}(Q\ge1)
\ge
\mathbb{P}(Z_1\ge1)>0.
\]
Thus, taking $c_1=1/4$, we obtain
\eqref{eq.large_prob_fixed_u} for some numerical constant $c_2>0$.

For every $t$,
\[
u^\top
y_{t,a_t}y_{t,a_t}^{\top}u
=
(y_{t,a_t}^{\top}u)^2
\ge
c_1^2
\mathbbm{1}\{|y_{t,a_t}^{\top}u|\ge c_1\}.
\]
Summing over batch $m$ and dividing by $n_m$ gives
\[
u^\top Bu
\ge
\frac{c_1^2}{n_m}
\sum_{t=t_{m-1}+1}^{t_m}
\mathbbm{1}\{|y_{t,a_t}^{\top}u|\ge c_1\}.
\]
Conditional on $\mathcal{H}_{t_{m-1}}$, the indicators in the preceding
sum are independent across $t$. Hence, by
\eqref{eq.large_prob_fixed_u} and a Chernoff bound,
\begin{equation}\label{eq.concentration_fixed_u}
\mathbb{P}\!\left(
u^\top Bu\ge\frac{c_1^2c_2}{2}
\,\middle|\,
\mathcal{H}_{t_{m-1}}
\right)
\ge
1-e^{-c_3n_m}
\end{equation}
for some numerical constant $c_3>0$.

We next upper-bound $\lambda_{\max}(B)$. In the rotated coordinate
system, write $y_{t,a_t}=v_t+w_t$, where
\[
v_t=(y_{t,a_t,1},\ldots,y_{t,a_t,d-1},0)^\top,
\qquad
w_t=(0,\ldots,0,y_{t,a_t,d})^\top.
\]
The vectors $v_t$ are independent Gaussian vectors whose first $d-1$
coordinates are independent standard Gaussians. Therefore, by standard
Wishart concentration (cf. Lemma~\ref{lemma.wishart}), with conditional
probability at least $1-e^{-\Omega(n_m)}$,
\[
\lambda_{\max}\!\left(
\frac1{n_m}
\sum_{t=t_{m-1}+1}^{t_m}
v_tv_t^\top
\right)
\le c_4
\]
for some numerical constant $c_4>0$.

For the last coordinate,
\[
w_{t,d}
\stackrel{d}{=}Q_t,
\]
where $Q_t=\max_{a\in[K]}Z_{t,a}$ in the nonzero-estimator case and
$Q_t=Z_{t,1}$ in the zero-estimator case, with
$\{Z_{t,a}\}_{a\in[K]}$ independent standard Gaussian variables. A
Gaussian tail bound and a union bound over the $K$ actions give
\[
|w_{t,d}|
\le
\sqrt{c_5\log(KT)}
\]
with probability at least $1-O(T^{-5})$ for each $t$. A union bound over
$t\in[T]$ therefore shows that, with probability at least
$1-O(T^{-4})$,
\[
\lambda_{\max}\!\left(
\frac1{n_m}
\sum_{t=t_{m-1}+1}^{t_m}
w_tw_t^\top
\right)
=
\frac1{n_m}
\sum_{t=t_{m-1}+1}^{t_m}w_{t,d}^2
\le
c_5\log(KT).
\]

Since $(a+b)(a+b)^\top
\preceq
2(aa^\top+bb^\top)$,
we have
\[
B
\preceq
2\left(
\frac1{n_m}\sum_{t=t_{m-1}+1}^{t_m}v_tv_t^\top
+
\frac1{n_m}\sum_{t=t_{m-1}+1}^{t_m}w_tw_t^\top
\right).
\]
Consequently, with conditional probability at least $1-e^{-\Omega(n_m)}-O(T^{-4})$,
we have
\begin{equation}\label{eq.lambda_max}
\lambda_{\max}(B)
\le
c_6\log(KT)
\end{equation}
for some numerical constant $c_6>0$.

Let $\mathcal{N}_d(\varepsilon)$ be an $\varepsilon$-net of the unit
sphere with
$|\mathcal{N}_d(\varepsilon)|\le(1+2/\varepsilon)^d$.
The standard net approximation
(cf. \cite[Section 2.3.1]{tao2012topics}) gives
\[
\lambda_{\min}(B)
\ge
\min_{u\in\mathcal{N}_d(\varepsilon)}
u^\top Bu
-
2\varepsilon\lambda_{\max}(B).
\]
Applying \eqref{eq.concentration_fixed_u} and a union bound over the net
gives
\[
\min_{u\in\mathcal{N}_d(\varepsilon)}
u^\top Bu
\ge
\frac{c_1^2c_2}{2}
\]
except with conditional probability at most
\[
\exp\!\left\{
d\log\left(1+\frac2\varepsilon\right)-c_3n_m
\right\}.
\]
On the event in \eqref{eq.lambda_max},
\[
\lambda_{\min}(B)
\ge
\frac{c_1^2c_2}{2}
-
2\varepsilon c_6\log(KT).
\]
Choose $\varepsilon
=
\frac{c_1^2c_2}{8c_6\log(KT)}$.
Then
\[
\lambda_{\min}(B)
\ge
\frac{c_1^2c_2}{4}.
\]
Because
\[
\log\left(1+\frac2\varepsilon\right)
=
O(\log\log(KT))
\]
and the grid in~\eqref{eq.minimax_grid} gives
$n_m=\Omega(d\sqrt{T})$, the total conditional failure probability is
$O(T^{-4})$. Thus, for some numerical constant $c_7>0$,
\[
B\succeq c_7I_d
\]
with conditional probability at least $1-O(T^{-4})$.

Finally,
\[
\frac1{n_m}
\sum_{t=t_{m-1}+1}^{t_m}
x_{t,a_t}x_{t,a_t}^\top
=
\Sigma^{1/2}B\Sigma^{1/2}
\succeq
c_7\Sigma
\succeq
c_7\frac{\kappa}{d}I_d.
\]
Multiplying by $n_m$ yields
\[
\lambda_{\min}\!\left(
\sum_{t=t_{m-1}+1}^{t_m}
x_{t,a_t}x_{t,a_t}^\top
\right)
\ge
c_7\frac{\kappa n_m}{d}.
\]

The conditional failure bound is uniform over
$\mathcal{H}_{t_{m-1}}$. Therefore, by the law of total probability, the
same $O(T^{-4})$ failure bound holds unconditionally.
}
\halmos
\subsection{Proof of Lemma~\ref{lemma.difference}}
\begin{proof}{Proof of Lemma~\ref{lemma.difference}.}
	By the standard algebra of linear regression, we have:
	\begin{align*}
	\hat{\theta}_m - \theta^\star = A_m^{-1}\sum_{t=1}^{t_m}x_{t,a_t} (r_{t,a_t} - x_{t,a_t}^\top \theta^\star). 
	\end{align*}
	Hence, conditioned on the contexts $\{x_{t,a_t}\}_{t\in [t_m]}$, the noise terms $r_{t,a_t} - x_{t,a_t}^\top \theta^\star$ are independent by the assumption, and each noise term $r_{t,a_t} - x_{t,a_t}^\top \theta^\star$ is $1$-sub-Gaussian.
	
	Next, we show that the random vector $\hat{\theta}_m - \theta^\star$ is $\sigma^2$-sub-Gaussian conditioned on the contexts with $\sigma^2 = \lambda_{\min}(A_m)^{-1}$.  To see this, we start by recalling that a centered (i.e. zero-mean) random vector $V$ is $v$-sub-Gaussian if the scalar random variable $\langle V, u\rangle$ is \StochRev{$v$-sub-Gaussian} for any unit vector $u$.
	\StochRev{Consequently, for any unit vector $u \in \mathbf{R}^d$, we have:}
	$$\langle	\hat{\theta}_m - \theta^\star , u \rangle = \langle A_m^{-1}\sum_{t=1}^{t_m}x_{t,a_t} (r_{t,a_t} - x_{t,a_t}^\top \theta^\star), u\rangle = \sum_{t=1}^{t_m} u^T A_m^{-1}x_{t,a_t} (r_{t,a_t} - x_{t,a_t}^\top \theta^\star).$$
		Since each \StochRev{term in the sum} is $(u^T A_m^{-1}x_{t,a_t})^2$-sub-Gaussian, and since all of them are independent (after being conditioned on  $\{x_{t,a_t}\}_{t\in [t_m]}$), their sum is also sub-Gaussian with the sub-Gaussian constant  equal to the sum of the \StochRev{sub-Gaussian} constants:
	\begin{align*}
	&\sum_{t=1}^{t_m}  (u^T A_m^{-1}x_{t,a_t})^2= \sum_{t=1}^{t_m}  u^T A_m^{-1}x_{t,a_t} x_{t,a_t}^T A_m^{-1} u
	= u^T A_m^{-1}\big( \sum_{t=1}^{t_m}  x_{t,a_t} x_{t,a_t}^T \big) A_m^{-1} u \\
	&= u^T A_m^{-1}A_m A_m^{-1} u =  u^T A_m^{-1} u \le \lambda_{\max}(A_m^{-1}) = \lambda_{\min}(A_m)^{-1}. 
	\end{align*}
	Since this inequality holds for every unit vector $u$, it follows that,
	conditional on the selected contexts,
	$\hat{\theta}_m-\theta^\star$ is a
	$\lambda_{\min}(A_m)^{-1}$-sub-Gaussian random vector.

	Proceeding further, by Lemma \ref{lemma.equator}, we have for each $m\in [M]$, with probability at least $1-O(T^{-4})$ 
	$\lambda_{\min}\left(\sum_{t=t_{m-1}+1}^{t_m} x_{t,a_t}x_{t,a_t}^\top \right) \ge c\cdot \frac{\kappa(t_m-t_{m-1})}{d}$. Consequently, by a union bound over all $M$ (which is at most $T$),
	we have with probability at least $1-O(T^{-3})$, $\lambda_{\min}\left(\sum_{t=t_{m-1}+1}^{t_m} x_{t,a_t}x_{t,a_t}^\top \right) \ge c\cdot \frac{\kappa(t_m-t_{m-1})}{d}$ for all $m \in [M]$.
	Since $\lambda_{\min}(X+Y)\ge \lambda_{\min}(X)+\lambda_{\min}(Y)$ for any symmetric matrices $X,Y$, 
	it then follows that with probability at least $1-O(T^{-3})$:
	\begin{align*}
	\lambda_{\min}(A_m) = \lambda_{\min}\left(\sum_{l=1}^m \sum_{t=t_{l-1}+1}^{t_l} x_{t,a_t}x_{t,a_t}^\top \right) \ge \sum_{l=1}^m \lambda_{\min}\left(\sum_{t=t_{l-1}+1}^{t_l} x_{t,a_t}x_{t,a_t}^\top \right) \ge \frac{c\kappa t_m}{d}.
	\end{align*}
	
	On the event
	$\lambda_{\min}(A_m)\ge c\kappa t_m/d$,
	the preceding argument shows that, conditional on the selected contexts,
	every one-dimensional projection of
	$\hat{\theta}_m-\theta^\star$ is sub-Gaussian with variance proxy at most
	$d/(c\kappa t_m)$. A standard net argument for sub-Gaussian random vectors
	therefore gives
	\[
	\|\hat{\theta}_m-\theta^\star\|_2
	\le
	Cd\sqrt{\frac{\log T}{\kappa t_m}}
	\]
	with conditional probability at least $1-O(T^{-3})$ over the reward noise.
	Combining this noise-concentration event with the preceding design-matrix
	event by a union bound proves the result.\halmos
\end{proof}

\section{Proofs of Propositions}
\label{appendix.proof-prop}
\subsection{Proof of Proposition~\ref{prop.heterogeneous_context}.}

We begin with a technical lemma. 
\begin{lemma}[Small-ball bound for a selected Gaussian vector]\label{lemma.selected-Gaussian}
Let $
X=(x_1,\dots,x_K)\sim \mathcal{N}(0,\Sigma)$ in $\mathbb{R}^{Kd}$,
where $
\Sigma \succeq I_{Kd}.$
Fix unit vectors \(u,v\in\mathbb{R}^d\), and let
\[
i^\star \in \operatorname*{arg\,max}_{1\le i\le K}
\langle u,x_i\rangle,
\]
with any measurable tie-breaking rule. Then there exist universal constants
\(c,c'>0\) such that
\[
\mathbb{P}\left(
\left|\langle v,x_{i^\star}\rangle\right|^2
\ge \frac{c}{\log(eK)}
\right)
\ge c'.
\]
\end{lemma}

\begin{proof}{Proof.}
\StochRev{Since $\Sigma\succeq I_{Kd}$, the matrix
$\Sigma-I_{Kd}$ is positive semidefinite. Let
\[
Y\sim\mathcal{N}(0,\Sigma-I_{Kd})
\quad\text{and}\quad
G=(g_1,\dots,g_K)\sim\mathcal{N}(0,I_{Kd})
\]
be independent. Then $Y+G\sim\mathcal{N}(0,\Sigma)$, so we may realize
$X$ as $X=Y+G$ on a common probability space.} Condition on
$
Y=(a_1,\dots,a_K).
$
It suffices to prove, uniformly over arbitrary deterministic vectors
\(a_1,\dots,a_K\in\mathbb{R}^d\), that if
$
x_i=a_i+g_i
$
and
$
I\in \operatorname*{arg\,max}_{1\le i\le K}
\langle u,x_i\rangle,
$
then
\begin{align}\label{eq:anticoncentration-target}
\mathbb{P}\left(
\left|\langle v,x_I\rangle\right|\le r
\,\middle|\,
Y
\right)
\le
C r\sqrt{\log(eK)}.
\end{align}
Note that averaging over $Y$ and choosing $r=\sqrt{\frac{c}{\log(eK)}}$ in \eqref{eq:anticoncentration-target} with a small $c>0$ proves the lemma. To prove \eqref{eq:anticoncentration-target}, we first establish a one-dimensional density estimate.

\medskip

\begin{lemma}\label{lemma.one-dimension}
Let $
S_i=m_i+Z_i, i=1,\dots,K,
$
where \(m_1,\dots,m_K\in\mathbb{R}\) are arbitrary and
\(Z_1,\dots,Z_K\) are independent standard Gaussian random variables.
\StochRev{Define
\[
M=\max_{1\le i\le K}S_i,
\qquad
I=\min\operatorname*{arg\,max}_{1\le i\le K}S_i,
\]
and let $p_i=\mathbb{P}(I=i)$. Since the $S_i$ have continuous
distributions, the maximizer is unique almost surely, so the tie-breaking
rule does not affect the result.}
Then the conditional distribution of \(M\) given \(I=i\) has a density \(f_i\)
satisfying
\[
\|f_i\|_{\infty}
\le
C_0\sqrt{\log\frac{e}{p_i}},
\]
where \(C_0>0\) is a universal constant.
\end{lemma}

\begin{proof}{Proof of Lemma \ref{lemma.one-dimension}.}
Let \(\phi\) and \(\Phi\) denote the standard Gaussian density and cumulative
distribution function, respectively, and write
$
\overline{\Phi}=1-\Phi.
$
Since ties occur with probability zero, the joint density of \(M\) and the event
\(\{I=i\}\) is
\[
h_i(t)
=
\phi(t-m_i)\prod_{j\ne i}\Phi(t-m_j).
\]
Consequently, the conditional density of \(M\) given \(I=i\) is
\StochRev{\[
f_i(t)
=
\frac{\phi(t-m_i)Q_i(t)}{p_i},
\qquad
Q_i(t):=\prod_{j\ne i}\Phi(t-m_j).
\]}
The function \(Q_i\) is nondecreasing. Therefore,
\[
\begin{aligned}
p_i
=
\int_{\mathbb{R}}\phi(s-m_i)Q_i(s)\,ds 
\ge
Q_i(t)\int_t^\infty\phi(s-m_i)\,ds =
Q_i(t)\overline{\Phi}(t-m_i).
\end{aligned}
\]
Since also \(Q_i(t)\le 1\), writing \(z=t-m_i\), we obtain
\begin{align*}
f_i(t)
\le
\min\left\{
\frac{\phi(z)}{p_i},
\frac{\phi(z)}{\overline{\Phi}(z)}
\right\}.
\end{align*}

If \(p_i>\frac{1}{2}\), then
$
f_i(t)
\le
\frac{\phi(z)}{p_i}
\le
2\phi(0).
$
\StochRev{Since $\sqrt{\log(e/p_i)}\ge1$, choosing
$C_0\ge2\phi(0)$ establishes the claimed bound in this case.} Suppose now
that \(p_i\le \frac{1}{2}\), and let \(z_i\ge 0\) be defined by
$
\overline{\Phi}(z_i)=p_i.
$
For \(z\le z_i\), the Gaussian Mills ratio gives
\[
f_i(t)
\le
\frac{\phi(z)}{\overline{\Phi}(z)}
\le
C(1+\max\{z,0\})
\le
C(1+z_i).
\]
For \(z\ge z_i\), the Gaussian Mills ratio again gives
\[
f_i(t)
\le
\frac{\phi(z)}{p_i}
\le
\frac{\phi(z_i)}{\overline{\Phi}(z_i)}
\le
C(1+z_i).
\]
Finally, since
$
z_i
\le
\sqrt{2\log\frac{1}{p_i}},
$
it follows that
\[
\|f_i\|_{\infty}
\le
C_0\sqrt{\log\frac{e}{p_i}},
\]
which proves the lemma. \halmos
\end{proof}

We now return to the vector problem in Lemma \ref{lemma.selected-Gaussian}. Write
$
v=\alpha u+\beta w,
$
where \(w\perp u\), \(\|w\|=1\), and
$
\alpha^2+\beta^2=1.
$
Define
\[
m_i=\langle u,a_i\rangle,
\qquad
b_i=\langle w,a_i\rangle,
\qquad
Z_i=\langle u,g_i\rangle,
\qquad
H_i=\langle w,g_i\rangle.
\]
Because \(u\perp w\) and \(g_1,\dots,g_K\) are independent standard Gaussian
vectors, the random variables
$
Z_1,\dots,Z_K,H_1,\dots,H_K
$
are mutually independent standard Gaussian random variables.

Set $
M=\max_{1\le i\le K}(m_i+Z_i)$. 
Then
\[
I\in \operatorname*{arg\,max}_{1\le i\le K}(m_i+Z_i),
\qquad
M=m_I+Z_I,
\]
and
\[
\begin{aligned}
\langle v,x_I\rangle
=
\alpha\langle u,a_I+g_I\rangle
+
\beta\langle w,a_I+g_I\rangle =
\alpha M+\beta b_I+\beta H_I.
\end{aligned}
\]
Moreover, \((M,I)\) is measurable with respect to
\((Z_1,\dots,Z_K)\), whereas \((H_1,\dots,H_K)\) is independent of
\((Z_1,\dots,Z_K)\). Consequently, conditional on \((M,I)\), we have 
$
H_I\sim\mathcal{N}(0,1).
$
We consider two cases.

\noindent\textbf{Case 1: \(\lvert\alpha\rvert\ge 2^{-1/2}\).} Let
$
p_i=\mathbb{P}(I=i\mid Y)
$. Conditional on \(I=i\), by Lemma \ref{lemma.one-dimension}, the density of
$
\alpha M+\beta b_i
$
is bounded by
\[
\frac{\|f_i\|_{\infty}}{|\alpha|}
\le
\sqrt{2}\,C_0\sqrt{\log\frac{e}{p_i}}.
\]
\StochRev{Let $q_i$ denote the conditional density of
$\alpha M+\beta b_i$ given $I=i$, and let $\mu$ be the distribution of
$\beta H_i$. By independence, the conditional density of
$\langle v,x_I\rangle$ given $I=i$ is $q_i*\mu$. Since $\mu$ is a
probability measure,
\[
\|q_i*\mu\|_\infty
\le \|q_i\|_\infty\mu(\mathbb{R})
=\|q_i\|_\infty.
\]
Thus, this conditional density is bounded by
$C_1\sqrt{\log(e/p_i)}$ for a universal constant $C_1$.} It follows that
\[
\begin{aligned}
\mathbb{P}\left(
|\langle v,x_I\rangle|\le r
\,\middle|\,
Y
\right)
=
\sum_{i=1}^K
p_i\,
\mathbb{P}\left(
|\langle v,x_I\rangle|\le r
\,\middle|\,
I=i,Y
\right) 
\le
2C_1r
\sum_{i=1}^K
p_i\sqrt{\log\frac{e}{p_i}}.
\end{aligned}
\]
\StochRev{Applying Jensen's inequality twice gives}
\[
\sum_{i=1}^K
p_i\sqrt{\log\frac{e}{p_i}}
\le
\sqrt{
\sum_{i=1}^K
p_i\log\frac{e}{p_i}
} \le \sqrt{\log(eK)}.
\]
Combining the above displays yields \eqref{eq:anticoncentration-target}. 

\noindent\textbf{Case 2: \(\lvert\alpha\rvert<2^{-1/2}\).} In this case,
$
|\beta|\ge 2^{-1/2}.
$
Since the density of $\beta H_I$ is upper bounded by a universal constant \(C_2\), and convolution cannot increase the $L^\infty$-norm, 
we get
\[
\mathbb{P}\left(
|\langle v,x_I\rangle|\le r
\,\middle|\,
Y
\right)
\le
2C_2r.
\]
Since \(\log(eK)\ge 1\), this implies \eqref{eq:anticoncentration-target} as well. \halmos 
\end{proof}

With Lemma \ref{lemma.selected-Gaussian}, we are now able to establish a counterpart of Lemma \ref{lemma.equator}.

\begin{lemma}[Selected design under heterogeneous Gaussian contexts]
\label{lemma.heterogeneous_design}
Under Assumption~\ref{assumption:heterogeneous_cov}, fix a batch $m$ and
condition on $\mathcal{H}_{t_{m-1}}$. Suppose that
\StochRev{$\hat{\theta}_{m-1}$ remains fixed throughout the batch} and that ties in the
greedy rule are resolved according to a fixed rule independent of the current
contexts. For the grid in~\eqref{eq.minimax_grid}, with
$M=O(\log\log T)$, there is a numerical constant $c>0$ such that, with
conditional probability at least $1-O(T^{-4})$,
\[
\lambda_{\min}\!\left(
\sum_{t=t_{m-1}+1}^{t_m}x_{t,a_t}x_{t,a_t}^{\top}
\right)
\ge
c\,\frac{\kappa(t_m-t_{m-1})}{d\log K}.
\]
\end{lemma}

\begin{proof}{Proof of Lemma~\ref{lemma.heterogeneous_design}.}
If $\hat{\theta}_{m-1}=0$, the fixed tie-breaking rule selects a fixed action,
and the claim follows from standard Gaussian sample-covariance concentration.
Suppose henceforth that $\hat{\theta}_{m-1}\ne0$. By Lemma \ref{lemma.selected-Gaussian} \StochRev{applied with $u$ chosen as the direction of} $\hat{\theta}_{m-1}$, for every unit vector $v\in \bR^d$ we have
\begin{align}\label{eq:result-1}
\bP\left( |\langle v, x_{t,a_t} \rangle|^2 \ge \frac{c\kappa}{d\log(eK)} \right) \ge c'.  
\end{align}
In addition, since $\Sigma \preceq \frac{1}{d}I_{Kd}$, the covariance matrix $\Sigma_a$ of each $x_{t,a}$ satisfies $\Sigma_a\preceq \frac{1}{d}I_d$. By Gaussian concentration and a union bound over $a\in [K]$, with probability $1-T^{-5}$, it holds that $\|x_{t,a}\|_2^2 \le C\log(KT)$ for every $a\in [K]$. By an additional union bound over $t$, with probability at least $1-T^{-4}$,
\begin{align}\label{eq:result-2}
\lambda_{\max}\left(\frac{1}{n_m}\sum_{t=t_{m-1}+1}^{t_m} x_{t,a_t}x_{t,a_t}^\top \right) \le C\log(KT). 
\end{align}
Now the claimed lemma follows from the same $\varepsilon$-net argument in the proof of Lemma \ref{lemma.equator} with \eqref{eq:result-1} and \eqref{eq:result-2}. 
\halmos
\end{proof}

\begin{proof}{Proof of Proposition~\ref{prop.heterogeneous_context}.}
	Take the $m$-th batch for any $m \ge 2$, along with any time step $t$ within it. From the definition of $a_t$, it follows that $x_{t,a_t}^\top \hat{\theta}_{m-1} \ge x_{t,a}^\top \hat{\theta}_{m-1}$ for all $a \in [K]$. Thus,
	\begin{equation*}
	\begin{aligned}
	\max_{a \in [K]} (x_{t,a} - x_{t,a_t})^\top \theta^\star &\le \max_{a \in [K]} (x_{t,a} - x_{t,a_t})^\top (\theta^\star - \hat{\theta}_{m-1}) \\
	&\le \max_{a, a' \in [K]} (x_{t,a} - x_{t,a'})^\top (\theta^\star - \hat{\theta}_{m-1}) \\
	&\le 2 \max_{a \in [K]} |x_{t,a}^\top (\theta^\star - \hat{\theta}_{m-1})|.
	\end{aligned}
	\end{equation*}
	For each $a\in[K]$, $x_{t,a}\sim\mathcal{N}(0,\Sigma_a)$ and is
	independent of $\hat{\theta}_{m-1}$. Consequently, by conditioning on earlier contexts and rewards, it follows that $x_{t,a}^\top (\theta^\star - \hat{\theta}_{m-1}) \sim \mathcal{N}(0, \sigma_a^2)$, where
	\begin{equation*}
	\sigma_a^2 = (\theta^\star - \hat{\theta}_{m-1})^\top \Sigma_a (\theta^\star - \hat{\theta}_{m-1}) \le \lambda_{\max}(\Sigma_a) \|\theta^\star - \hat{\theta}_{m-1}\|_2^2 \le \frac{\|\theta^\star - \hat{\theta}_{m-1}\|_2^2}{d}.
	\end{equation*}
	Applying a Gaussian tail bound and a union bound over $a\in[K]$ \StochRev{gives}
	with probability at least $1-O(T^{-3})$,
	\[
	\max_{a\in[K]} (x_{t,a} - x_{t,a_t})^\top \theta^\star \le 2 \max_{a\in[K]} |x_{t,a}^\top (\theta^\star - \hat{\theta}_{m-1})| = O \left( \|\theta^\star - \hat{\theta}_{m-1}\|_2 \cdot \sqrt{\frac{\log(KT)}{d}} \right).
	\]
	{Combining Lemmas~\ref{lemma.heterogeneous_design}
	and~\ref{lemma.difference} and applying a union bound, we find that there
	exists a numerical constant $C'>0$ such that, with probability at least
	$1-O(T^{-3})$, the instantaneous regret at time $t$ is bounded by}
	\[
	\max_{a\in[K]} (x_{t,a} - x_{t,a_t})^\top \theta^\star \le C' \sqrt{\log(KT) \log T} \cdot \sqrt{\frac{d\log K}{\kappa t_{m-1}}}.
	\]
	Applying a union bound over $t \in [T]$, we find the cumulative regret after the initial batch is at most:
	\begin{equation}
	\label{eq.prop.heterogeneous.later_batch}
	\sum_{m=2}^M C' \sqrt{\log(KT) \log T} \cdot t_m \sqrt{\frac{d\log K}{\kappa t_{m-1}}} \le C'' \sqrt{\frac{\log(KT) \log K\log T}{\kappa}} M \cdot \beta\sqrt{d}
	\end{equation}
	with probability $1 - O(T^{-2})$, where the inequality follows from the grid definition in \eqref{eq.minimax_grid}.
	
	Concerning the first batch, the immediate regret at any time $t$ is no more than the maximum of $K$ Gaussian random variables $\mathcal{N}(0, (\theta^\star)^\top \Sigma_a \theta^\star)$. Since $\|\theta^\star\|_2 \le 1$ and $\lambda_{\max}(\Sigma_a) \le 1/d$, the instantaneous regret is at most $C'''\sqrt{\log(KT)/d}$ for some constant $C''' > 0$ with probability at least $1 - O(T^{-3})$. 	Using a union bound for $t \in [t_1]$, the total regret during the first batch is bounded by:
	\begin{equation}
	\label{eq.prop.heterogeneous.first_batch}
	C''' \sqrt{\log(KT)/d} \cdot t_1 = C''' \sqrt{\log(KT)} \cdot \beta\sqrt{d}.
	\end{equation}
	The desired regret bound follows with high probability by combining \eqref{eq.prop.heterogeneous.later_batch}, \eqref{eq.prop.heterogeneous.first_batch}, and the choice of $\beta$ in Algorithm~\ref{algo.pure-exp} (noting $M = O(\log \log T)$), which also holds in expectation.\halmos
	\end{proof}

\ZH{

\subsection{Proof of Proposition~\ref{prop.action_dependent}.}
	\begin{proof}{Proof of Proposition~\ref{prop.action_dependent}.}
		We first adapt Lemma~\ref{lemma.equator} to show that every action's covariance matrix grows linearly. For a fixed action $a \in [K]$ and a fixed unit vector $u \in \mathbb{R}^d$, we analyze the expected projection of the selected contexts. Let $Z_{-a} = \max_{a' \neq a} x_{t,a'}^\top \hat{\theta}_{m-1, a'}$ represent the maximum estimated reward among all other arms. Because the contexts $x_{t,a'}$ for $a' \neq a$ are drawn independently of $x_{t,a}$, $Z_{-a}$ is a random variable strictly independent of $x_{t,a}$.
		
		Notice that for each $a' \neq a$, the inner product $x_{t,a'}^\top \hat{\theta}_{m-1, a'}$ is a zero-mean Gaussian random variable (or exactly 0 if the estimate is the zero vector). Thus, the probability that it is less than or equal to 0 is at least $1/2$. By independence across actions:
		\[
		\mathbb{P}(Z_{-a} \le 0) = \prod_{a' \neq a} \mathbb{P}\left(x_{t,a'}^\top \hat{\theta}_{m-1, a'} \le 0\right) \ge \left(\frac{1}{2}\right)^{K-1}.
		\]
		Let $E$ denote the event $\{Z_{-a} \le 0\}$. This event is completely independent of $x_{t,a}$. For action $a$ to be selected ($a_t = a$), it is sufficient that $E$ occurs and simultaneously $x_{t,a}^\top \hat{\theta}_{m-1, a} \ge 0$. We must also ensure that the selected context has a sufficiently large projection along the arbitrary direction $u$. 
		
		Because \StochRev{$x_{t,a} \sim \mathcal{N}(0, \Sigma_a)$ is symmetric with $\lambda_{\min}(\Sigma_a) \ge \kappa/d$}, standard Gaussian anti-concentration guarantees the existence of absolute constants $c_1, p_1 > 0$ (depending only on $\kappa$) such that:
		\[
		\mathbb{P}\left( |u^\top x_{t,a}| \ge \frac{c_1}{\sqrt{d}} \quad \text{and} \quad x_{t,a}^\top \hat{\theta}_{m-1, a} \ge 0 \right) \ge p_1.
		\]
		Because $E$ is independent of $x_{t,a}$, the joint probability that action $a$ is selected and provides a large projection along $u$ is strictly bounded away from zero:
		\[
		\mathbb{P}\left( a_t = a \text{ and } (u^\top x_{t,a})^2 \ge \frac{c_1^2}{d} \right) \ge \mathbb{P}(E) \cdot p_1 \ge \left(\frac{1}{2}\right)^{K-1} p_1 \triangleq p_0 > 0.
		\]
		Crucially, $p_0$ is a strictly positive constant independent of the current parameter estimates. The expected contribution to the covariance matrix in direction $u$ at each step is at least $p_0 c_1^2 / d$. 	By applying a standard Chernoff bound over the batch length $t_m - t_{m-1}$ to establish concentration for this fixed direction $u$, bounding the maximum eigenvalue via Wishart matrix concentration, and taking a union bound over an $\varepsilon$-net on the unit sphere, we obtain the uniform lower bound.
		With probability at least $1 - O(T^{-4})$:
		\[
		\lambda_{\min}\left(\sum_{t=t_{m-1}+1}^{t_m} \mathbbm{1}(a_t = a) x_{t,a} x_{t,a}^\top\right) \ge c' \left(\frac{1}{2}\right)^{K} \cdot \frac{\kappa}{d} (t_m - t_{m-1}) \quad \forall a \in [K],
		\]
		for some absolute numerical constant $c' > 0$.
		
		Given the uniform lower bound on the minimum eigenvalue, standard least-squares analysis (analogous to Lemma~\ref{lemma.difference}, assuming conditional independence enforced by the master sample-splitting algorithm) ensures that the estimation error for every action decays optimally.
		With probability at least $1 - O(T^{-3})$:
		\[
		\forall a \in [K], \quad \|\hat{\theta}_{m-1,a} - \theta^\star_a\|_2 \le C \cdot 2^{K/2} d \sqrt{\frac{\log T}{\kappa t_{m-1}}}.
		\]

		We now bound the instantaneous regret at time $t$ in batch $m \ge 2$. Let $a^\star = \arg\max_{a} x_{t,a}^\top \theta^\star_a$ be the true optimal action, and $a_t = \arg\max_{a} x_{t,a}^\top \hat{\theta}_{m-1,a}$ be the action chosen by the greedy policy. The instantaneous regret is:
		\begin{align*}
		r_{t,a^\star} - r_{t,a_t} &= x_{t,a^\star}^\top \theta^\star_{a^\star} - x_{t,a_t}^\top \theta^\star_{a_t} \\
		&\le x_{t,a^\star}^\top \theta^\star_{a^\star} - x_{t,a_t}^\top \theta^\star_{a_t} + \left( x_{t,a_t}^\top \hat{\theta}_{m-1,a_t} - x_{t,a^\star}^\top \hat{\theta}_{m-1,a^\star} \right) \\
		&= \left( x_{t,a^\star}^\top \theta^\star_{a^\star} - x_{t,a^\star}^\top \hat{\theta}_{m-1,a^\star} \right) + \left( x_{t,a_t}^\top \hat{\theta}_{m-1,a_t} - x_{t,a_t}^\top \theta^\star_{a_t} \right) \\
		&\le 2 \max_{a \in [K]} |x_{t,a}^\top (\hat{\theta}_{m-1,a} - \theta^\star_a)|,
		\end{align*}
		where the first inequality critically uses $x_{t,a^\star}^\top \hat{\theta}_{m-1,a^\star} \le x_{t,a_t}^\top \hat{\theta}_{m-1,a_t}$ (by definition of the greedy choice). Because $x_{t,a}$ is independent of $\hat{\theta}_{m-1,a}$, the term $x_{t,a}^\top (\hat{\theta}_{m-1,a} - \theta^\star_a)$ is a zero-mean Gaussian with variance at most $\|\hat{\theta}_{m-1,a} - \theta^\star_a\|_2^2 / d$. Applying a union bound over the $K$ actions and substituting \StochRev{the preceding estimation-error bound}, the instantaneous regret is bounded by:
		\[
		r_{t,a^\star} - r_{t,a_t} \le O \left( \max_{a \in [K]} \|\hat{\theta}_{m-1,a} - \theta^\star_a\|_2 \cdot \sqrt{\frac{\log(KT)}{d}} \right) \le C' \cdot 2^{K/2} \sqrt{\log(KT) \log T} \cdot \sqrt{\frac{d}{\kappa t_{m-1}}}.
		\]

		Remarkably, this algebraic bound mirrors the instantaneous regret bound derived for the shared-parameter setting, except for the additional exponential dependence on the number of actions. Summing this instantaneous regret over the batches $m=2, \dots, M$ using the geometric grid $t_m = \lfloor \beta\sqrt{t_{m-1}}\rfloor$ yields the cumulative regret:
		\[
		\sum_{m=2}^M C' \cdot 2^{K/2} \sqrt{\log(KT) \log T} \cdot t_m \sqrt{\frac{d}{\kappa t_{m-1}}} \le C'' \cdot 2^{K/2} \sqrt{\frac{\log(KT) \log T}{\kappa}} M \cdot \beta\sqrt{d}.
		\]
		Bounding the initial batch trivially by $O(t_1/\sqrt{d}) = O(\beta\sqrt{d})$ and substituting the optimal parameter $\beta = \Theta\left(\sqrt{T} \cdot \left(\frac{T}{d^2}\right)^{\frac{1}{2(2^M-1)}}\right)$ recovers the stated upper bound.\end{proof}
}
\ZH{\subsection{Proof of Proposition~\ref{prop.general_distribution}.}
\begin{proof}{Proof of Proposition~\ref{prop.general_distribution}.}
		We use the conditional diversity condition (Condition~2 of Assumption~\ref{assumption:general_distribution}) to obtain lower bounds on the covariance of the observed contexts along arbitrary directions.
		
		Let $B_m = t_m - t_{m-1}$ be the batch length. At time $t$, the greedy algorithm selects $a_t = \arg\max_a x_{t,a}^\top \hat{\theta}_{m-1}$. Let $v = \hat{\theta}_{m-1} / \|\hat{\theta}_{m-1}\|_2$ be the normalized greedy direction. We define the filtration $\mathcal{F}_t = \sigma\left( \{x_{t,a}^\top v\}_{a=1}^K \right)$. Conditioned on the parallel projections $\mathcal{F}_t$, the greedy action index $a_t$ is deterministically fixed.
		
			For \StochRev{any unit vector} $u \in \mathbb{S}^{d-1}$, we orthogonally decompose it as $u = \alpha v + \sqrt{1-\alpha^2} u_\perp$, where $u_\perp \perp v$ and $\alpha \in [-1, 1]$. We evaluate the expected energy of the selected context $x_{t,a_t}$ along $u$ conditioned on $\mathcal{F}_t$. Because $a_t$ is fixed under $\mathcal{F}_t$, the expectation distributes directly over the selected arm:
		\begin{align*}
		\mathbb{E}\left[ (u^\top x_{t,a_t})^2 \;\middle|\; \mathcal{F}_t \right] &= \alpha^2 (v^\top x_{t,a_t})^2 + (1-\alpha^2) \mathbb{E}\left[ (u_\perp^\top w_{t,a_t})^2 \;\middle|\; \mathcal{F}_t \right] \\
		&\quad + 2\alpha\sqrt{1-\alpha^2} (v^\top x_{t,a_t}) \mathbb{E}\left[ u_\perp^\top w_{t,a_t} \;\middle|\; \mathcal{F}_t \right].
		\end{align*}
		Crucially, because contexts are drawn independently across arms, conditioning on the parallel projections $\mathcal{F}_t$ provides no information about the orthogonal component of arm $a_t$ other than what is correlated through $x_{t,a_t}^\top v$. By Condition 2, the conditional mean is exactly zero, causing the cross-term to vanish: $\mathbb{E}[u_\perp^\top w_{t,a_t} \mid \mathcal{F}_t] = 0$. 
		
		Similarly, Condition 2 guarantees the orthogonal variance is bounded away from zero: $\mathbb{E}[(u_\perp^\top w_{t,a_t})^2 \mid \mathcal{F}_t] \ge \frac{c_1}{d}$. Thus, the conditional energy satisfies:
		\[
		\mathbb{E}\left[ (u^\top x_{t,a_t})^2 \;\middle|\; \mathcal{F}_t \right] \ge \alpha^2 (v^\top x_{t,a_t})^2 + (1-\alpha^2) \frac{c_1}{d}.
		\]
		Taking the unconditional expectation, we evaluate the parallel projection. The greedy selection explicitly maximizes the projection, so its expected energy is lower bounded by the energy of any single generic arm: $\mathbb{E}[(v^\top x_{t,a_t})^2] \ge \mathbb{E}[(v^\top x_{t,1})^2] \ge \lambda_{\min}(\Sigma) \ge \frac{\kappa}{d}$.
		
		Combining these yields a uniform, unconditional lower bound for the energy in \emph{any} direction $u$:
		\[
		\mathbb{E}\left[ (u^\top x_{t,a_t})^2 \right] \ge \alpha^2 \frac{\kappa}{d} + (1-\alpha^2) \frac{c_1}{d} \ge \frac{\min(\kappa, c_1)}{d}.
		\]
		Let $c' = \min(\kappa, c_1)$. The resulting lower bound is uniform over directions $u \in \mathbb{S}^{d-1}$. Because the greedy algorithm relies on $\hat{\theta}_{m-1}$ computed from previous batches, the contexts within batch $m$ remain conditionally independent, allowing us to apply the Matrix Chernoff bound. With probability at least $1 - \mathcal{O}(T^{-4})$, the minimum eigenvalue for batch $m$ is bounded below by
		\[
		\lambda_{\min}(A_m) \ge \frac{c'}{2d} B_m.
		\]
	
		Given the uniform bound $\lambda_{\min}\left(\sum_{l=1}^{m-1} A_l\right) \ge \frac{c'}{2d} t_{m-1}$, the unregularized OLS estimation error $\hat{\theta}_{m-1} - \theta^\star$ is a sub-Gaussian random vector with variance proxy bounded by $\mathcal{O}(d / (c' t_{m-1}))$. Applying the standard sub-exponential tail bound for the squared $\ell_2$-norm, we tightly recover the estimation error with probability at least $1 - \mathcal{O}(T^{-3})$:
		\[
		\|\hat{\theta}_{m-1} - \theta^\star\|_2 \le C d \sqrt{\frac{\log T}{t_{m-1}}}.
		\]
		
		For any time step $t$ within batch $m \ge 2$, let $a^\star = \arg\max_{a} x_{t,a}^\top \theta^\star$. The instantaneous regret is algebraically bounded by:
		\begin{align*}
		x_{t,a^\star}^\top \theta^\star - x_{t,a_t}^\top \theta^\star &\le \left( x_{t,a^\star}^\top \theta^\star - x_{t,a^\star}^\top \hat{\theta}_{m-1} \right) + \left( x_{t,a_t}^\top \hat{\theta}_{m-1} - x_{t,a_t}^\top \theta^\star \right) \\
		&\le 2 \max_{a \in [K]} |x_{t,a}^\top (\theta^\star - \hat{\theta}_{m-1})|.
		\end{align*}
		Conditioned on $\hat{\theta}_{m-1}$, each variable $Y_a = x_{t,a}^\top (\theta^\star - \hat{\theta}_{m-1})$ is a zero-mean sub-Gaussian with a variance proxy bounded by $\mathcal{O}\left(\frac{1}{d} \|\hat{\theta}_{m-1} - \theta^\star\|_2^2\right)$. Because we take the maximum over $K$ independent sub-Gaussian variables, we naturally incur the optimal $\sqrt{\log K}$ penalty:
		\[
		\mathbb{P} \left( \max_{a \in [K]} |Y_a| > \sigma \sqrt{2 \log(2K/\delta)} \right) \le \delta.
		\]
		Setting $\delta = T^{-3}$ and plugging in the estimation error bound above, the instantaneous regret is bounded with high probability by:
		\[
		\mathcal{O} \left( \|\hat{\theta}_{m-1} - \theta^\star\|_2 \cdot \sqrt{\frac{\log(KT)}{d}} \right) \le C'' \sqrt{\log(KT) \log T} \cdot \sqrt{\frac{d}{t_{m-1}}}.
		\]
		
		Summing this instantaneous regret bound over the length of each batch using the geometric grid $t_m = \lfloor \beta\sqrt{t_{m-1}} \rfloor$ yields:
		\[
		\sum_{m=2}^M C'' \sqrt{\log(KT) \log T} \cdot t_m \sqrt{\frac{d}{t_{m-1}}} \le C''' \sqrt{\log(KT) \log T} \cdot M \cdot \beta \sqrt{d}.
		\]
		By setting the initial grid scaling parameter $\beta = \Theta\left(\sqrt{T} \cdot (T/d^2)^{\frac{1}{2(2^M-1)}}\right)$ and bounding the pure exploration regret in the first batch trivially by $\mathcal{O}(\sqrt{\log K / d})$, we obtain the cumulative regret bound
		\[
		\tilde{\mathcal{O}} \left( \sqrt{dT \log K} \left(\frac{T}{d^2}\right)^{\frac{1}{2(2^M-1)}} \right).
		\]
\halmos
	\end{proof}		
}

\section{Proof of Theorem~\ref{thm.problem-dependent}}
\label{appendix.problem-dependent}
\subsection{Regret Analysis for the Upper Bound}
\ZH{
\begin{lemma}\label{lemma.difference2}
	For each $m\in [M]$, let the estimator $\hat{\theta}_m$ be computed using the sample-split dataset $T^{(m)}$ constructed by partitioning each batch into $M$ equal intervals. The regression matrix is defined as $A_m = \sum_{t \in T^{(m)}} x_{t,a_t}x_{t,a_t}^\top$. Because each batch is split evenly, the sample size is $|T^{(m)}| = t_m / M$. 
    
    If the rewards are mutually independent given the selected contexts, then with probability at least $1-O(T^{-3})$,
	\begin{align*}
	\|\hat{\theta}_m - \theta^\star\|_2 \le Cd\cdot \sqrt{\frac{M \log T}{\kappa t_m}}
	\end{align*}
	for a numerical constant $C>0$ independent of $(K,T,d,m,\kappa)$.
\end{lemma}
\begin{proof}{Proof of Lemma~\ref{lemma.difference2}.}
	By the standard algebra of linear regression over the sample-split dataset $T^{(m)}$, we have:
	\begin{align*}
	\hat{\theta}_m - \theta^\star = A_m^{-1}\sum_{t \in T^{(m)}}x_{t,a_t} (r_{t,a_t} - x_{t,a_t}^\top \theta^\star). 
	\end{align*}
	Conditioned on the selected contexts $\{x_{t,a_t}\}_{t \in T^{(m)}}$, the noise terms $r_{t,a_t} - x_{t,a_t}^\top \theta^\star$ are independent by assumption, and each is $1$-sub-Gaussian.
	
		Using the identical sub-Gaussian trace argument from the original non-split analysis, the random vector $\hat{\theta}_m - \theta^\star$ is conditionally $\sigma^2$-sub-Gaussian with variance proxy $\sigma^2 = \lambda_{\min}(A_m)^{-1}$. To see this, for \StochRev{any unit vector} $u \in \mathbf{R}^d$:
	\begin{align*}
	\sum_{t \in T^{(m)}}  (u^\top A_m^{-1}x_{t,a_t})^2 &= u^\top A_m^{-1}\left( \sum_{t \in T^{(m)}}  x_{t,a_t} x_{t,a_t}^\top \right) A_m^{-1} u \\
	&= u^\top A_m^{-1}A_m A_m^{-1} u =  u^\top A_m^{-1} u \le \lambda_{\min}(A_m)^{-1}. 
	\end{align*}
	
	Proceeding further, by our context diversity assumptions and sequential independence via sample-splitting, Lemma \ref{lemma.equator} guarantees that the minimum eigenvalue grows linearly with the number of samples. Because the dataset $T^{(m)}$ is constructed by uniformly partitioning the intervals into $M$ splits, the effective sample size is exactly $|T^{(m)}| = t_m / M$. 
	
	Applying the matrix Chernoff bound uniformly over the split dataset guarantees that, with probability at least $1-O(T^{-3})$, the minimum eigenvalue is bounded below by:
	\begin{align*}
	\lambda_{\min}(A_m) = \lambda_{\min}\left(\sum_{t \in T^{(m)}} x_{t,a_t}x_{t,a_t}^\top \right) \ge \frac{c \kappa |T^{(m)}|}{d} = \frac{c \kappa t_m}{M d}.
	\end{align*}
	
	Finally, substituting this eigenvalue bound, the vector $\hat{\theta}_m - \theta^\star$ obeys a $\left(\frac{M d}{c \kappa t_m}\right)$-sub-Gaussian tail limit. Consequently, the squared norm $\|\hat{\theta}_m - \theta^\star\|_2^2$ behaves as a sub-exponential random variable. 
	
	Conditioned on the high-probability eigenvalue event for the contexts, standard sub-exponential concentration secures the claimed upper bound on the norm with a further failure probability bounded by $O(T^{-3})$ over the rewards. Taking a union bound successfully completes the proof.
	\halmos
\end{proof}

\begin{proof}{Proof of the Upper Bound in Theorem~\ref{thm.problem-dependent}.}
{
Let $Z_1,\ldots,Z_K$ be independent standard Gaussian variables and define
\[
C_K=\mathbb{E}\!\left[\max_{a\in[K]}Z_a\right],
\qquad
\widetilde C_K=
\mathbb{E}\!\left[\left(\max_{a\in[K]}Z_a\right)^2\right].
\]
Then $C_K=\Theta(\sqrt{\log K})$ and $\widetilde C_K\ge1$ for $K\ge2$.
Fix $t$ in batch $m\ge2$ and condition on the full pre-batch history
$\mathcal{H}_{t_{m-1}}$. Under this conditioning,
$\hat{\theta}_{m-1}$ is fixed, whereas the current contexts remain mutually
independent across actions and independent of the history.

Define
\[
\tilde{x}_{t,a}=\Sigma^{-1/2}x_{t,a},
\qquad
\tilde{\theta}^{\star}=\Sigma^{1/2}\theta^\star,
\qquad
\hat{\tilde{\theta}}_{m-1}=\Sigma^{1/2}\hat{\theta}_{m-1}.
\]
The vectors $\{\tilde{x}_{t,a}\}_{a\in[K]}$ are independent
$\mathcal{N}(0,I_d)$ vectors. Moreover,
$x_{t,a}^\top\theta^\star
=\tilde{x}_{t,a}^\top\tilde{\theta}^\star$ and
$x_{t,a}^\top\hat{\theta}_{m-1}
=\tilde{x}_{t,a}^\top\hat{\tilde{\theta}}_{m-1}$, so the transformation
preserves both rewards and greedy action selection. If
$\hat{\tilde{\theta}}_{m-1}\ne0$, let
$v=\hat{\tilde{\theta}}_{m-1}/\|\hat{\tilde{\theta}}_{m-1}\|_2$ and write
\[
\tilde{x}_{t,a}=S_{t,a}v+W_{t,a},
\qquad
S_{t,a}=v^\top\tilde{x}_{t,a}.
\]
Mutual independence and Gaussian orthogonality imply that the scores
$\{S_{t,a}\}_{a\in[K]}$ are independent standard Gaussians and are jointly
independent of all orthogonal components
$\{W_{t,a}\}_{a\in[K]}$. Since $a_t$ is determined only by the scores,
\[
S_{t,a_t}\stackrel{d}{=}\max_{a\in[K]}Z_a,
\qquad
W_{t,a_t}\sim\mathcal{N}(0,I_d-vv^\top),
\]
and these two variables are independent. Consequently,
\[
\mathbb{E}\!\left[
\tilde{x}_{t,a_t}\tilde{x}_{t,a_t}^\top
\mid\mathcal{H}_{t_{m-1}}
\right]
=
\widetilde C_Kvv^\top+(I_d-vv^\top)
\succeq I_d.
\]
If $\hat{\tilde{\theta}}_{m-1}=0$, the fixed context-independent
tie-breaking rule selects a fixed action, so the same conditional expectation
equals $I_d$. Lemma~\ref{lemma.equator}, applied to the sample-split
observations, therefore gives the required high-probability lower bound on
the whitened design matrix.

Let $a_t^\star$ denote the optimal action. Independence across actions also
gives
\[
\mathbb{E}\!\left[
\tilde{x}_{t,a_t^\star}^{\top}\tilde{\theta}^\star
\mid\mathcal{H}_{t_{m-1}}
\right]
=C_K\|\tilde{\theta}^\star\|_2.
\]
For the greedy action, the same score--orthogonal decomposition yields
\[
\mathbb{E}\!\left[
\tilde{x}_{t,a_t}^{\top}\tilde{\theta}^\star
\mid\mathcal{H}_{t_{m-1}}
\right]
=C_KG_{m-1},
\]
where
\[
G_{m-1}
=
\begin{cases}
\displaystyle
\frac{\tilde{\theta}^{\star\top}\hat{\tilde{\theta}}_{m-1}}
{\|\hat{\tilde{\theta}}_{m-1}\|_2},
& \hat{\tilde{\theta}}_{m-1}\ne0,\\[1.2ex]
0,&\hat{\tilde{\theta}}_{m-1}=0.
\end{cases}
\]
Thus the exact conditional expected regret is
\begin{equation}
\mathbb{E}\!\left[r_t\mid\mathcal{H}_{t_{m-1}}\right]
=
C_K\left(\|\tilde{\theta}^\star\|_2-G_{m-1}\right).
\label{eq:exact_geometry_whitened}
\end{equation}
}

By the Law of Total Expectation, we decompose the unconditional expected instantaneous regret using a high-probability "good" event. Let $\mathcal{E}_{m-1}$ be the event where the sample-split whitened design matrix tightly concentrates ($\tilde{A}_{m-1} \succeq c \frac{t_{m-1}}{M} I_d$) and the sub-Gaussian noise is uniformly bounded. 

By Lemma \ref{lemma.difference2}, the failure probability is $\mathbb{P}(\mathcal{E}_{m-1}^c) \le \mathcal{O}(T^{-3})$. Because the expected maximum single-step squared reward is bounded by $\mathcal{O}(\log K)$ for Gaussian contexts, applying the Cauchy-Schwarz inequality bounds the expected instantaneous regret of the failure event by $\mathbb{E}[r_t \mathbf{1}_{\mathcal{E}_{m-1}^c}] \le \sqrt{\mathbb{E}[r_t^2]\mathbb{P}(\mathcal{E}_{m-1}^c)} = \mathcal{O}(\sqrt{\log K \cdot T^{-3}})$. Summed over the horizon, this contributes a negligible $\mathcal{O}(T^{-0.5})$ to the total expected regret. 

Thus, we focus on the expected regret under the good event, $\mathbb{E}[r_t \mathbf{1}_{\mathcal{E}_{m-1}}]$. We securely bound the trigonometric difference in the whitened space by splitting into two exhaustive sub-cases based on the magnitude of the estimation error:

\textit{Case 1: Small Error ($\|\tilde{\theta}^\star - \hat{\tilde{\theta}}_{m-1}\|_2 \le \frac{1}{2}\|\tilde{\theta}^\star\|_2$).} 
By the reverse triangle inequality, $\|\hat{\tilde{\theta}}_{m-1}\|_2 \ge \frac{1}{2}\|\tilde{\theta}^\star\|_2$. For any two vectors $x,y$, expanding the squared Euclidean distance yields $\|x - y\|_2^2 = \|x\|_2^2 + \|y\|_2^2 - 2x^\top y$. By the AM-GM inequality, $2\|x\|_2\|y\|_2 \le \|x\|_2^2 + \|y\|_2^2$. Substituting this into our expansion gives $2\|x\|_2\|y\|_2 - 2x^\top y \le \|x - y\|_2^2$. Dividing by $2\|y\|_2$ yields the elementary algebraic inequality:
\begin{equation*}
    \|x\|_2 - \frac{x^\top y}{\|y\|_2} \le \frac{\|x - y\|_2^2}{2\|y\|_2}.
\end{equation*}
Substituting $x = \tilde{\theta}^\star$ and $y = \hat{\tilde{\theta}}_{m-1}$, and applying the bound $\|\hat{\tilde{\theta}}_{m-1}\|_2 \ge \frac{1}{2}\|\tilde{\theta}^\star\|_2$, we rigorously obtain:
\begin{equation*}
    \left(\|\tilde{\theta}^\star\|_2 - \frac{\tilde{\theta}^{\star\top} \hat{\tilde{\theta}}_{m-1}}{\|\hat{\tilde{\theta}}_{m-1}\|_2}\right) \mathbf{1}_{\mathcal{E}_{m-1}} \le \frac{\|\tilde{\theta}^\star - \hat{\tilde{\theta}}_{m-1}\|_2^2}{2\|\hat{\tilde{\theta}}_{m-1}\|_2} \mathbf{1}_{\mathcal{E}_{m-1}} \le \frac{\|\tilde{\theta}^\star - \hat{\tilde{\theta}}_{m-1}\|_2^2}{\|\tilde{\theta}^\star\|_2} \mathbf{1}_{\mathcal{E}_{m-1}}.
\end{equation*}

\textit{Case 2: Large Error ($\|\tilde{\theta}^\star - \hat{\tilde{\theta}}_{m-1}\|_2 > \frac{1}{2}\|\tilde{\theta}^\star\|_2$).}
{If $\hat{\tilde{\theta}}_{m-1}\ne0$, the geometric difference is at most
$2\|\tilde{\theta}^\star\|_2$. If
$\hat{\tilde{\theta}}_{m-1}=0$, it equals
$\|\tilde{\theta}^\star\|_2$ and satisfies the same bound. Hence, in both
cases,}
{
\begin{equation*}
    \left(\|\tilde{\theta}^\star\|_2-G_{m-1}\right)
    \mathbf{1}_{\mathcal{E}_{m-1}}
    \le 2\|\tilde{\theta}^\star\|_2
    < 8 \frac{\|\tilde{\theta}^\star -
    \hat{\tilde{\theta}}_{m-1}\|_2^2}
    {\|\tilde{\theta}^\star\|_2}.
\end{equation*}
}

In both cases, the geometric regret multiplier is bounded by $8 \frac{\|\tilde{\theta}^\star - \hat{\tilde{\theta}}_{m-1}\|_2^2}{\|\tilde{\theta}^\star\|_2}$ without requiring any hidden assumptions on the size of $T$. 

Under the sample splitting procedure defined in Algorithm 5, the estimator $\hat{\tilde{\theta}}_{m-1}$ is computed using the independent dataset $T^{(m-1)}$. Rather than evaluating the unconditional expected trace—which could technically diverge for unregularized OLS—we directly bound the restricted expectation in the whitened space. 

Because the concentration occurs directly in the isotropic whitened space under event $\mathcal{E}_{m-1}$, the whitened estimation error is deterministically bounded by $\|\hat{\tilde{\theta}}_{m-1} - \tilde{\theta}^\star\|_2^2 \le \mathcal{O}\left(\frac{d M \log T}{t_{m-1}}\right)$, and crucially, this bound is entirely free of $\kappa$. We safely bound the conditional trace:
\begin{equation*}
    \mathbb{E}[\|\hat{\tilde{\theta}}_{m-1} - \tilde{\theta}^\star\|_2^2 \mathbf{1}_{\mathcal{E}_{m-1}}] \le \mathcal{O}\left( \frac{d M \log T}{t_{m-1}} \right) \mathbb{P}(\mathcal{E}_{m-1}) \le \mathcal{O}\left( \frac{d M \log T}{t_{m-1}} \right).
\end{equation*}

Substituting this rigorously bounded trace back into our expected instantaneous regret equation, and converting the whitened norm back to the original space via $\|\tilde{\theta}^\star\|_2 = \sqrt{\theta^{\star \top} \Sigma \theta^\star} \ge \sqrt{\kappa/d} \|\theta^\star\|_2$, we obtain the expected instantaneous regret under the good event for the $m$-th batch:
\begin{align*}
    \mathbb{E}[r_t \mathbf{1}_{\mathcal{E}_{m-1}}] &\le C_K \frac{\mathbb{E}[8\|\hat{\tilde{\theta}}_{m-1} - \tilde{\theta}^\star\|_2^2 \mathbf{1}_{\mathcal{E}_{m-1}}]}{\|\tilde{\theta}^\star\|_2} \\
    &\le 8 C_K \frac{\mathcal{O}(d M \log T / t_{m-1})}{\sqrt{\kappa/d} \|\theta^\star\|_2} = \mathcal{O}\left( \frac{d^{3/2} M \sqrt{\log K} \log T}{\sqrt{\kappa} t_{m-1} \|\theta^\star\|_2} \right).
\end{align*}

For the initial pure-exploration batch ($m=1$), uniform sampling incurs an expected maximum reward gap safely bounded by $\mathcal{O}(\sqrt{(\log K)/d})$. Over $t_1 = \Theta(b d^2)$ steps, this purely exploratory regret absorbs into the subsequent limits.

Integrating our instantaneous upper bound over the current batch duration $\Delta t_m = t_m - t_{m-1}$ yields the expected cumulative regret for batch $m \ge 2$:
\begin{equation*}
    \sum_{t=t_{m-1}+1}^{t_m} \mathbb{E}[r_t] \le \mathcal{O}\left( \frac{d^{3/2} M \sqrt{\log K} \log T}{\sqrt{\kappa} \|\theta^\star\|_2} \right) \frac{\Delta t_m}{t_{m-1}}.
\end{equation*}

Given our optimal geometric grid $t_m = \lfloor b t_{m-1} \rfloor$ with scaling factor $b = \Theta((T/d^2)^{1/M})$, the fractional length of the batch evaluates to $\frac{\Delta t_m}{t_{m-1}} \le \frac{(b-1)t_{m-1}}{t_{m-1}} < b = \Theta\left(\left(\frac{T}{d^2}\right)^{1/M}\right)$.

Summing this cumulative regret uniformly across all $M$ batches introduces an additional scaling factor of $M$. We absorb the logarithmic and batch-count penalties, $\mathcal{O}(M^2 \log T)$, into the $\mathsf{polylog}(T)$ term:
\begin{equation*}
    \mathbb{E} [R_T(\pi) \mid \theta^\star] \le \mathsf{polylog}(T) \cdot \frac{d^{3/2} \sqrt{\log K}}{\sqrt{\kappa} \|\theta^\star\|_2} \left( \frac{T}{d^2} \right)^{\frac{1}{M}}.
\end{equation*}
\end{proof}

}

\ZH{
\subsection{Regret Analysis for the Lower Bound} 
The proof adapts the targeted localized prior technique from the lower bound proof of Theorem~\ref{thm.stochastic} to the scaled regret metric $\|\theta^\star\|_2 \cdot \mathbb{E}[R_T]$. A key simplification arises because multiplying the geometric regret identity by $\|\theta^\star\|_2$ cancels the $1/\|\theta^\star\|_2$ denominator, leaving the orthogonal variance as the sole quantity to control. This avoids the global prior and its accompanying $\log T$ penalty.

The algorithm defines a batch grid $0 = t_0 < t_1 < \dots < t_M = T$. During the $m$-th batch ($t_{m-1} < t \le t_m$), the policy purely relies on the history $\mathcal{H}_{t_{m-1}}$ collected up to step $t_{m-1}$.

By Lemma~\ref{lemma.geometric_regret_identity}, the expected conditional instantaneous regret for any step $t \in (t_{m-1}, t_m]$ satisfies:
\[
\mathbb{E}_{X_t}[r_t \mid \theta^\star] \ge \frac{C_K}{2\sqrt{d}\|\theta^\star\|_2} \|(I - u_t u_t^\top)\theta^\star\|_2^2,
\]
where $u_t$ is the normalized conditionally expected selected context. Multiplying both sides by the parameter norm $\|\theta^\star\|_2$ neatly isolates the orthogonal variance:
\begin{equation}
    \|\theta^\star\|_2 \mathbb{E}_{X_t}[r_t \mid \theta^\star] \ge \frac{C_K}{2\sqrt{d}} \|(I - u_t u_t^\top)\theta^\star\|_2^2. \label{eq:scaled_regret_identity}
\end{equation}

Let $s = \min\{m \in [M] : t_m \ge d^2\}$. The adversary constructs $M-s+1$ distinct testing priors targeted at different batch intervals. For any targeted batch $j \in \{s+1, \dots, M\}$, the adversary selects the prior $\rho_j = \mathcal{N}(0, \frac{\nu_j^2}{d} I_d)$ with scale $\nu_j = \frac{d}{2\sqrt{\gamma_K t_{j-1}}}$. Because $t_{j-1} \ge t_s \ge d^2$, we have $\nu_j \le 1/2$, so $2\nu_j \le 1$.

Let $E_j = \{\|\theta^\star\|_2 \le 2\nu_j\}$, ensuring the parameter remains within the unit ball. Restricting \eqref{eq:scaled_regret_identity} to event $E_j$ and taking the expectation over $\rho_j$:
\[
\mathbb{E}_{\rho_j}\left[\|\theta^\star\|_2 \cdot \mathbb{E}[r_t] \cdot \mathbbm{1}_{E_j}\right] \ge \frac{C_K}{2\sqrt{d}}\, \mathbb{E}_{\rho_j}\left[\|(I - u_t u_t^\top)\theta^\star\|_2^2 \cdot \mathbbm{1}_{E_j}\right].
\]
Using $\mathbbm{1}_{E_j}+\mathbbm{1}_{E_j^c}=1$ and linearity of expectation, we have $\mathbb{E}[X\mathbbm{1}_{E_j}]=\mathbb{E}[X]-\mathbb{E}[X\mathbbm{1}_{E_j^c}]$. Applying Lemma~\ref{lemma.orthogonal_trace} to the orthogonal variance under $\rho_j$, and noting the tail error $\mathbb{E}_{\rho_j}[\|\theta^\star\|_2^2 \mathbbm{1}_{\|\theta^\star\|_2 > 2\nu_j}] \le 0.092\nu_j^2$ for $d \ge 2$:
\[
\mathbb{E}_{\rho_j}\left[\|(I - u_t u_t^\top)\theta^\star\|_2^2 \cdot \mathbbm{1}_{E_j}\right] \ge \frac{(d-1)^2}{d^2/\nu_j^2 + \gamma_K t_{m-1}} - 0.092\nu_j^2.
\]
For any step $t \le t_j$, the associated batch $m$ satisfies $t_{m-1} \le t_{j-1}$. Substituting $\nu_j$ yields $d^2/\nu_j^2 + \gamma_K t_{m-1} \le 4\gamma_K t_{j-1} + \gamma_K t_{j-1} = 5\gamma_K t_{j-1}$. The trace term evaluates to $\frac{(d-1)^2}{5\gamma_K t_{j-1}} = \frac{4(d-1)^2}{5d^2}\nu_j^2 \ge 0.2\nu_j^2$ for $d \ge 2$. Subtracting the tail error gives a net positive margin of $(0.2 - 0.092)\nu_j^2 = 0.108\nu_j^2$. Therefore, the per-step Bayesian expected scaled regret for every step $1 \le t \le t_j$ satisfies:
\[
\mathbb{E}_{\rho_j}\left[\|\theta^\star\|_2 \cdot \mathbb{E}[r_t] \cdot \mathbbm{1}_{E_j}\right] \ge \frac{0.108\, C_K}{2\sqrt{d}} \nu_j^2 = \frac{0.054\, C_K}{\sqrt{d}} \cdot \frac{d^2}{4\gamma_K t_{j-1}} = \frac{c_0\, C_K\, d^{3/2}}{\gamma_K\, t_{j-1}},
\]
where $c_0 = 0.0135$. Summing from step 1 to $t_j$:
\begin{equation}
\mathbb{E}_{\rho_j}\left[\|\theta^\star\|_2 \cdot \mathbb{E}[R_T] \cdot \mathbbm{1}_{E_j}\right] \ge \sum_{t=1}^{t_j} \frac{c_0\, C_K\, d^{3/2}}{\gamma_K\, t_{j-1}} = \frac{c_0\, C_K}{\gamma_K} \cdot d^{3/2} \cdot \frac{t_j}{t_{j-1}}. \label{eq:pd_lower_bound_tj}
\end{equation}
By standard minimax theory, since $E_j \subseteq \{\|\theta^\star\|_2 \le 1\}$:
\[
\sup_{\|\theta^\star\|_2 \le 1} \|\theta^\star\|_2 \cdot \mathbb{E}[R_T(\textbf{Alg}, \theta^\star)] \ge \mathbb{E}_{\rho_j}\left[\|\theta^\star\|_2 \cdot \mathbb{E}[R_T] \cdot \mathbbm{1}_{E_j}\right].
\]

By mirroring the argument for the initialization batch $j = s$ (with $\nu_s = 1/2$ and $E_s = \{\|\theta^\star\|_2 \le 1\}$), the same analysis yields a bound of $c_0'\, C_K\, t_s / \sqrt{d}$, where $c_0' > 0$ is a universal constant. Since the minimax regret dominates each adversarial target, we acquire the sequence maximum:
\[
\sup_{\|\theta^\star\|_2 \le 1} \|\theta^\star\|_2 \cdot \mathbb{E}[R_T] \ge c_1\, C_K \max\left\{\frac{t_s}{\sqrt{d}},\; \frac{d^{3/2}\, t_{s+1}}{t_s},\; \frac{d^{3/2}\, t_{s+2}}{t_{s+1}},\; \dots,\; \frac{d^{3/2}\, T}{t_{M-1}}\right\},
\]
where $c_1 > 0$ absorbs the structural constants $\gamma_K$, $c_0$, and $c_0'$.

Let $\alpha$ denote the above maximum. From $\alpha \ge t_s / \sqrt{d}$ we obtain $t_s \le \alpha\sqrt{d}$. Taking the product of the remaining $M - s$ inequalities:
\[
\alpha^{M-s} \ge d^{3(M-s)/2} \cdot \frac{T}{t_s} \ge d^{3(M-s)/2} \cdot \frac{T}{\alpha\sqrt{d}}.
\]
Rearranging:
\[
\alpha^{M-s+1} \ge d^{(3(M-s)-1)/2} \cdot T.
\]
Therefore $\alpha \ge d^{(3(M-s)-1)/(2(M-s+1))} \cdot T^{1/(M-s+1)}$. A direct computation shows:
\begin{align*}
\frac{3(M-s)-1}{2(M-s+1)} &= \frac{3}{2} - \frac{2}{M-s+1},
\end{align*}
so that $d^{(3(M-s)-1)/(2(M-s+1))} \cdot T^{1/(M-s+1)} = d^{3/2} \cdot (T/d^2)^{1/(M-s+1)}$. Since $s \ge 1$ implies $M - s + 1 \le M$, and $T/d^2 \ge 2^M \ge 1$:
\[
\alpha \ge d^{3/2} \left(\frac{T}{d^2}\right)^{1/(M-s+1)} \ge d^{3/2} \left(\frac{T}{d^2}\right)^{1/M}.
\]
Re-substituting $\alpha$ with $C_K = \Omega(\sqrt{\log K})$ and $\gamma_K = \mathcal{O}(1)$:
\[
\sup_{\|\theta^\star\|_2 \le 1} \|\theta^\star\|_2 \cdot \mathbb{E}[R_T(\textbf{Alg}, \theta^\star)] \ge c \cdot \sqrt{\log K} \cdot d^{3/2} \left(\frac{T}{d^2}\right)^{1/M}.
\]
\halmos
}

\bibliographystyle{plainnat}
\bibliography{di}

\end{document}